%% file: main.tex
\documentclass[10pt,twocolumn,letterpaper]{article}

\usepackage[pagenumbers]{main}

\usepackage[table]{xcolor}
\usepackage{microtype}
\usepackage{graphicx}
\usepackage{booktabs}
\usepackage{amsmath}
\usepackage{amssymb}
\usepackage{mathtools}
\usepackage{amsthm}

\usepackage{color}
\usepackage{enumitem}
\usepackage{multirow}
\usepackage{makecell}

\usepackage{algorithmic}
\usepackage{algorithm}
\usepackage{etoolbox,siunitx}

\usepackage{fontawesome}
\usepackage{academicons}

\definecolor{robo_blue}{RGB}{66, 133, 244}
\definecolor{robo_red}{RGB}{231, 66, 52}
\definecolor{robo_yellow}{RGB}{251, 189, 5}
\definecolor{robo_green}{RGB}{51, 168, 82}
\definecolor{robo_gray}{RGB}{165, 165, 165}

\definecolor{lblue}{RGB}{66, 133, 244}

\usepackage[pagebackref,breaklinks,colorlinks,citecolor=lblue]{hyperref}

\def\paperID{1}
\def\confName{RoboDrive}
\def\confYear{2024}

\title{\includegraphics[height=0.47cm]{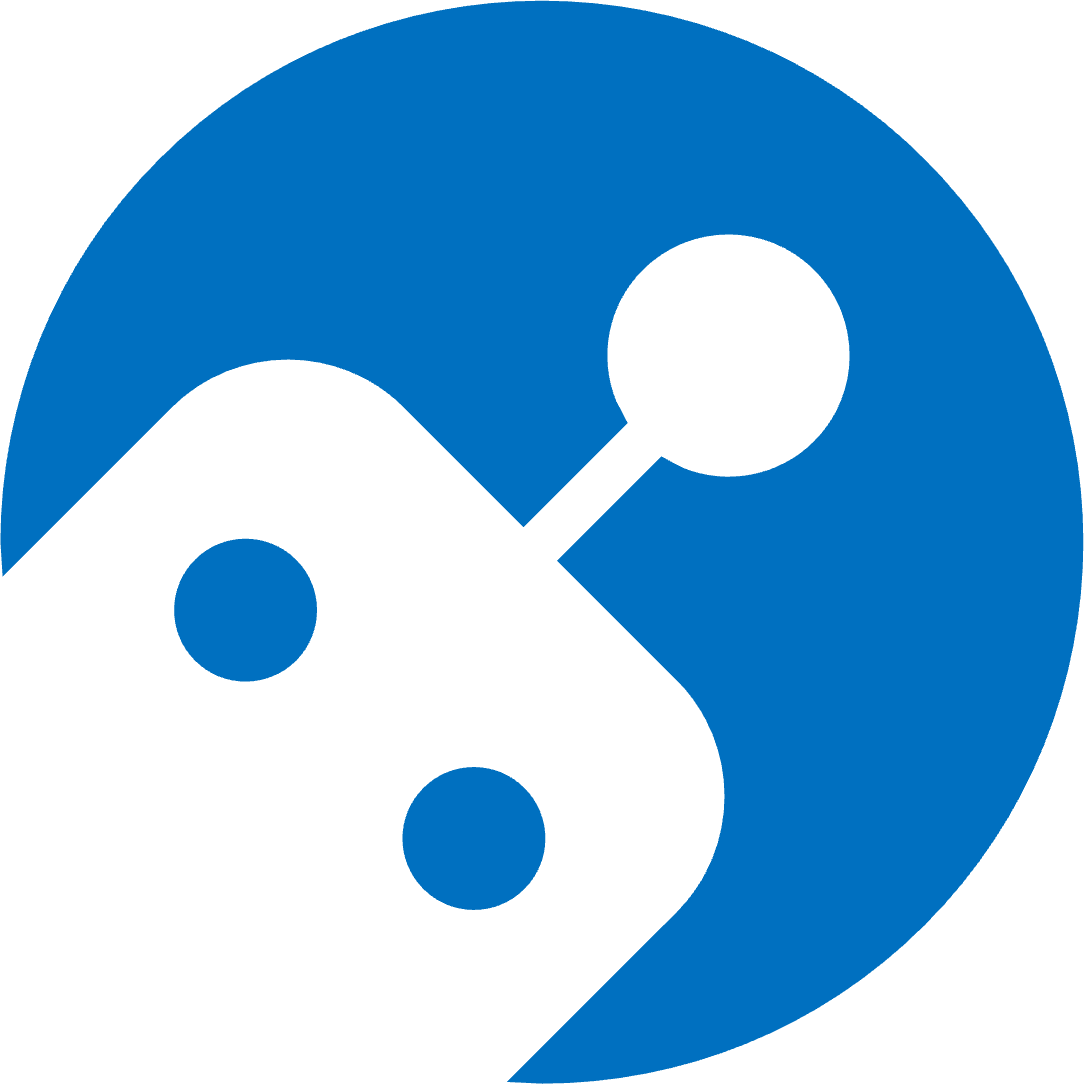}~The RoboDrive Challenge:\\Drive Anytime Anywhere in Any Condition}

\author{
    \textbf{Challenge \& Workshop Organizers}\\
    Lingdong Kong \quad Shaoyuan Xie \quad Hanjiang Hu \quad Yaru Niu \quad Wei Tsang Ooi\\
    Benoit R. Cottereau \quad Lai Xing Ng \quad Yuexin Ma \quad Wenwei Zhang \quad Liang Pan\\
    Kai Chen \quad Ziwei Liu\\
    \url{https://robodrive-24.github.io}\\
\\
    \textbf{Technical Committee}\\
    Weichao Qiu \quad Wei Zhang\\
\\
    \textbf{Challenge Participants}\\
    Xu Cao \quad Hao Lu \quad Ying-Cong Chen \quad Caixin Kang \quad Xinning Zhou \quad Chengyang Ying\\
    Wentao Shang \quad Xingxing Wei \quad Yinpeng Dong \quad Bo Yang \quad Shengyin Jiang \quad Zeliang Ma\\
    Dengyi Ji \quad Haiwen Li \quad Xingliang Huang \quad Yu Tian \quad Genghua Kou \quad Fan Jia \quad Yingfei Liu\\
    Tiancai Wang \quad Ying Li \quad Xiaoshuai Hao \quad Yifan Yang \quad Hui Zhang \quad Mengchuan Wei\\
    Yi Zhou \quad Haimei Zhao \quad Jing Zhang \quad Jinke Li \quad Xiao He \quad Xiaoqiang Cheng \quad Bingyang Zhang\\
    Lirong Zhao \quad Dianlei Ding \quad Fangsheng Liu \quad Yixiang Yan \quad Hongming Wang \quad Nanfei Ye\\
    Lun Luo \quad Yubo Tian \quad Yiwei Zuo \quad Zhe Cao \quad Yi Ren \quad Yunfan Li \quad Wenjie Liu \quad Xun Wu\\
    Yifan Mao \quad Ming Li \quad Jian Liu \quad Jiayang Liu \quad Zihan Qin \quad Cunxi Chu \quad Jialei Xu\\
    Wenbo Zhao \quad Junjun Jiang \quad Xianming Liu \quad Ziyan Wang \quad Chiwei Li \quad Shilong Li\\
    Chendong Yuan \quad Songyue Yang \quad Wentao Liu \quad Peng Chen \quad Bin Zhou \quad Yubo Wang\\
    Chi Zhang \quad Jianhang Sun \quad Hai Chen \quad Xiao Yang \quad Lizhong Wang \quad Dongyi Fu\\
    Yongchun Lin \quad Huitong Yang \quad Haoang Li \quad Yadan Luo \quad Xianjing Cheng \quad Yong Xu 
}

\begin{document}

\maketitle

\input{sections/0_abstract}

\clearpage
\twocolumn[{%
\renewcommand\twocolumn[1][]{#1}%
\begin{center}
    \centering
    \vspace{-8pt}
    \captionsetup{type=figure}
    \includegraphics[width=\textwidth]{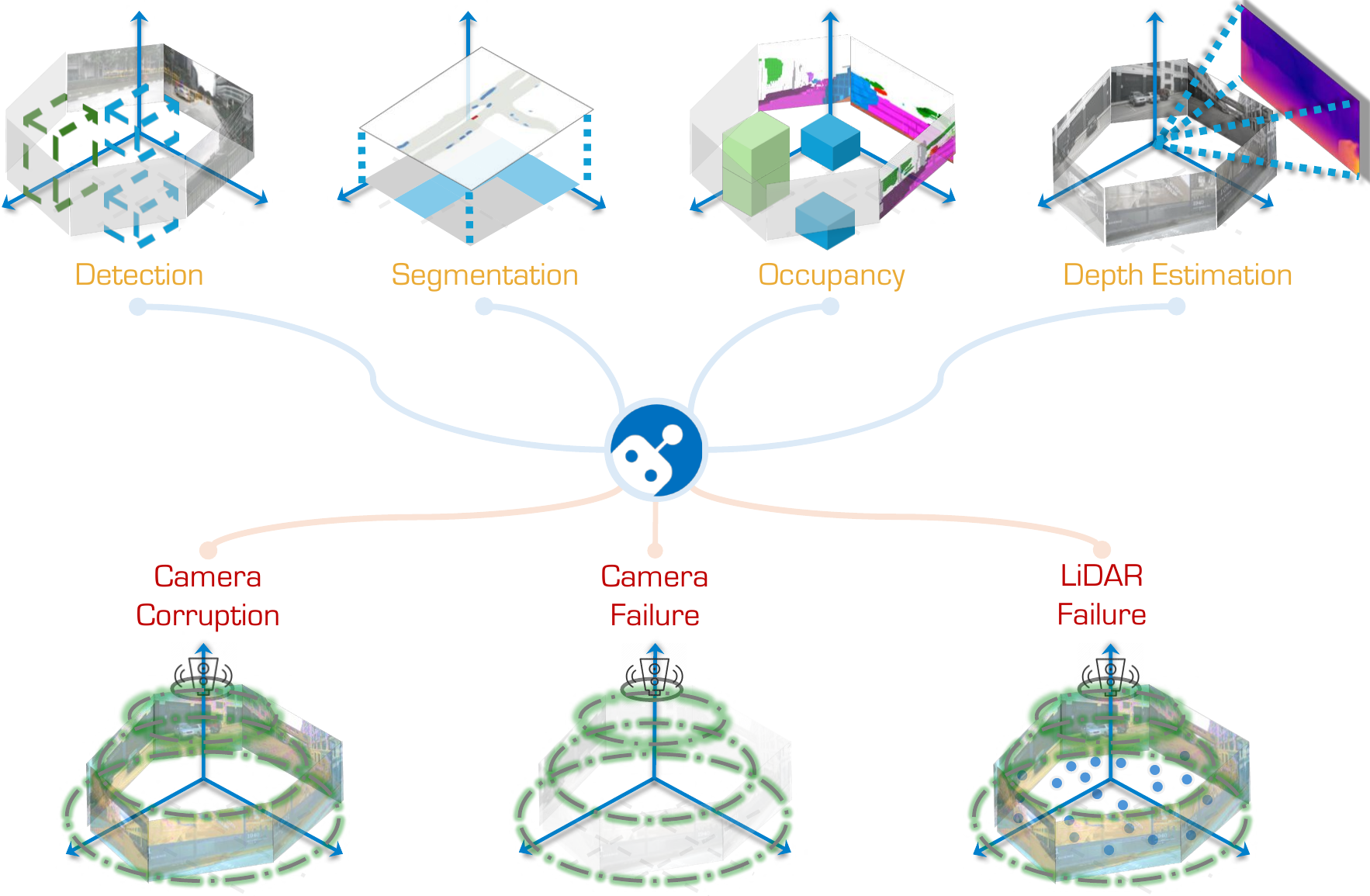}
    \vspace{-13.5pt}
    \captionof{figure}{\textbf{Challenge overview}. The 2024 RoboDrive challenge aims to facilitate and encourage innovative solutions for tackling mainstream driving perception tasks under out-of-distribution (OoD) scenarios that occur in the real world. We are particularly interested in enhancing the OoD robustness of 3D object detection, HD map segmentation, semantic occupancy prediction, and multi-view depth estimation algorithms in challenging and unprecedented conditions, such as camera corruptions, camera failures, and LiDAR failures.}
    \label{fig:teaser}
    \vspace{6.36pt}
\end{center}
}]

\input{sections/1_intro}
\input{sections/2_related_work}
\input{sections/3_approach}

\input{sections/4_solutions}

\input{sections/5_conclusion}

\input{sections/6_appendix}

{\small
\bibliographystyle{ieeenat_fullname}
\bibliography{main}
}

\end{document}

%% file: sections/0_abstract.tex
\begin{abstract}
\vspace{-0.2cm}
In the realm of autonomous driving, robust perception under out-of-distribution conditions is paramount for the safe deployment of vehicles. Challenges such as adverse weather, sensor malfunctions, and environmental unpredictability can severely impact the performance of autonomous systems. The 2024 RoboDrive Challenge was crafted to propel the development of driving perception technologies that can withstand and adapt to these real-world variabilities. Focusing on four pivotal tasks -- BEV detection, map segmentation, semantic occupancy prediction, and multi-view depth estimation -- the competition laid down a gauntlet to innovate and enhance system resilience against typical and atypical disturbances. This year's challenge consisted of five distinct tracks and attracted 140 registered teams from 93 institutes across 11 countries, resulting in nearly one thousand submissions evaluated through our servers. The competition culminated in 15 top-performing solutions, which introduced a range of innovative approaches including advanced data augmentation, multi-sensor fusion, self-supervised learning for error correction, and new algorithmic strategies to enhance sensor robustness. These contributions significantly advanced the state of the art, particularly in handling sensor inconsistencies and environmental variability. Participants, through collaborative efforts, pushed the boundaries of current technologies, showcasing their potential in real-world scenarios. Extensive evaluations and analyses provided insights into the effectiveness of these solutions, highlighting key trends and successful strategies for improving the resilience of driving perception systems. This challenge has set a new benchmark in the field, providing a rich repository of techniques expected to guide future research in this field. All resources have been publicly available on the challenge website to encourage ongoing community engagement and the development of robust driving perception.
\vspace{-0.4cm}
\end{abstract}

%% file: sections/1_intro.tex
\section{Introduction}
\label{sec:intro}

The evolution of vision-centric autonomous driving hinges critically on robust perception systems that interpret and respond accurately to complex and dynamic environments \cite{caesar2020nuscenes,geiger2013vision,sun2020scalability}. As shown in \cref{fig:teaser}, central to these systems are four mainstream perception tasks: bird's eye view (BEV) detection, HD map segmentation, semantic occupancy prediction, and multi-view depth estimation \cite{ma2022vision}. Mastery of these tasks ensures that vehicles can navigate safely and interact intelligently with their surroundings \cite{li20224bev,li2023bevdepth,huang2021bevdet,yang2023bevformer}.

BEV detection creates a comprehensive aerial perspective of the vehicle's surroundings, which is crucial for understanding the placement and movement of various elements such as vehicles, pedestrians, and static obstacles \cite{li2022bevformer}. This overview aids in route planning and obstacle avoidance by providing a clear picture of the environment from above \cite{huang2021bevdet}.
Map segmentation delineates drivable areas from potential hazards by segmenting road elements \cite{li2022hdmapnet}. It plays a pivotal role in navigation by clarifying paths and identifying safe zones, ensuring that the vehicle can plan routes efficiently and react to changes in the driving environment \cite{peng2023bevsegformer}. Semantic occupancy prediction forecasts the potential occupancy around the vehicle with corresponding semantic information \cite{wei2023surroundocc,tian2024occ3d}. This prediction is vital for dynamic driving decisions, helping to identify areas that might soon be occupied by other vehicles or pedestrians, thereby enhancing situational awareness \cite{cao2022monoscene}. Monocular, stereo, and multi-view depth estimation utilize data from multiple sensors to generate accurate depth maps of the surrounding environment \cite{wei2023surrounddepth,watson2019hints}. This task is essential for precise distance measurement and spatial understanding, especially in complex scenarios where single-view data might be insufficient \cite{godard2019monodepth2}.

Autonomous vehicles must operate reliably under a variety of conditions, many of which may deviate significantly from the environments seen during the training of their perception systems \cite{xie2023robobev,xie2024benchmarking}. These out-of-distribution (OoD) scenarios, such as adverse weather, sensor malfunctions, and unpredictable environmental changes, pose significant challenges \cite{hendrycks2019benchmarking}. They can severely impact the performance of perception systems, leading to degraded situational awareness and potentially unsafe driving decisions \cite{kong2023robo3d,kong2024robodepth,beemelmanns2024multicorrupt}.

To address these challenges, the 2024 RoboDrive Challenge was conceived. This competition aims to propel the development of driving perception technologies capable of withstanding and adapting to real-world variabilities. By focusing on the aforementioned tasks, the challenge encourages the exploration of innovative approaches that enhance system resilience against typical and atypical disturbances, such as camera corruptions, camera failures, and LiDAR failures (see \cref{fig:teaser}). 

This year's RoboDrive Challenge, held at the 41st IEEE Conference on Robotics and Automation (ICRA 2024), featured five distinct tracks centered around these tasks. It attracted significant international participation, with 140 teams from 93 institutions across 11 countries submitting nearly one thousand entries. The contributions from this challenge have significantly advanced the state of the art, particularly in handling sensor inconsistencies and environmental variability. A total of 18 camera corruption and failure cases and 3 LiDAR failure scenarios are involved in this competition, providing a holistic evaluation for the OoD robustness of different driving perception models.

The participants employed various advanced methodologies to tackle the challenge, including:
\begin{itemize}
    \item \textit{Advanced Data Augmentation Techniques:} Teams implemented sophisticated augmentation strategies such as frequency domain manipulations and realistic environmental simulations. These methods were designed to train robust models capable of handling unexpected variations, thereby enhancing their adaptability to real-world conditions.
    
    \item \textit{Multi-Sensor Fusion:} Demonstrating the power of integrated systems, several solutions effectively merged data from multiple modalities, including cameras and LiDAR. This integration not only enhanced detection reliability but also maintained accuracy, especially under conditions where sensor functionality might be compromised.
    
    \item \textit{Self-Supervised Learning for Sensor Error Correction:} By employing self-supervised learning techniques such as masked modeling and contrastive learning, participants could reconstruct and refine data from corrupted sensors, substantially boosting the resilience of their systems.
    
    \item \textit{Innovative Algorithmic Approaches:} Through the development of novel algorithms, teams were able to achieve robust feature extraction, intricate sensor data fusion, and improved predictive accuracy. These innovations mark significant strides in the field of robust perception.
    
    \item \textit{Systematic Robustness against Environmental and Sensor Variabilities:} Addressing a wide range of adversities -- from adverse weather conditions to variations in lighting and sensor anomalies -- participants ensured that their systems delivered consistent performance across an array of challenging scenarios.
\end{itemize}

The collective efforts and achievements of this year's RoboDrive Challenge highlight the dynamic and collaborative nature of the field. These pioneering efforts underscore the challenge's role in pushing the boundaries of technological innovation and fostering a competitive yet cooperative environment for exploring new solutions. The competition not only brought forward novel methodologies but also set the stage for future developments in autonomous driving technologies. This synergy of competition and innovation is vital for accelerating progress and encouraging the practical application of research findings in real-world scenarios.

The remainder of this paper is organized as follows. \cref{sec:related_work} reviews recent advancements in autonomous driving perception and summarizes relevant challenges and competitions. \cref{sec:robodrive} details the key statistics, public resources, and terms and conditions of this challenge. \cref{sec:solution} presents detailed discussions of the innovative solutions and implementation details from the top-performing teams across the five tracks. \cref{sec:discussions} and \cref{sec:conclusion} offer discussion of future research directions and concluding remarks, respectively. Finally, \cref{sec:appendix} contains acknowledgments and additional information about this challenge.

%% file: sections/2_related_work.tex
\section{Related Work}
\label{sec:related_work}

\subsection{BEV Detection}
3D object detection is a cornerstone task in autonomous driving systems. Traditional methods use point clouds as the input for extracting spatial information \cite{lang2019pointpillars,yin2021center,cbgs}. Recently, the domain has witnessed notable advancements in monocular 3D object detection, largely attributed to popular benchmarks~\cite{geiger2013vision,caesar2020nuscenes,sun2020scalability}. This progress has been driven by a range of research studies. Nevertheless, monocular systems face limitations due to their dependence on a single viewpoint and limited data, which hampers their capacity to manage complex scenarios. To overcome these limitations, extensive benchmarks have been developed, offering surrounding view data that enriches the field of multi-view 3D object detection and propels the evolution of advanced detection methodologies~\cite{caesar2020nuscenes,sun2020scalability}.
The multi-view 3D object detection research can be primarily divided into dense bird’s eye view (BEV)-based~\cite{li2022bevformer,li2023bevdepth, huang2021bevdet,jiang2023polarformer, yang2023bevformer} and sparse query-based algorithms~\cite{wang2022detr3d,liu2022petr, liu2023petrv2,lin2022sparse4d}. Dense BEV-based algorithms convert multi-view image features into a dense BEV representation using the Lift-Splat-Shoot (LSS) module~\cite{li2023bevdepth,huang2021bevdet} or directly using attention-based transformation~\cite{li2022bevformer}. To mitigate overfitting from LSS, BEVDet~\cite{huang2021bevdet} recommends data augmentation on BEV features. BEVDepth \cite{li2023bevdepth} introduces depth estimation supervision to enhance LSS precision. Conversely, sparse query-based algorithms such as DETR3D~\cite{wang2022detr3d} and PETR~\cite{liu2022petr} utilize learnable 3D object queries and position encoding to engage with multi-view 2D features. Recent studies have integrated temporal information to refine these methods further~\cite{park2022time, huang2022bevdet4d}.

\subsection{BEV Map Segmentation}
High-definition maps (or HD maps) are crucial for providing detailed information about road environments, which is essential for autonomous vehicles. Traditionally, HD maps are manually annotated, though some studies leverage SLAM algorithms to create HD maps from repeated LiDAR sensor scans~\cite{dube2017online,lee2013robust,mendes2016icp, shan2020lio, yang2018robust}. These methods require extensive data collection, prolonged iterative processes, and costly human annotations. Consequently, generating local maps directly from onboard sensors has emerged as a cost-effective alternative. Moreover, the capability to create semantic maps online enhances system redundancy. HDMapNet~\cite{li2022hdmapnet} was the first to address local semantic map learning, focusing on constructing semantic maps online using data from LiDAR sensors and cameras. It introduced a method to derive BEV features from sensor inputs and predict vectorized map elements. BEVSegFormer~\cite{peng2023bevsegformer} later introduced multi-camera deformable attention to converting image-view features into BEV representations for semantic map creation. More recently, BEVerse~\cite{zhang2022beverse} proposes a multi-task learning framework to enhance the performance further.

\subsection{Semantic Occupancy Prediction}
Semantic occupancy prediction aims to predict the occupancy and semantic labels of 3D scenes, unlike occupancy grid mapping~\cite{moravec1985high, thrun2002probabilistic}, which generates probabilistic maps from range measurements and assumes static scenes. This task does not require range sensors, making it more versatile. Semantic Scene Completion (SSC) closely relates to semantic occupancy prediction, inferring dense 3D occupancy with semantic labels. MonoScene \cite{cao2022monoscene} employs a U-Net for this from monocular RGB images. TPVFormer \cite{huang2023tri} extends to multi-camera input but lacks dense supervision. Recently, improved pipelines are now able to generate dense occupancy ground truth for better training~\cite{wei2023surroundocc, tian2024occ3d}. Methods like VoxFormer~\cite{li2023voxformer} and OccDepth~\cite{miao2023occdepth} use depth-aware techniques to further enhance the performance. Another task that is closely related to SSC is LiDAR semantic segmentation \cite{milioto2019rangenet++,kong2023conDA,liu2023uniseg,liu2023segment,liu2024m3net}. Sharing a similar objective, the LiDAR segmentation models aim to segment the holistic 3D scenes acquired by the LiDAR sensor \cite{choy2019minkowski,kong2022laserMix,kong2023rethinking,kong2024lasermix2}. It is worth mentioning that the SSC labels are often obtained by concatenating multi-frame semantic LiDAR point clouds \cite{chen2023clip2Scene,hong20224dDSNet,li2023sscbench}.

\subsection{Multi-View Depth Estimation}
Depth estimation from multi-view inputs is an emerging area of research that addresses the limitations of monocular systems by synthesizing information across multiple sensory inputs \cite{vehicles,mvsnet}. This approach is vital for achieving precise depth measurements in environments where single-viewpoint data is insufficient to make reliable navigational decisions \cite{AFNet}. For example, unlike traditional monocular depth estimation, modern in-vehicle sensing systems can provide a comprehensive $360^{\circ}$ view of the surroundings \cite{caesar2020nuscenes,geiger2013vision}, enhancing the vehicle's ability to perceive and interact with its environment more effectively. Notable works in this field include SurroundDepth \cite{wei2023surrounddepth} and S3Depth \cite{S3Depth}, which propose leveraging consistency across different views to enhance depth predictions. Although these models demonstrate significant improvements in depth accuracy by integrating features from multiple cameras, their robustness under challenging conditions remains unknown \cite{kong2024robodepth,kong2023robodepth_benchmark}. A robust depth estimation system that can suppress real-world corruptions is crucial for navigation and obstacle avoidance in autonomous driving.

\subsection{Out-of-Distribution Robustness}
The resilience of models against unexpected disturbances, \ie, out-of-distribution (OoD), is a critical area of focus in machine learning, particularly under conditions that deviate from typical training environments. Several benchmarks such as ImageNet-C~\cite{hendrycks2019benchmarking}, ObjectNet~\cite{barbu2019objectnet}, and ImageNetV2~\cite{recht2019imagenet} have been established to systematically evaluate the robustness of 2D image classifiers against a diverse range of corruptions. For example, ImageNet-C \cite{hendrycks2019benchmarking} introduces controlled perturbations to pristine ImageNet \cite{imagenet} samples, including compression artifacts and motion blur, to test classifier resilience. Conversely, ObjectNet \cite{barbu2019objectnet} challenges models with a test set characterized by significant variations in object rotation, background, and viewpoint, emphasizing the importance of adaptability in real-world applications. Hendrycks \etal \cite{hendrycks2021many} highlight the direct correlation between a model's ability to withstand synthetic corruptions and its performance in practical settings. In the realm of 3D perception, recent initiatives have aimed to extend these concepts to more complex models.  
RoboDepth \cite{kong2024robodepth,kong2023challenge} establishes a robustness benchmark for monocular depth estimation under various corruptions. Robo3D \cite{kong2023robo3d} aims to probe the robustness of LiDAR-based perception models in adverse weather conditions and sensor malfunctions. RoboBEV \cite{xie2023robobev,xie2024benchmarking} and MetaBEV propose to understand the BEV detection model's robustness under external distributions. Most recently, MultiCorrupt \cite{beemelmanns2024multicorrupt} combines several 2D and 3D corruptions together to further analyze the robustness of multi-modal perception models.

\subsection{Relevant Challenges}
Numerous competitions have significantly propelled the field of depth estimation and robust visual perception. The Robust Vision Challenge (RVC) \cite{RVC} explored a broad spectrum of scene understanding tasks including reconstruction, optical flow, semantic segmentation, and depth prediction, pushing the boundaries of visual systems' capabilities.
The Dense Depth for Autonomous Driving (DDAD) Challenge \cite{DDAD} specialized in depth estimation under diverse urban conditions, crucial for autonomous navigation. The SeasonDepth Depth Prediction Challenge \cite{hu2022seasondepth} adapted depth models to various lighting and seasonal conditions, while the Monocular Depth Estimation Challenge (MDEC) \cite{MDEC,MDEC2} focused on depth perception in complex natural landscapes. The Argoverse Stereo Competition \cite{ArgoverseStereo} targeted real-time stereo depth estimation for driving applications, and the NTIRE 2023 Challenge \cite{NTIRE} tackled depth estimation from non-Lambertian surfaces, dealing with specular and transparent materials.
The RoboDepth Challenge \cite{kong2023robodepth_challenge} introduced robust OoD depth estimation to combat real-world corruptions, including adverse weather, sensor failures, and noise contamination. Our competition shares a similar spirit with RoboDepth \cite{kong2023robodepth_challenge} and further extends such endeavors to the mainstream driving perception tasks, \ie, BEV detection, HD map segmentation, semantic occupancy prediction, and multi-view depth estimation. We hope this competition could facilitate and encourage future research on designing more robust driving perception techniques.

%% file: sections/3_approach.tex
\section{RoboDrive Challenge 2024}
\label{sec:robodrive}

Our competition presents an opportunity to test the limits of driving technologies under out-of-distribution conditions. This year, the challenge emphasizes the resilience and adaptability of driving perception systems across various unpredictable scenarios. The sections below outline the specifics of data preparation, evaluation metrics, baseline models, and highlight the notable outcomes of the competition.

\subsection{Overview}
The 2024 RoboDrive Challenge, held in conjunction with the 41st IEEE Conference on Robotics and Automation (ICRA), represents a significant stride toward advancing autonomous driving technologies under challenging conditions. This annual competition is designed to test and enhance the robustness of autonomous driving systems against a spectrum of real-world challenges such as adverse weather, sensor malfunctions, and environmental unpredictability, which are critical for the safe deployment of autonomous vehicles. This year's challenge emphasized innovation in driving perception across four essential tasks: BEV detection, map segmentation, semantic occupancy prediction, and multi-view depth estimation. These tasks are pivotal in developing systems that can perceive and react to dynamic and potentially hazardous environments accurately. The competition structured five tracks to address these tasks, attracting participation from 140 teams representing 93 international institutes, which together made nearly one thousand submissions.

\subsection{Challenge Tracks}
The following five tracks are established in this competition:
\begin{itemize}
    \item Track 1: Robust BEV Detection
    \item Track 2: Robust Map Segmentation
    \item Track 3: Robust Occupancy Prediction
    \item Track 4: Robust Depth Estimation
    \item Track 5: Robust Multi-Modal BEV Detection
\end{itemize}

\subsection{Challenge Phases}
We hosted two phases for this competition, which are:
\begin{itemize}
    \item Phase 1: Preliminary exploration, from January to March.
    \item Phase 2: Final designs and solutions, from April to May.
\end{itemize}

\subsection{Datasets}
To ensure a rigorous evaluation of participating models, the challenge utilized datasets specifically designed to simulate a wide range of real-world perturbations and sensor malfunctions. These datasets comprise synthetic and real-world data that have been meticulously altered to include common disturbances such as adverse weather, sensor noise, and lighting variations. This approach allows us to thoroughly assess the adaptability and robustness of each solution in conditions that mimic real-world unpredictability.

\subsubsection{Camera Corruptions}
A total of 18 corruption types are used in this competition, including \textit{brightness}, \textit{low-light}, \textit{fog}, \textit{frost}, \textit{snow}, \textit{contrast}, \textit{defocus blur}, \textit{glass blur}, \textit{motion blur}, \textit{zoom blur}, \textit{elastic transform}, \textit{quantization}, \textit{Gaussian noise}, \textit{impulse noise}, \textit{shot noise}, \textit{ISO noise}, \textit{pixelate}, and \textit{JPEG compression}. The rationale of these corruptions is elaborated as follows.

\begin{itemize}
    \item \textbf{Brightness and Low-Light Images.} Variations in lightness or darkness can dramatically affect image visibility and depth quality. Brightness levels are influenced by lighting intensity and object reflectivity, while low-light images often suffer from low visibility and increased noise due to underexposure or dark conditions.

    \item \textbf{Atmospheric Conditions.} Fog and frost introduce scattering and coating effects that blur images and distort details, crucial for accurate depth estimation. Fog reduces contrast and clarity by scattering light, whereas frost forms ice on lenses, leading to distorted images.

    \item \textbf{Precipitation.} Snow can obscure details and alter the perceived depth by covering scenes with a layer of snowflakes, leading to errors in object detection and depth estimation.

    \item \textbf{Contrast and Focus Issues.} Contrast defines the clarity of an image based on the luminance difference, which can be severely affected under varying light conditions. Defocus and glass blur occur when images are out of focus or shot through an uneven glass medium, blurring the details needed for precise depth estimation.

    \item \textbf{Motion and Zoom Blur.} These corruptions arise when there is relative motion between the camera and the subject or due to rapid focal length changes, respectively, both distorting the spatial orientation and clarity of the scene.

    \item \textbf{Image Quality Degradations.} Elastic transformations, color quantization, and compression artifacts from formats like JPEG alter the original fidelity of images, which can mislead depth estimation processes. Noise variations such as Gaussian, impulse, shot, and ISO noise introduce random inaccuracies across the image data, complicating the task of reliable depth measurement.

    \item \textbf{Resolution Issues.} Pixelation occurs when images are too low in resolution, making individual pixels visible and degrading the overall image quality.
\end{itemize}

\subsubsection{LiDAR Failures}
In addition to the 18 malfunction scenarios on cameras, we also consider the following 3 sensor failure cases that have certain likelihood to occur on the LiDAR sensors.
\begin{itemize}
    \item \textbf{Points Dropping}, which refers to the phenomenon where certain data points that a LiDAR sensor collects are lost or not recorded. This can occur due to various factors such as sensor malfunctions, adverse environmental conditions, or issues with data transmission. Points dropping leads to incomplete or sparse point clouds, which can significantly affect the accuracy and reliability of 3D representations.
    
    \item \textbf{Angular Range Restriction}, which is defined as restricting the LiDAR to capture points only within a specific angular range degree relative to the ego vehicle's coordinate system. This selective sensing can be due to intentional configuration, aiming to focus on the most relevant data directly in front of the vehicle. However, this results in a significant reduction in surrounding environmental data, potentially overlooking critical obstacles and hazards located outside of this narrow view range, thus affecting navigation and safety systems.
    
    \item \textbf{Beam Dropping}, which occurs when one or more of the LiDAR sensor's beams fail to function properly. This failure can be attributed to hardware issues, such as damaged emitters or detectors, or software faults that mismanage beam activation. Beam dropping results in gaps within the collected data, impacting the uniformity and completeness of the scanned area.

\end{itemize}

\subsection{Evaluation Metrics}
In this competition, to maintain fairness in evaluating the performances across different teams, we adopt several conventional perception metrics. The metrics used for each of the five tracks are defined as follows.

\noindent\textbf{BEV Detection.}
Following the convention, we adopt the nuScenes Detection Score, \ie, \texttt{NDS} for evaluating the performance in the BEV detection tracks (Track 1 and Track 5) of this competition. The \texttt{NDS} score is calculated over the mean Average Precision (AP) and the True Positive (TP) metrics, which can be defined as follows:
\begin{equation}
    \label{eq:map}
    \text{mAP}= \frac{1}{|\mathbb{C}||\mathbb{D}|}\sum_{c\in\mathbb{C}}\sum_{d\in\mathbb{D}}\text{AP}_{c,d}~,
\end{equation}
\begin{equation}
    \label{eq:nds}
    \text{NDS} = \frac{1}{10}~[5\text{mAP}+\sum_{\text{mTP}\in\mathbb{TP}}(1-\min(1, \text{mTP}))] ~,
\end{equation}
where mTP denotes the mean True Positive score. For a detailed definition of the \texttt{NDS} score, please refer to the official nuScenes dataset~\cite{caesar2020nuscenes}.

\noindent\textbf{Map Segmentation.}
For the robust map segmentation track, we adopt the mean Intersection of Union, \ie, \texttt{mIoU}, to evaluate the segmentation model's performance. In line with the standard protocol, the \texttt{IoU} score can be calculated as:
\begin{equation}
\text{IoU}_{i} = \frac{\text{TP}_{i}}{\text{TP}_{i} + \text{FP}_{i} + \text{FN}_{i}}~,
\label{eq:iou}
\end{equation}
where $\text{TP}_{i}, \text{FP}_{i}, \text{FN}_{i}$ are true-positives, false-positives, and false-negatives for class $i$, respectively, on HD maps. \texttt{mIoU} is the average value of \texttt{IoU} scores for all semantic classes.

\noindent\textbf{Semantic Occupancy Prediction.}
Similar to the HD map segmentation task, we use the \texttt{mIoU} metric to evaluate the performances of semantic occupancy prediction models in Track 4. The calculation of \texttt{IoU} scores on voxels is the same as \cref{eq:iou}. \texttt{mIoU} is the average value of \texttt{IoU} scores for all semantic classes in this track.

\noindent\textbf{Depth Estimation.}
For the robust multi-view depth estimation track, we adopt the conventional reporting of \texttt{Abs Rel} (error rate) and $\delta_1$ (accuracy) for measuring the depth estimation performance. \texttt{Abs Rel} measures the absolute relative difference between the ground-truth (\texttt{gt}) and the prediction (\texttt{pred}), which can be calculated as follows:
\begin{equation}
\text{Abs Rel} = \frac{1}{|D|}\sum_{pred\in D}\frac{|gt - pred|}{gt}~.
\end{equation}
The $\delta$ metric, on the other hand, is the depth estimation accuracy given the threshold:
\begin{equation}
\delta_t = \frac{1}{|D|}|\{\ pred\in D | \max{(\frac{gt}{pred}, \frac{pred}{gt})< 1.25^t}\}| \times 100\%~,
\end{equation}
where $\delta_1 = \delta<1.25, \delta_2 = \delta<1.25^2, \delta_3 = \delta<1.25^3$ are the three conventionally used accuracy scores in prior works \cite{eigen2014depth,godard2019monodepth2,watson2019hints}. The \texttt{Abs Rel} metric is adopted as the main indicator during comparisons.

\input{tables/track1}

\subsection{Baseline Models}
In this section, we provide more details on the baseline models used in each of the five tracks in this competition.

\subsubsection{Track 1}
This track challenges participants to employ advanced learning algorithms for BEV detection. The task encompasses the identification and localization of dynamic objects such as vehicles and pedestrians, as well as the handling of complex environmental elements like road signs and static obstacles. Accuracy and sensitivity to detail are paramount in this track.
\begin{itemize}
    \item \textbf{Baseline Model:} BEVFormer \cite{li2022bevformer}.
    \item \textbf{Estimated Training Requirements:} The model requires 8x NVIDIA GeForce RTX 3090 GPUs and typically completes its training cycle within 1 to 2 days.
    \item \textbf{Baseline Performance:} Our baseline model achieved an NDS of $31.24\%$ and a mAP of $18.82\%$ on our robustness evaluation set.
    \item \textbf{Evaluation Server:} All the submissions were evaluated on our Track 1 server at the CodaLab platform, accessible at \url{https://codalab.lisn.upsaclay.fr/competitions/17135}.
\end{itemize}

\subsubsection{Track 2}
This track challenges participants to use advanced learning algorithms for precise map segmentation on high-resolution BEV images. The task includes a detailed analysis of various urban geographical features such as lane markings, sidewalks, and green spaces. Additionally, this track tests the participants' ability to handle image segmentation under varying lighting, weather conditions, and noise levels.
\begin{itemize}
    \item \textbf{Baseline Model:} BEVerse \cite{zhang2022beverse}.
    \item \textbf{Estimated Training Requirements:} The model requires 8x NVIDIA GeForce RTX 3090 GPUs and typically completes its training cycle within 1 to 2 days.
    \item \textbf{Baseline Performance:} Our baseline model achieved a mIoU of $17.33\%$ on our robustness evaluation set.
    \item \textbf{Evaluation Server:} All the submissions were evaluated on our Track 2 server at the CodaLab platform, accessible at \url{https://codalab.lisn.upsaclay.fr/competitions/17062}.
\end{itemize}

\input{tables/track2}
\input{tables/track3}
\input{tables/track4}

\subsubsection{Track 3}
This track requires participants to use advanced learning algorithms to predict the occupancy and corresponding semantic information of urban roads and their surroundings. The task involves accurately forecasting traffic flow, pedestrian dynamics, and static obstacles, which are crucial for the real-time response and decision-making capabilities of autonomous driving systems.
\begin{itemize}
    \item \textbf{Baseline Model:} SurroundOcc \cite{wei2023surroundocc}.
    \item \textbf{Estimated Training Requirements:} The model requires 8x NVIDIA GeForce RTX 3090 GPUs and typically completes its training cycle within 2 to 3 days.
    \item \textbf{Baseline Performance:} Our baseline model achieved a mIoU of $11.30\%$ on our robustness evaluation set.
    \item \textbf{Evaluation Server:} The submissions were evaluated on our Track 3 server at the CodaLab platform, accessible at \url{https://codalab.lisn.upsaclay.fr/competitions/17063}.
\end{itemize}

\subsubsection{Track 4}
This track challenges participants to use advanced learning algorithms for depth estimation from data captured by cameras at multiple viewpoints. Unlike traditional monocular or binocular depth estimation, this task focuses on leveraging surround camera data to construct a comprehensive and systematic 3D environmental depth estimation model. The task demands that algorithms effectively handle variations in viewpoint, lighting changes, and other environmental factors that impact depth estimation.
\begin{itemize}
    \item \textbf{Baseline Model:} SurroundDepth \cite{wei2023surrounddepth}.
    \item \textbf{Estimated Training Requirements:} The model requires 8x NVIDIA GeForce RTX 3090 GPUs and typically completes its training cycle within 2 to 3 days.
    \item \textbf{Baseline Performance:} Our baseline model achieved an Abs Rel of $0.348$, an RMSE of $7.102$, and an a1 score of $62.3\%$ on our robustness evaluation set.
    \item \textbf{Evaluation Server:} The submissions were evaluated on our Track 4 server at the CodaLab platform, accessible at \url{https://codalab.lisn.upsaclay.fr/competitions/17226}.
\end{itemize}

\subsubsection{Track 5}

This track requires participants to use advanced learning algorithms to fuse data from different sensors, such as cameras and LiDAR, to enhance the accuracy of Bird's Eye View (BEV) 3D object detection. This track encourages participants to demonstrate their innovative capabilities in multi-modal data fusion and processing, while ensuring the robustness and efficiency of their algorithms.
\begin{itemize}
    \item \textbf{Baseline Model:} BEVFusion \cite{liu2023bevfusion}.
    \item \textbf{Estimated Training Requirements:} The model requires 8x NVIDIA GeForce RTX 3090 GPUs and typically completes its training cycle within 1 to 2 days.
    \item \textbf{Baseline Performance:} Our baseline model achieved an NDS of $42.86\%$ and a mAP of $24.50\%$ on our robustness evaluation set.
    \item \textbf{Evaluation Server:} The submissions were evaluated on our Track 5 server at the CodaLab platform, accessible at \url{https://codalab.lisn.upsaclay.fr/competitions/17137}.
\end{itemize}

\input{tables/track5}

\subsection{Challenge Results}
This competition showcased a remarkable advancement in the robustness and effectiveness of perception models, particularly when benchmarked against pre-established baselines. Analysis of the solutions submitted revealed several innovative strategies that significantly contributed to enhancing model robustness and performance.

\noindent\textbf{Leaderboards.}
We observe a substantial enhancement in the robustness of the different driving perception models when compared to the established baselines. As shown in \cref{tab:res-track1}, the performance of the baseline BEVFormer \cite{li2022bevformer} has been largely improved from $22.8$ NDS to $52.1$ NDS by Team \textcolor{robo_blue}{DeepVision}. This corresponds to a $56\%$ relative improvement. The other two winning teams, \ie, \textcolor{robo_blue}{Ponyville Autonauts Ltd.} and \textcolor{robo_blue}{CyberBEV}, achieved $50.2$ NDS and $49.0$ NDS, respectively. As shown in \cref{tab:res-track2} and \cref{tab:res-track3}, similar trends also occur in Track 2 and Track 3, where Team \textcolor{robo_blue}{SafeDrive-SSR} and Team \textcolor{robo_blue}{ViewFormer} have exhibited clear advantages over the baseline models and other participants in the leaderboards (with a $33.1\%$ mIoU and a $13.6\%$ mIoU higher than the baseline BEVerse \cite{zhang2022beverse} and SurroundOcc \cite{wei2023surroundocc}). The results from Tab.~\ref{tab:res-track4} highlight Team \textcolor{robo_blue}{HIT-AIIA}'s leading performance in depth estimation under adverse conditions, as observed by their low error rates in scenarios like low-light and frost, which exhibited a notable improvement over the baseline SurroundDepth \cite{wei2023surrounddepth}. In Tab.~\ref{tab:res-track5}, Team \textcolor{robo_blue}{Safedrive} excelled in the robust multi-modal BEV detection track, achieving the highest NDS and mAP scores ($10.6$ NDS and $17.9$ mAP higher than the baseline BEVFusion \cite{liu2023bevfusion}). This underscores their effective integration of multi-modal data to handle diverse environmental challenges. These findings demonstrate considerable progress in enhancing the robustness of models against a broad spectrum of real-world corruptions.

\noindent\textbf{Discussions \& Analyses.}
Throughout the solutions analyzed, several techniques have emerged as particularly effective in augmenting model's robustness under OoD scenarios:
\begin{itemize}
    \item \textit{Data Augmentation}. Data augmentation represents the most prevalent strategy employed across all teams. Notably, powerful augmentation techniques such as Augmix~\cite{hendrycks2019augmix}, AugFFT, and DeepAug, among others, have been widely implemented. It is important to note that to ensure fair comparions, the use of corruptions similar to those employed in generating the dataset is explicitly prohibited as a means of data augmentation during model training. Nevertheless, the integration of robust and varied data augmentation methods continues to be an effective approach for enhancing model resilience.

    \item \textit{Temporal Fusion}. In addition to data augmentation, leveraging temporal fusion has also become a common practice among the teams. Team \textcolor{robo_blue}{SafeDriveSSR} implemented both temporal fusion and temporal-spatial consistency checks, achieving the highest results in the robust map segmentation track and significantly surpassing the second-ranked team. Furthermore, by utilizing temporal fusion, Team \textcolor{robo_blue}{ViewFormer} enhanced their model with streaming temporal attention, which contributed to their first-place victory in the robust occupancy prediction track. This approach underscores the efficacy of incorporating temporal dynamics into model architectures to improve robustness and performance in scenarios where data evolves over time.

    \item \textit{Robust Backbone}. Empowering the model with a strong backbone architecture is another strategy implemented by top-ranked solutions. The adoption of powerful backbones such as EVA ViT~\cite{fang2023eva}, Swin Transformer~\cite{liu2021swin}, and DINOv2~\cite{oquab2023dinov2} illustrates this trend. It is particularly interesting to observe how participants adapt these advanced vision backbones to specific driving perception tasks. The integration of such robust architectures not only enhances the model's capability to handle complex visual data but also improves its generalization across different environmental conditions and scenarios. This adaptation is pivotal for advancing the state-of-the-art in driving perception.
\end{itemize}
These insights demonstrate substantial progress in enhancing the robustness of models against a wide spectrum of real-world conditions and corruptions, setting a new benchmark in the field of robust autonomous driving perception.

%% file: tables/track1.tex
\begin{table*}[t]
\centering
\caption{The challenge results (Phase 2) of \textbf{Track 1: Robust BEV Detection} in the 2024 RoboDrive Challenge. All scores are given in percentage ($\%$). The higher the scores, the better the overall robustness. The best score for each corruption scenario is highlighted in \textbf{bold}.}
\resizebox{\textwidth}{!}{%
\begin{tabular}{l|c|cccccccccccccccccc}
\toprule
\textbf{Team Name} & \rotatebox{90}{\textcolor{robo_blue}{$\bullet$} \textbf{NDS}} & \rotatebox{90}{\textcolor{robo_blue}{$\circ$} Brightness~} & \rotatebox{90}{\textcolor{robo_blue}{$\circ$} Low-Light~} & \rotatebox{90}{\textcolor{robo_blue}{$\circ$} Fog} & \rotatebox{90}{\textcolor{robo_blue}{$\circ$} Frost} & \rotatebox{90}{\textcolor{robo_blue}{$\circ$} Snow} & \rotatebox{90}{\textcolor{robo_blue}{$\circ$} Contrast} & \rotatebox{90}{\textcolor{robo_blue}{$\circ$} Defocus Blur~} & \rotatebox{90}{\textcolor{robo_blue}{$\circ$} Glass Blur} & \rotatebox{90}{\textcolor{robo_blue}{$\circ$} Motion Blur~} & \rotatebox{90}{\textcolor{robo_blue}{$\circ$} Zoom Blur} & \rotatebox{90}{\textcolor{robo_blue}{$\circ$} Elastic Transform~} & \rotatebox{90}{\textcolor{robo_blue}{$\circ$} Quantization} & \rotatebox{90}{\textcolor{robo_blue}{$\circ$} Gaussian Noise~} & \rotatebox{90}{\textcolor{robo_blue}{$\circ$} Impulse Noise} & \rotatebox{90}{\textcolor{robo_blue}{$\circ$} Shot Noise} & \rotatebox{90}{\textcolor{robo_blue}{$\circ$} ISO Noise} & \rotatebox{90}{\textcolor{robo_blue}{$\circ$} Pixelate} & \rotatebox{90}{\textcolor{robo_blue}{$\circ$} JPEG Compression~} 
\\\midrule\midrule
\rowcolor{robo_blue!50}\multicolumn{20}{l}{\textcolor{white}{\textbf{Winning Teams}}}
\\
\rowcolor{robo_blue!18} \textcolor{robo_blue}{\faTrophy~\textbf{DeepVision}} & $\mathbf{52.1}$ & $39.5$ & $\mathbf{65.3}$ & $36.7$ & $\mathbf{49.8}$ & $\mathbf{62.3}$ & $\mathbf{51.8}$ & $53.5$ & $\mathbf{49.1}$ & $41.8$ & $\mathbf{37.1}$ & $\mathbf{45.3}$ & $\mathbf{67.5}$ & $\mathbf{71.2}$ & $\mathbf{52.9}$ & $\mathbf{59.6}$ & $\mathbf{56.7}$ & $\mathbf{56.9}$ & $40.8$
\\ 
\rowcolor{robo_blue!10} \faTrophy~Ponyville & $50.2$ & $\mathbf{43.1}$ & $62.7$ & $\mathbf{37.5}$ & $46.6$ & $60.9$ & $49.2$ & $\mathbf{58.4}$ & $46.5$ & $\mathbf{44.7}$ & $18.8$ & $44.8$ & $66.7$ & $70.6$ & $42.4$ & $56.0$ & $50.8$ & $56.8$ & $\mathbf{47.5}$
\\ 
\rowcolor{robo_blue!5} \faTrophy~CyberBEV & $49.0$ & $42.1$ & $61.4$ & $37.0$ & $46.3$ & $60.5$ & $47.0$ & $57.1$ & $44.9$ & $43.5$ & $17.1$ & $43.8$ & $65.9$ & $69.1$ & $40.9$ & $56.1$ & $49.4$ & $55.3$ & $45.1$
\\\midrule
\rowcolor{robo_red!10}\multicolumn{20}{l}{\textcolor{robo_red}{\textbf{Other Teams}}}
\\
Safedrive-promax  & $48.1$ & $39.2$ & $59.8$ & $37.4$ & $39.8$ & $62.5$ & $47.6$ & $59.0$ & $43.9$ & $41.4$ & $14.3$ & $45.4$ & $63.3$ & $68.5$ & $42.3$ & $53.3$ & $49.5$ & $56.2$ & $42.2$
\\
drivingClass  &  $47.8$ & $39.1$ & $60.0$ & $28.1$ & $48.7$ & $56.3$ & $44.1$ & $52.5$ & $46.9$ & $38.9$ & $34.9$ & $44.3$ & $62.9$ & $63.8$ & $50.9$ & $51.6$ & $52.0$ & $56.3$ & $33.4$
\\
BUPTMM & $43.5$ & $37.7$ & $55.1$ & $30.8$ & $41.1$ & $51.2$ & $45.2$ & $45.1$ & $41.1$ & $38.6$ & $29.6$ & $40.5$ & $56.1$ & $58.7$ & $41.0$ & $48.5$ & $42.4$ & $46.8$ & $34.0$
\\
LDC & $40.7$ & $31.7$ & $54.8$ & $29.5$ & $36.1$ & $47.7$ & $37.8$ & $46.6$ & $36.2$ & $33.5$ & $25.2$ & $35.4$ & $55.7$ & $58.6$ & $40.6$ & $46.9$ & $40.4$ & $43.1$ & $33.3$
\\
googa & $40.1$ & $34.5$ & $46.5$ & $27.5$ & $30.8$ & $54.4$ & $41.3$ & $46.3$ & $40.1$ & $32.5$ & $11.5$ & $44.6$ & $49.3$ & $49.0$ & $35.7$ & $47.0$ & $45.5$ & $50.0$ & $35.4$ 
\\
Maodouu & $37.8$ & $27.6$ & $47.7$ & $28.7$ & $34.4$ & $46.4$ & $34.5$ & $37.4$ & $36.1$ & $32.7$ & $25.5$ & $30.1$ & $51.9$ & $56.4$ & $36.3$ & $44.9$ & $38.2$ & $41.1$ & $31.1$ 
\\\midrule
\rowcolor{gray!10}\multicolumn{20}{l}{\textcolor{gray}{\textbf{Baseline}}}
\\
BEVFormer & $22.8$ & $28.5$ & $34.9$ & $21.4$ & $10.5$ & $28.0$ & $15.7$ & $28.7$ & $21.1$ & $19.2$ & $6.4$ & $35.0$ & $26.9$ & $20.6$ & $12.3$ & $24.9$ & $25.4$ & $30.3$ & $21.3$
\\ 
\bottomrule
\end{tabular}%
}
\label{tab:res-track1}
\end{table*}

%% file: tables/track2.tex
\begin{table*}[t]
\centering
\caption{The challenge results (Phase 2) of \textbf{Track 2: Robust Map Segmentation} in the 2024 RoboDrive Challenge. All scores are given in percentage ($\%$). The higher the scores, the better the overall robustness. The best score for each corruption scenario is highlighted in \textbf{bold}.}
\resizebox{\textwidth}{!}{%
\begin{tabular}{l|c|cccccccccccccccccc}
\toprule
\textbf{Team Name} & \rotatebox{90}{\textcolor{robo_blue}{$\bullet$} \textbf{mIoU}} & \rotatebox{90}{\textcolor{robo_blue}{$\circ$} Brightness~} & \rotatebox{90}{\textcolor{robo_blue}{$\circ$} Low-Light~} & \rotatebox{90}{\textcolor{robo_blue}{$\circ$} Fog} & \rotatebox{90}{\textcolor{robo_blue}{$\circ$} Frost} & \rotatebox{90}{\textcolor{robo_blue}{$\circ$} Snow} & \rotatebox{90}{\textcolor{robo_blue}{$\circ$} Contrast} & \rotatebox{90}{\textcolor{robo_blue}{$\circ$} Defocus Blur~} & \rotatebox{90}{\textcolor{robo_blue}{$\circ$} Glass Blur} & \rotatebox{90}{\textcolor{robo_blue}{$\circ$} Motion Blur~} & \rotatebox{90}{\textcolor{robo_blue}{$\circ$} Zoom Blur} & \rotatebox{90}{\textcolor{robo_blue}{$\circ$} Elastic Transform~} & \rotatebox{90}{\textcolor{robo_blue}{$\circ$} Quantization} & \rotatebox{90}{\textcolor{robo_blue}{$\circ$} Gaussian Noise~} & \rotatebox{90}{\textcolor{robo_blue}{$\circ$} Impulse Noise} & \rotatebox{90}{\textcolor{robo_blue}{$\circ$} Shot Noise} & \rotatebox{90}{\textcolor{robo_blue}{$\circ$} ISO Noise} & \rotatebox{90}{\textcolor{robo_blue}{$\circ$} Pixelate} & \rotatebox{90}{\textcolor{robo_blue}{$\circ$} JPEG Compression~}
\\\midrule\midrule
\rowcolor{robo_blue!50}\multicolumn{20}{l}{\textcolor{white}{\textbf{Winning Teams}}}
\\
\rowcolor{robo_blue!18} \textcolor{robo_blue}{\faTrophy~\textbf{SafeDriveSSR}} & $\mathbf{48.8}$ & $\mathbf{54.6}$ & $\mathbf{71.1}$ & $28.6$ & $23.1$ & $\mathbf{54.5}$ & $\mathbf{54.6}$ & $\mathbf{64.8}$ & $\mathbf{51.2}$ & $\mathbf{44.7}$ & $21.8$ & $52.1$ & $\mathbf{40.4}$ & $\mathbf{58.5}$ & $\mathbf{46.1}$ & $\mathbf{37.2}$ & $\mathbf{64.2}$ & $55.2$ & $\mathbf{54.9}$
\\ 
\rowcolor{robo_blue!15} \faTrophy~CrazyFriday & $34.5$ & $42.7$ & $40.4$ & $29.8$ & $\mathbf{25.0}$ & $41.7$ & $17.1$ & $42.4$ & $34.8$ & $33.2$ & $\mathbf{22.4}$ & $48.0$ & $31.5$ & $33.7$ & $23.6$ & $26.0$ & $38.0$ & $45.8$ & $45.6$ 
\\ 
\rowcolor{robo_blue!5} \faTrophy~Samsung & $29.8$ & $42.4$ & $28.7$ & $\mathbf{30.4}$ & $9.1$ & $8.6$ & $13.8$ & $21.6$ & $43.1$ & $24.8$ & $19.6$ & $\mathbf{75.2}$ & $18.7$ & $23.5$ & 14.3 & $17.7$ & $31.4$ & $\mathbf{68.2}$ & $44.5$
\\\midrule
\rowcolor{robo_red!10}\multicolumn{20}{l}{\textcolor{robo_red}{\textbf{Other Teams}}}
\\
hm.unilab & $28.2$ & $29.6$ & $30.2$ & $25.8$ & $11.1$ & $21.1$ & $7.3$ & $23.8$ & $30.0$ & $21.0$ & $17.0$ & $45.7$ & $14.5$ & $40.4$ & $36.2$ & $27.6$ & $47.0$ & $45.4$ & $33.0$
\\ 
hyt1407 & $26.4$ & $18.6$ & $37.6$ & $23.5$ & $10.4$ & $22.8$ & $16.8$ & $40.1$ & $30.2$ & $30.1$ & $22.8$ & $42.9$ & $20.2$ & $23.2$ & $13.5$ & $15.2$ & $28.8$ & $40.3$ & $38.6$ 
\\ 
Oliguy & $26.4$ & $18.9$ & $36.8$ & $23.6$ & $10.9$ & $23.6$ & $16.7$ & $39.2$ & $30.4$ & $29.6$ & $22.6$ & $42.8$ & $19.4$ & $23.7$ & $13.5$ & $15.6$ & $28.4$ & $40.7$ & $38.4$ 
\\
Yu\_Tian\_1995 & $25.9$ & $33.1$ & $34.6$ & $30.2$ & $10.8$ & $19.7$ & $14.6$ & $39.2$ & $31.2$ & $23.8$ & $19.6$ & $45.8$ & $19.1$ & $19.1$ & $13.4$ & $11.8$ & $28.6$ & $40.5$ & $30.9 $
\\\midrule
\rowcolor{gray!10}\multicolumn{20}{l}{\textcolor{gray}{\textbf{Baseline}}}
\\
BEVerse & $15.7$ & $21.4$ & $14.1$ & $19.2$ & $6.8$ & $3.2$ & $3.7$ & $18.9$ & $27.9$ & $9.2$ & $17.6$ & $44.6$ & $11.4$ & $5.2$ & $2.1$ & $2.4$ & $9.6$ & $36.8$ & $28.0$ 
\\ \bottomrule
\end{tabular}%
}
\label{tab:res-track2}
\end{table*}


%% file: tables/track3.tex
\begin{table*}[t]
\centering
\caption{The challenge results (Phase 2) of \textbf{Track 3: Robust Occupancy Prediction} in the 2024 RoboDrive Challenge. All scores are given in percentage ($\%$). The higher the scores, the better the overall robustness. The best score for each corruption scenario is highlighted in \textbf{bold}.}
\resizebox{\textwidth}{!}{%
\begin{tabular}{l|c|cccccccccccccccccc}
\toprule
\textbf{Team Name} & \rotatebox{90}{\textcolor{robo_blue}{$\bullet$} \textbf{mIoU}} & \rotatebox{90}{\textcolor{robo_blue}{$\circ$} Brightness~} & \rotatebox{90}{\textcolor{robo_blue}{$\circ$} Low-Light~} & \rotatebox{90}{\textcolor{robo_blue}{$\circ$} Fog} & \rotatebox{90}{\textcolor{robo_blue}{$\circ$} Frost} & \rotatebox{90}{\textcolor{robo_blue}{$\circ$} Snow} & \rotatebox{90}{\textcolor{robo_blue}{$\circ$} Contrast} & \rotatebox{90}{\textcolor{robo_blue}{$\circ$} Defocus Blur~} & \rotatebox{90}{\textcolor{robo_blue}{$\circ$} Glass Blur} & \rotatebox{90}{\textcolor{robo_blue}{$\circ$} Motion Blur~} & \rotatebox{90}{\textcolor{robo_blue}{$\circ$} Zoom Blur} & \rotatebox{90}{\textcolor{robo_blue}{$\circ$} Elastic Transform~} & \rotatebox{90}{\textcolor{robo_blue}{$\circ$} Quantization} & \rotatebox{90}{\textcolor{robo_blue}{$\circ$} Gaussian Noise~} & \rotatebox{90}{\textcolor{robo_blue}{$\circ$} Impulse Noise} & \rotatebox{90}{\textcolor{robo_blue}{$\circ$} Shot Noise} & \rotatebox{90}{\textcolor{robo_blue}{$\circ$} ISO Noise} & \rotatebox{90}{\textcolor{robo_blue}{$\circ$} Pixelate} & \rotatebox{90}{\textcolor{robo_blue}{$\circ$} JPEG Compression~}
\\ \midrule\midrule
\rowcolor{robo_blue!50}\multicolumn{20}{l}{\textcolor{white}{\textbf{Winning Teams}}}
\\
\rowcolor{robo_blue!18} \textcolor{robo_blue}{\faTrophy~\textbf{ViewFormer}} & $\mathbf{22.3}$ & $\mathbf{21.4}$ & $\mathbf{29.3}$ & $\mathbf{16.8}$ & $\mathbf{15.8}$ & $\mathbf{31.0}$ & $\mathbf{17.9}$ & $\mathbf{27.7}$ & $\mathbf{20.7}$ & $\mathbf{18.9}$ & $6.6$ & $\mathbf{27.3}$ & $\mathbf{20.5}$ & $\mathbf{25.4}$ & $\mathbf{19.8}$ & $\mathbf{22.5}$ & $\mathbf{25.8}$ & $\mathbf{30.0}$ & $\mathbf{19.8}$ 
\\ 
\rowcolor{robo_blue!10} \faTrophy~APEC Blue & $10.3$ & $13.3$ & $12.2$ & $8.7$ & $9.7$ & $7.8$ & $5.2$ & $13.0$ & $14.0$ & $4.3$ & $\textbf{7.7}$ & $15.3$ & $18.2$ & $8.8$ & $9.2$ & $10.8$ & $11.7$ & $15.3$ & $10.8$
\\ 
\rowcolor{robo_blue!5} \faTrophy~hm.unilab & $8.9$ & $13.1$ & $12.9$ & $7.9$ & $3.1$ & $7.8$ & $4.5$ & $11.8$ & $7.1$ & $4.3$ & $4.9$ & $14.6$ & $6.6$ & $9.0$ & $7.9$ & $10.0$ & $10.9$ & $15.0$ & $9.5$ 
\\\midrule
\rowcolor{robo_red!10}\multicolumn{20}{l}{\textcolor{robo_red}{\textbf{Other Teams}}}
\\
LVGroupHFUT & $8.8$ & $13.7$ & $13.0$ & $6.9$ & $3.2$ & $6.6$ & $4.0$ & $10.9$ & $7.2$ & $4.4$ & $4.9$ & $14.3$ & $6.6$ & $9.0$ & $8.3$ & $10.2$ & $11.2$ & $14.9$ & $9.2$
\\\midrule
\rowcolor{gray!10}\multicolumn{20}{l}{\textcolor{gray}{\textbf{Baseline}}}
\\
SurroundOcc & $8.7$ & $12.1$ & $13.1$ & $7.8$ & $3.7$ & $7.5$ & $4.4$ & $11.3$ & $6.9$ & $4.3$ & $4.9$ & $14.3$ & $9.5$ & $8.1$ & $7.0$ & $9.2$ & $9.8$ & $14.9$ & $9.7$
\\ 
\bottomrule
\end{tabular}%
}
\label{tab:res-track3}
\end{table*}

%% file: tables/track4.tex
\begin{table*}[t]
\centering
\caption{The challenge results (Phase 2) of \textbf{Track 4: Robust Depth Estimation} in the 2024 RoboDrive Challenge. All scores are given in percentage ($\%$). The lower the scores, the better the overall robustness. The best score for each corruption scenario is highlighted in \textbf{bold}.}
\resizebox{\textwidth}{!}{%
\begin{tabular}{l|c|cccccccccccccccccc}
\toprule
\textbf{Team Name} & \rotatebox{90}{\textcolor{robo_blue}{$\bullet$} \textbf{Abs Rel}} & \rotatebox{90}{\textcolor{robo_blue}{$\circ$} Brightness~} & \rotatebox{90}{\textcolor{robo_blue}{$\circ$} Low-Light~} & \rotatebox{90}{\textcolor{robo_blue}{$\circ$} Fog} & \rotatebox{90}{\textcolor{robo_blue}{$\circ$} Frost} & \rotatebox{90}{\textcolor{robo_blue}{$\circ$} Snow} & \rotatebox{90}{\textcolor{robo_blue}{$\circ$} Contrast} & \rotatebox{90}{\textcolor{robo_blue}{$\circ$} Defocus Blur~} & \rotatebox{90}{\textcolor{robo_blue}{$\circ$} Glass Blur} & \rotatebox{90}{\textcolor{robo_blue}{$\circ$} Motion Blur~} & \rotatebox{90}{\textcolor{robo_blue}{$\circ$} Zoom Blur} & \rotatebox{90}{\textcolor{robo_blue}{$\circ$} Elastic Transform~} & \rotatebox{90}{\textcolor{robo_blue}{$\circ$} Quantization} & \rotatebox{90}{\textcolor{robo_blue}{$\circ$} Gaussian Noise~} & \rotatebox{90}{\textcolor{robo_blue}{$\circ$} Impulse Noise} & \rotatebox{90}{\textcolor{robo_blue}{$\circ$} Shot Noise} & \rotatebox{90}{\textcolor{robo_blue}{$\circ$} ISO Noise} & \rotatebox{90}{\textcolor{robo_blue}{$\circ$} Pixelate} & \rotatebox{90}{\textcolor{robo_blue}{$\circ$} JPEG Compression~}
\\\midrule\midrule
\rowcolor{robo_blue!50}\multicolumn{20}{l}{\textcolor{white}{\textbf{Winning Teams}}}
\\
\rowcolor{robo_blue!18} \textcolor{robo_blue}{\faTrophy~\textbf{HIT-AIIA}} & $\mathbf{18.7}$ & $11.7$ & $\mathbf{12.5}$ & $34.3$ & $\mathbf{24.4}$ & $\mathbf{16.9}$ & $\mathbf{15.2}$ & $\mathbf{15.6}$ & $\mathbf{16.4}$ & $29.6$ & $23.7$ & $\mathbf{20.8}$ & $19.3$ & $16.2$ & $\mathbf{19.5}$ & $\mathbf{12.9}$ & $\mathbf{18.9}$ & $\mathbf{14.2}$ &$\mathbf{14.9}$
\\ 
\rowcolor{robo_blue!10} \faTrophy~BUAA-Trans & $20.5$ & $\mathbf{10.5}$ & $17.6$ & $\mathbf{33.7}$ & $26.8$ & $20.0$ & $25.3$ & $18.8$ & $21.4$ & $\mathbf{26.1}$ & $\mathbf{22.7}$ & $21.5$ & $\mathbf{18.4}$ & $\mathbf{15.3}$ & $21.5$ & $13.6$ & $20.1$ & $17.3$ & $17.6$
\\ 
\rowcolor{robo_blue!5} \faTrophy~CUSTZS & $26.4$ & $16.4$ & $25.7$ & $37.9$ & $30.1$ & $20.4$ & $23.8$ & $23.1$ & $25.6$ & $36.4$ & $36.1$ & $27.5$ & $24.1$ & $26.1$ & $26.6$ & $18.6$ & $33.6$ & $21.5$ & $22.6$ 
\\ \midrule
\rowcolor{robo_red!10}\multicolumn{20}{l}{\textcolor{robo_red}{\textbf{Other Teams}}}
\\
Highway & $29.2$ & $16.8$ & $20.9$ & $43.5$ & $32.9$ & $29.7$ & $32.1$ & $24.9$ & $25.3$ & $38.6$ & $33.4$ & $28.4$ & $30.0$ & $35.7$ & $30.4$ & $22.0$ & $39.8$ & $18.8$ & $21.6$ 
\\\midrule
\rowcolor{gray!10}\multicolumn{20}{l}{\textcolor{gray}{\textbf{Baseline}}}
\\
SurroundDepth & $30.4$ & $16.8$ & $21.0$ & $46.4$ & $33.0$ & $29.7$ & $32.1$ & $25.5$ & $25.3$ & $45.1$ & $43.0$ & $28.4$ & $31.8$ & $35.7$ & $31.0$ & $22.0$ & $39.8$ & $18.8$ & $21.6$
\\ 
\bottomrule
\end{tabular}%
}
\label{tab:res-track4}
\end{table*}

%% file: tables/track5.tex
\begin{table}[t]
\centering
\caption{The challenge results (Phase 2) of \textbf{Track 5: Robust Multi-Modal BEV Detection} in the 2024 RoboDrive Challenge. All scores are given in percentage ($\%$). The NDS and mAP scores are the higher the better, while the other scores are the lower the better. The best score for each corruption scenario is highlighted in \textbf{bold}.}
\resizebox{\linewidth}{!}{%
\begin{tabular}{l|c|c|ccccc}
\toprule
\textbf{Team} & \rotatebox{90}{\textcolor{robo_blue}{$\bullet$} \textbf{NDS}} & \rotatebox{90}{\textcolor{robo_blue}{$\bullet$} \textbf{mAP}} & \rotatebox{90}{\textcolor{robo_blue}{$\circ$} mATE} & \rotatebox{90}{\textcolor{robo_blue}{$\circ$} mASE~} & \rotatebox{90}{\textcolor{robo_blue}{$\circ$} mAOE~} & \rotatebox{90}{\textcolor{robo_blue}{$\circ$} mAVE~} & \rotatebox{90}{\textcolor{robo_blue}{$\circ$} mAAE~}
\\ \midrule\midrule
\rowcolor{robo_blue!50}\multicolumn{8}{l}{\textcolor{white}{\textbf{Winning Teams}}}
\\
\rowcolor{robo_blue!18} \textcolor{robo_blue}{\faTrophy~\textbf{safedrive}} & $\mathbf{49.7}$ & $\mathbf{39.5}$ & $54.0$ & $\mathbf{26.5}$ & $39.6$ & $59.4$ & $\mathbf{21.1}$
\\ 
\rowcolor{robo_blue!10} \faTrophy~Ponyville & $48.2$ & $36.4$ & $56.7$ & $27.2$ & $\mathbf{38.4}$ & $56.2$ & $21.4$
\\ 
\rowcolor{robo_blue!5} \faTrophy~HITSZ & $46.6$ & $32.7$ & $\mathbf{41.4}$ & $28.1$ & $60.0$ & $\mathbf{45.8}$ & $22.6$
\\\midrule
\rowcolor{robo_red!10}\multicolumn{8}{l}{\textcolor{robo_red}{\textbf{Other Teams}}}
\\
UMIC & $42.8$ & $26.4$ & $51.6$ & $27.6$ & $29.4$ & $72.3$ & $23.1$
\\
UESTC & $41.7$ & $24.8$ & $43.5$ & $29.3$ & $63.6$ & $47.8$ & $23.0$ 
\\\midrule
\rowcolor{gray!10}\multicolumn{8}{l}{\textcolor{gray}{\textbf{Baseline}}}
\\
BEVFusion & $39.1$ & $21.6$ & $44.0$ & $30.0$ & $53.0$ & $64.6$ & $25.1$
\\ 
\bottomrule
\end{tabular}%
}
\label{tab:res-track5}
\end{table}

%% file: sections/4_solutions.tex
\section{Challenge Solutions}
\label{sec:solution}
In this section, we summarize the key components of each of the fifteen solutions that contributed to this challenge.

\subsection{Track 1: Robust BEV Detection}
This section introduces the key innovations and implementation details of the three winning solutions in Track 1.
\input{subsections/track1}

\subsection{Track 2: Robust Map Segmentation}
This section introduces the key innovations and implementation details of the three winning solutions in Track 2.
\input{subsections/track2}

\subsection{Track 3: Robust Occupancy Prediction}
This section introduces the key innovations and implementation details of the three winning solutions in Track 3.
\input{subsections/track3}

\subsection{Track 4: Robust Depth Estimation}
This section introduces the key innovations and implementation details of the three winning solutions in Track 4.

\input{subsections/track4}

\subsection{Track 5: Robust Multi-Modal BEV Detection}
This section introduces the key innovations and implementation details of the three winning solutions in Track 5.
\input{subsections/track5}

%% file: subsections/track1.tex
\begin{figure*}[t]
    \centering
    \includegraphics[width=\linewidth]{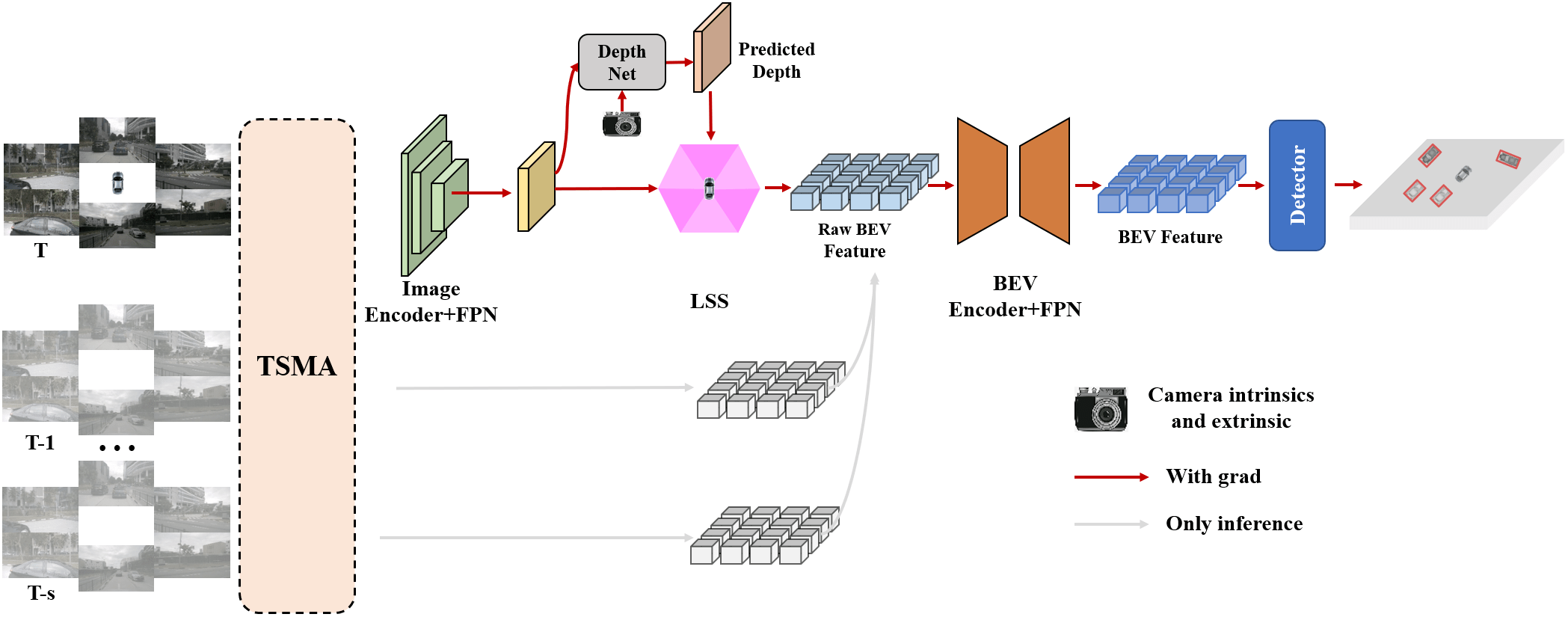}
    \caption{The TSMA-BEV framework: Multi-view images undergo sequence-consistent augmentations, are processed through an image encoder for 2D feature extraction, transformed into 3D space by a view transformer, and finally, detection is performed using concatenated temporal features for enhanced accuracy.}
    \label{fig:track1_deepvision_fig1}
\end{figure*}

\begin{figure}[t]
    \centering
    \includegraphics[width=\linewidth]{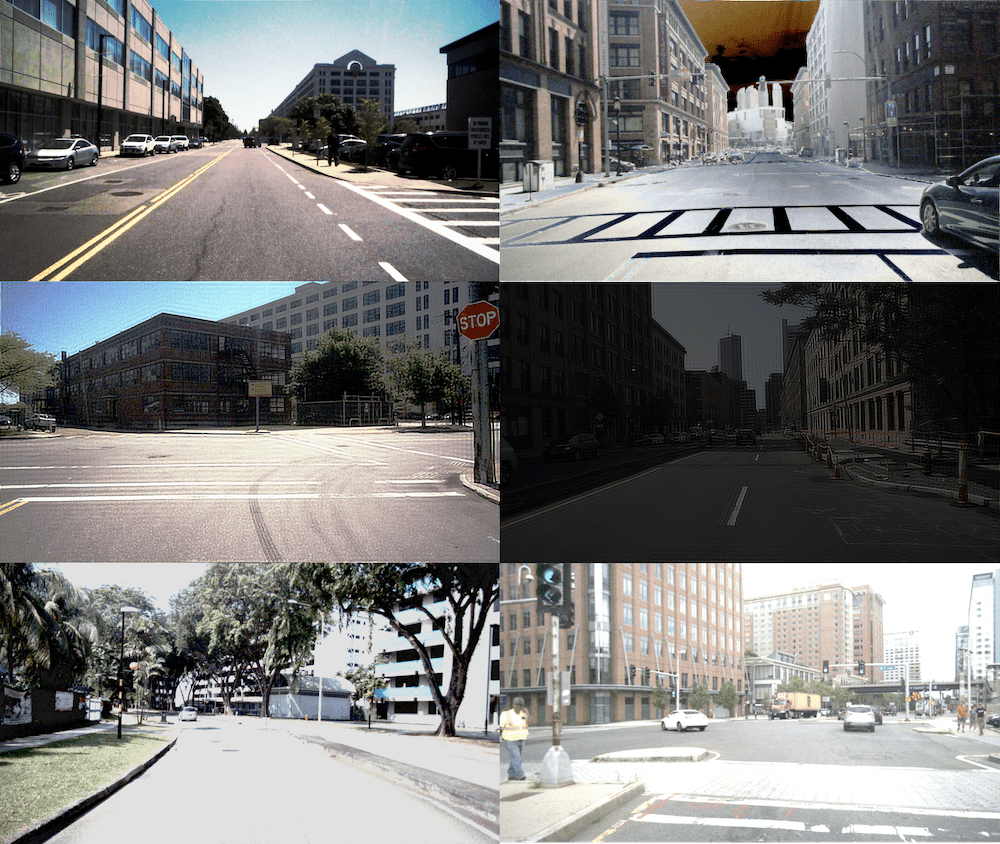}
    \caption{Examples of AugFFT-generated images, demonstrating the variety of frequency domain adjustments used to simulate and prepare for diverse environmental conditions, enhancing the model's generalization capabilities.}
    \label{fig:track1_deepvision_fig4}
\end{figure}

\begin{figure*}[t]
    \centering
    \includegraphics[width=\linewidth]{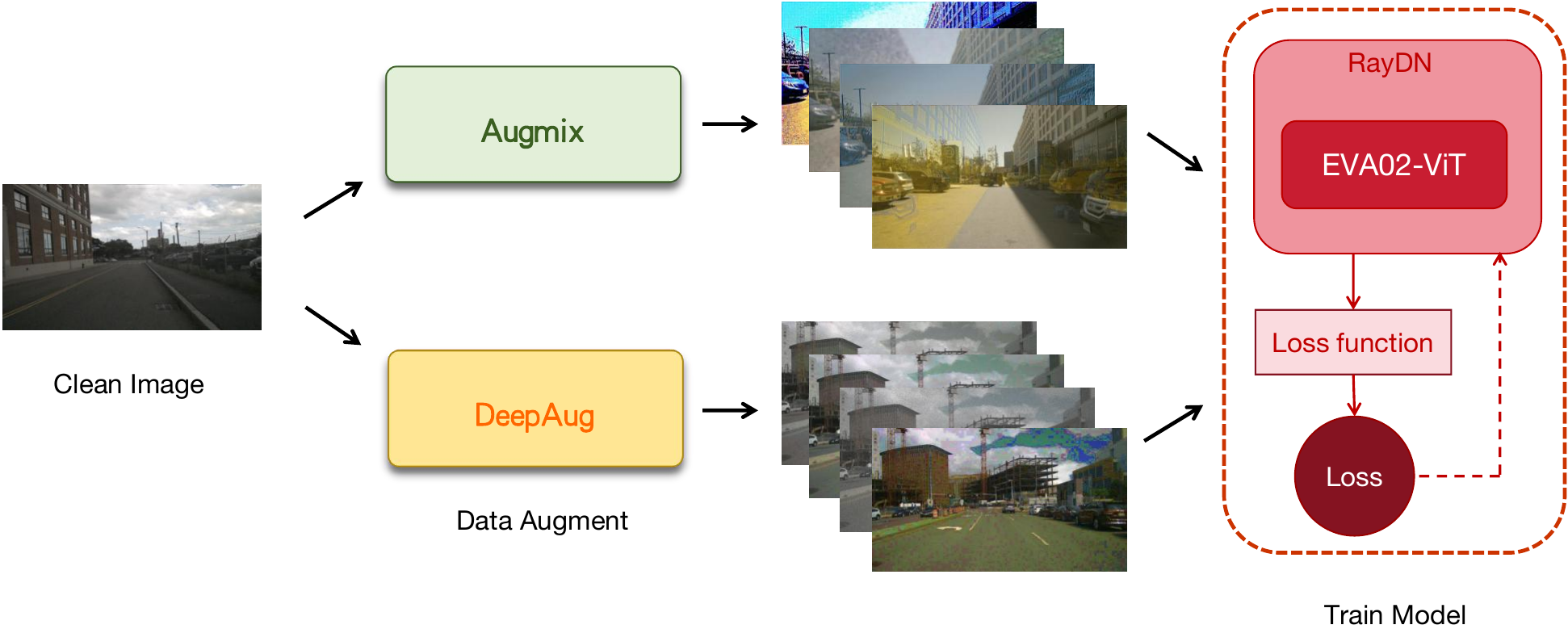}
    \caption{Pipeline of the Multi-View Enhancer (MVE) method developed by Team \textcolor{robo_blue}{Ponyville}, illustrating the integration of advanced computational techniques from data augmentation through to model training and object detection.}
    \label{fig:track1_ponyville_fig1}
\end{figure*}

\begin{figure}[t]
    \centering
    \includegraphics[width=\linewidth]{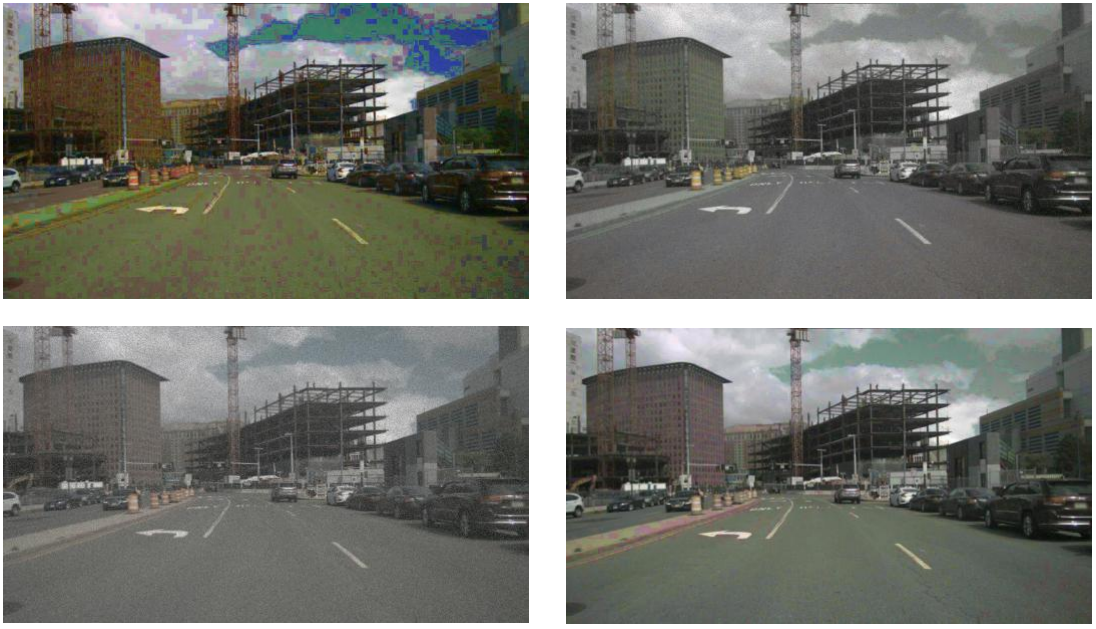}
    \caption{Examples of Augmix-enhanced data~\cite{hendrycks2019augmix} by Team \textcolor{robo_blue}{Ponyville}.}
    \label{fig:track1_ponyville_fig2}
\end{figure}

\begin{figure}[t]
    \centering
    \includegraphics[width=\linewidth]{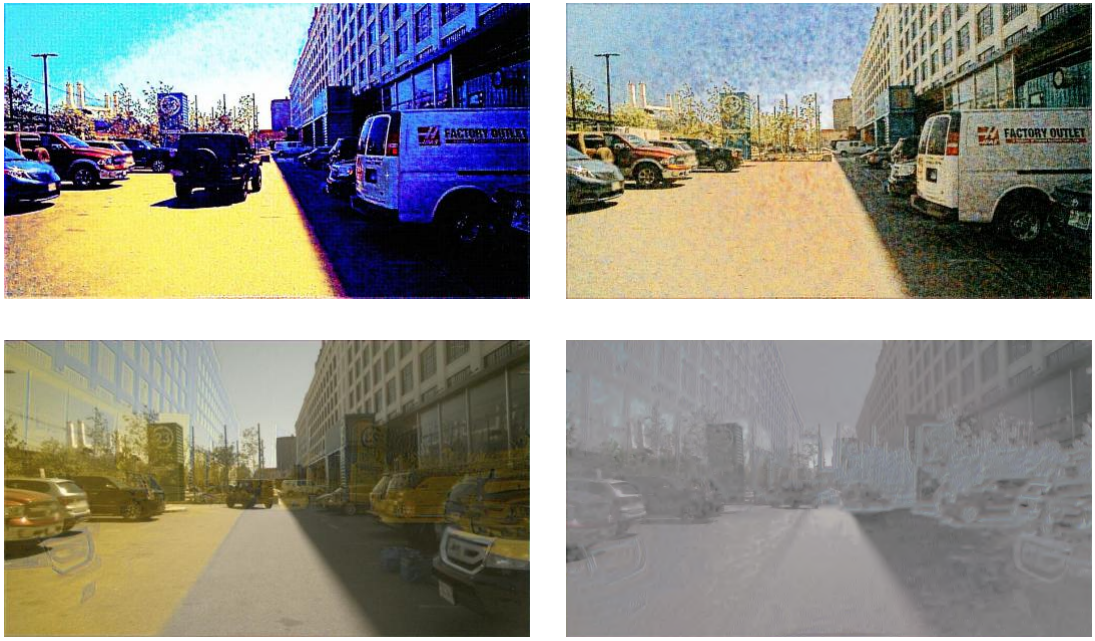}
    \caption{Examples of DeepAug-enhanced data by Team \textcolor{robo_blue}{Ponyville}.}
    \label{fig:track1_ponyville_fig3}
\end{figure}

\begin{figure*}[t]
    \centering
    \includegraphics[width=\linewidth]{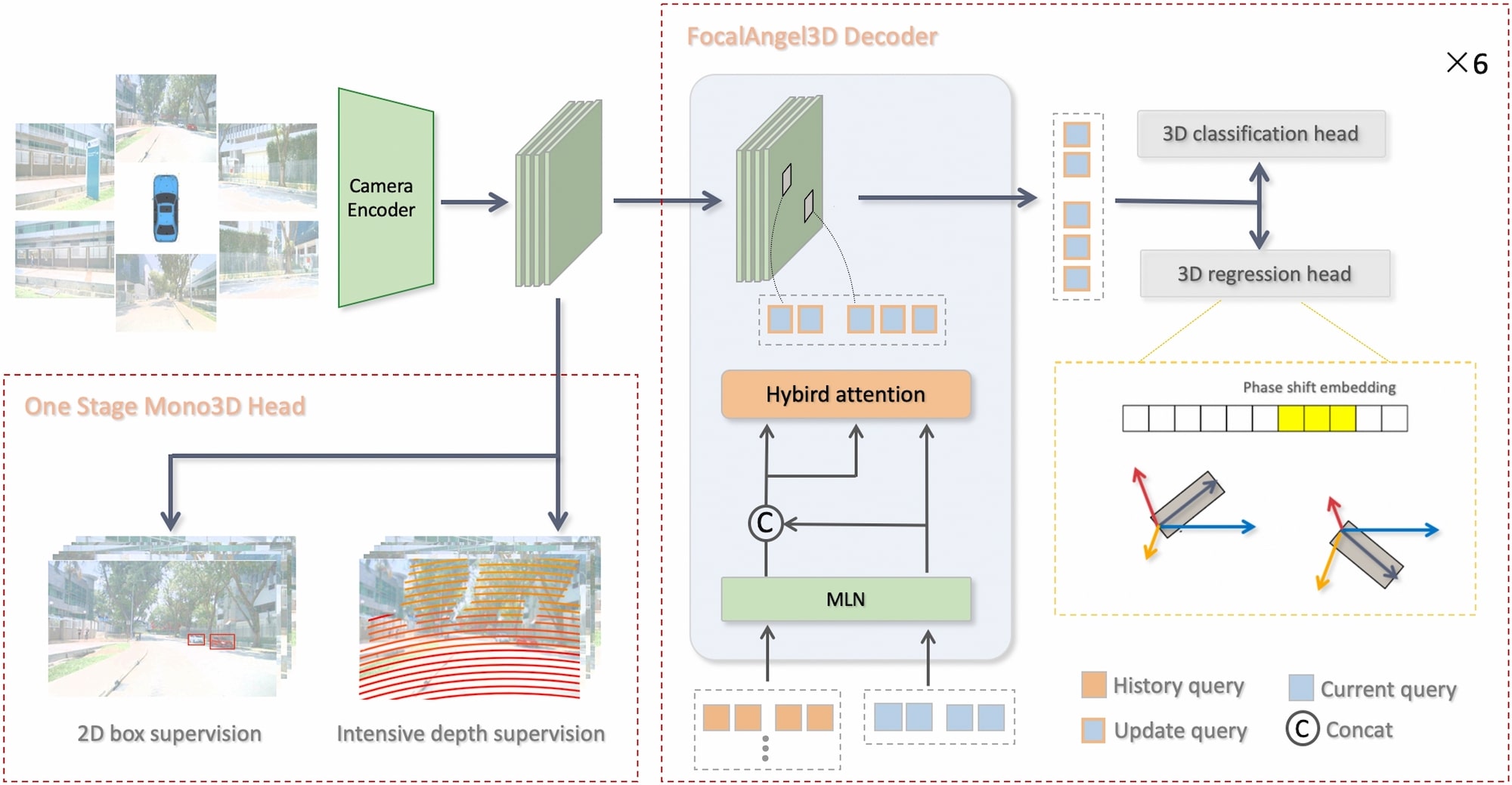}
    \caption{The framework of FocalAngle3D: Surrounding images are input to the image encoder to extract image features, then processed through one stage mono3D prediction head, which includes YOLOX and the intensive depth network. The second stage is the FocalAngle3D decoder, incorporating temporal and spatial fusion modules for enhanced detection accuracy.}
    \label{fig:cyberbev_overview}
\end{figure*}

\begin{figure}[t]
    \centering
    \includegraphics[width=\linewidth]{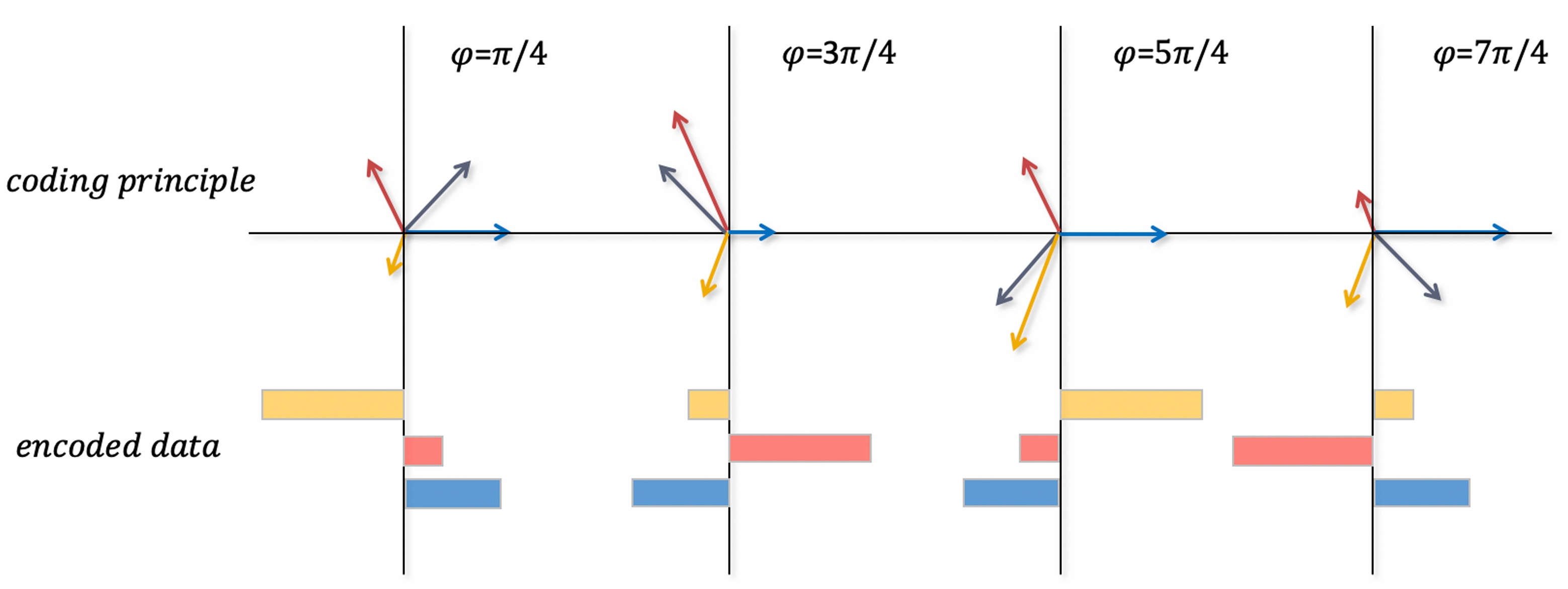}
    \caption{Phase Shift Coder (PSC) implementation: Illustration of the coded principle for four different angles and the corresponding coded values using phase-shift encoding.}
    \label{fig:cyberbev_psc}
\end{figure}

\subsubsection{Team \textcolor{robo_blue}{DeepVision}}
This team has crafted the \textbf{TSMA-BEV} (Temporal Sequence Mix Augmentation - Bird's Eye View) framework to significantly bolster the robustness of camera-only 3D object detection systems. Their innovative approach focuses on overcoming challenges presented by real-world corruptions and sensor inconsistencies through the integration of two novel strategies: Augmentation Fast Fourier Transform (AugFFT) and Sequential Mix Augmentation (SeqMixAug).

\noindent\textbf{\faLightbulbO~Key Innovations:}
\begin{itemize}
    \item \textbf{AugFFT:} This technique involves manipulating the frequency domain of images through stochastic frequency cut-offs and amplitude scaling. This process generates a diverse array of training images, each reflecting different potential environmental disturbances. The aim is to enhance the model's generalization ability across a wide range of conditions, making it robust against variations that it has not been explicitly trained on.
    \item \textbf{SeqMixAug:} This method extends beyond traditional single-frame augmentation by ensuring consistency in augmentation across temporal sequences. It is particularly vital for models that utilize temporal data, helping to maintain learning stability and prevent discrepancies that can arise due to the randomness in augmentation settings across adjacent frames.
\end{itemize}

\noindent\textbf{\faGear~Implementation Details:}\\
The TSMA-BEV framework integrates a novel view transformer to convert 2D image features into a 3D space, utilizing estimated depth from multi-camera inputs to build accurate BEV representations. This setup is crucial for aligning and concatenating volume features from current and historical frames, which enhances the predictive accuracy and robustness of the system in dynamic environments. \cref{fig:track1_deepvision_fig1} provides an overview of the view transformation process. The AugFFT module applies stochastic frequency cut-offs and amplitude scaling to introduce a spectrum of visual variations that simulate potential real-world disturbances. This aspect of the methodology is particularly effective in creating a training dataset that robustly prepares the model for unforeseen environmental changes. The SeqMixAug strategy maintains consistent augmentation parameters across sequences, ensuring that the temporal data used by the model does not introduce discontinuities, thus aiding in the accurate prediction of object trajectories and movements. The training regimen is carefully structured, starting with short-term fusion to ensure quick convergence and followed by the integration of long-term data~\cite{park2022time} and using mixed precision to optimize computational efficiency and depth of training. \cref{fig:track1_deepvision_fig4} highlights the practical effects of their augmentation techniques with augmented image samples.

\subsubsection{Team \textcolor{robo_blue}{Ponyville Autonauts Ltd.}}
This team developed a Multi-View Enhancer (MVE) method to substantially improve the robustness of 3D object detection across multiple camera perspectives. Their approach builds upon the RayDN~\cite{liu2024ray} architecture but introduces significant enhancements with the integration of ImageNet~\cite{recht2019imagenet} pre-trained the EVA ViT-Large backbone~\cite{fang2023eva}. This modified backbone ensures deeper and more robust feature extraction capabilities. The model is further enhanced through a strategic combination of Augmix~\cite{hendrycks2019augmix} and DeepAug data augmentation techniques, meticulously tailored to avoid overlapping with the corruptions encountered in the challenge's test sets.

\noindent\textbf{\faLightbulbO~Key Innovations:}
\begin{itemize}
    \item \textbf{EVA ViT-Large Backbone:} The adoption of a sophisticated Transformer-based model that has been extensively trained on ImageNet \cite{imagenet} for superior feature extraction. This backbone is integral for enhancing the precision of object detection from 2D images converted into a 3D space.
    \item \textbf{Advanced Data Augmentation:} Implementation of Augmix and DeepAug techniques, which are engineered to significantly bolster the model’s resilience against variable environmental conditions. These methods introduce realistic variations that prepare the model to perform reliably under diverse and unforeseen operational conditions.
\end{itemize}

\noindent\textbf{\faGear~Implementation Details:}\\
The overall pipeline of MVE is illustrated in \cref{fig:track1_ponyville_fig1}. It is designed to tackle the inherent challenges of 3D object detection in dynamic driving environments by integrating a novel pipeline for camera-only detection, a sophisticated feature extraction backbone, and innovative data augmentation techniques. Depth-aware hard negative sampling is employed to refine the detection capabilities, enhancing the model's adaptability to varied environmental conditions. The training process is structured to progressively evolve from clean, unaltered datasets to increasingly complex scenarios, effectively building a model that is remarkably resilient. The examples of Augmix and DeepAug data augmentation are shown in \cref{fig:track1_ponyville_fig2} and \cref{fig:track1_ponyville_fig3}, highlighting how these technologies create diverse training data augmentation.

\subsubsection{Team \textcolor{robo_blue}{CyberBEV}}
This team developed the \textbf{FocalAngle3D} framework, enhancing the robustness of 3D object detection through a two-stage model that integrates advanced feature extraction and angle encoding techniques. Building on the StreamPETR~\cite{wang2023exploring} paradigm, their model leverages the 2D bounding box auxiliary supervision provided by RepDETR3D \cite{wang2023exploring}, and incorporates an intensive depth prediction network to enhance feature extraction~\cite{li2023bevdepth}. This contributes to more accurate 3D bounding box regression. For orientation estimation, they employ a novel Phase-Shifting Coder (PSC), which encodes the orientation angle with three phase-shifted encoding channels, which enhances the model’s robustness in capturing subtle angular variations.

\noindent\textbf{\faLightbulbO~Key Innovations:}
\begin{itemize}
    \item \textbf{Intensive Depth Prediction Network:} An innovative addition that integrates precise depth information into the extracted 2D features, significantly improving the positional accuracy of 3D detection boxes.
    \item \textbf{Phase-Shifting Coder (PSC):} A robust encoding method for orientation angle prediction, leveraging additional angle information to better capture subtle angular variations and differences.
\end{itemize}

\noindent\textbf{\faGear~Implementation Details:}\\
The FocalAngle3D framework, as illustrated in \cref{fig:cyberbev_overview}, is specifically tailored to address robust detection in challenging scenarios, such as low visibility and adverse weather conditions. The model starts with the StreamPETR architecture, known for its efficient spatiotemporal modeling capabilities, and introduces deformable attention mechanisms~\cite{zhu2020deformable} to optimize feature extraction across spatial domains. The depth prediction network employs a dense prediction strategy that extends across the detection plane, allowing for a richer and more detailed depth understanding. This setup is complemented by the PSC, which adds dimensionality to the encoding of orientation angles, surpassing traditional sine-cosine methods in capturing minute angular variations. The encoding technique of the PSC is depicted in \cref{fig:cyberbev_psc}, showcasing how angle information is enhanced. The overall pipeline integrates these components into a cohesive system, beginning with image input through the intensive depth prediction network and the YOLOX-based 2D detection network~\cite{ge2021yolox}, flowing into the FocalAngle3D decoder where temporal and spatial fusion takes place.

%% file: subsections/track2.tex
\begin{figure*}[t]
    \centering
    \includegraphics[width=\linewidth]{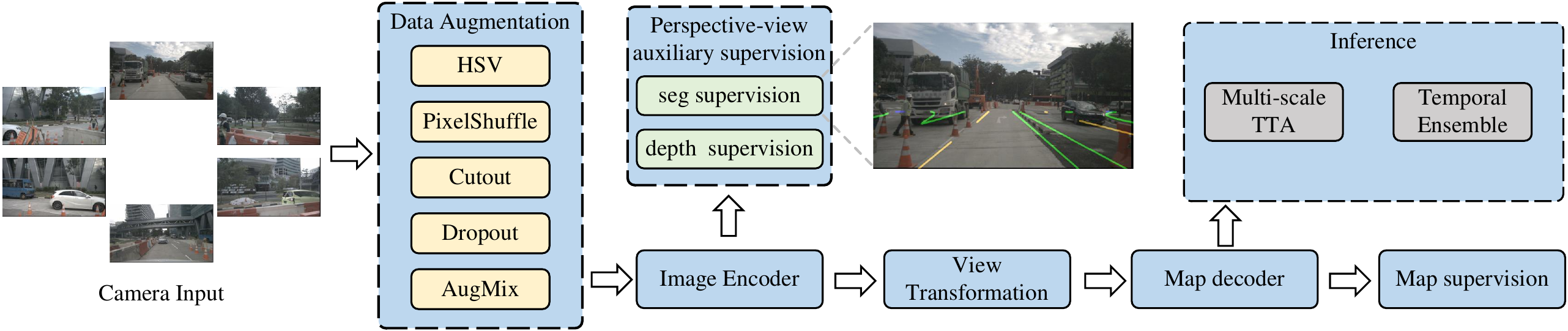}
    \caption{Overview of the SafeMapSSR framework. This diagram illustrates the integration of model modifications, advanced data augmentation, and post-processing strategies to enhance robustness in map segmentation tasks.}
    \label{fig:safedrivessr_pipeline}
\end{figure*}

\begin{figure}[t]
    \centering
    \includegraphics[width=\linewidth]{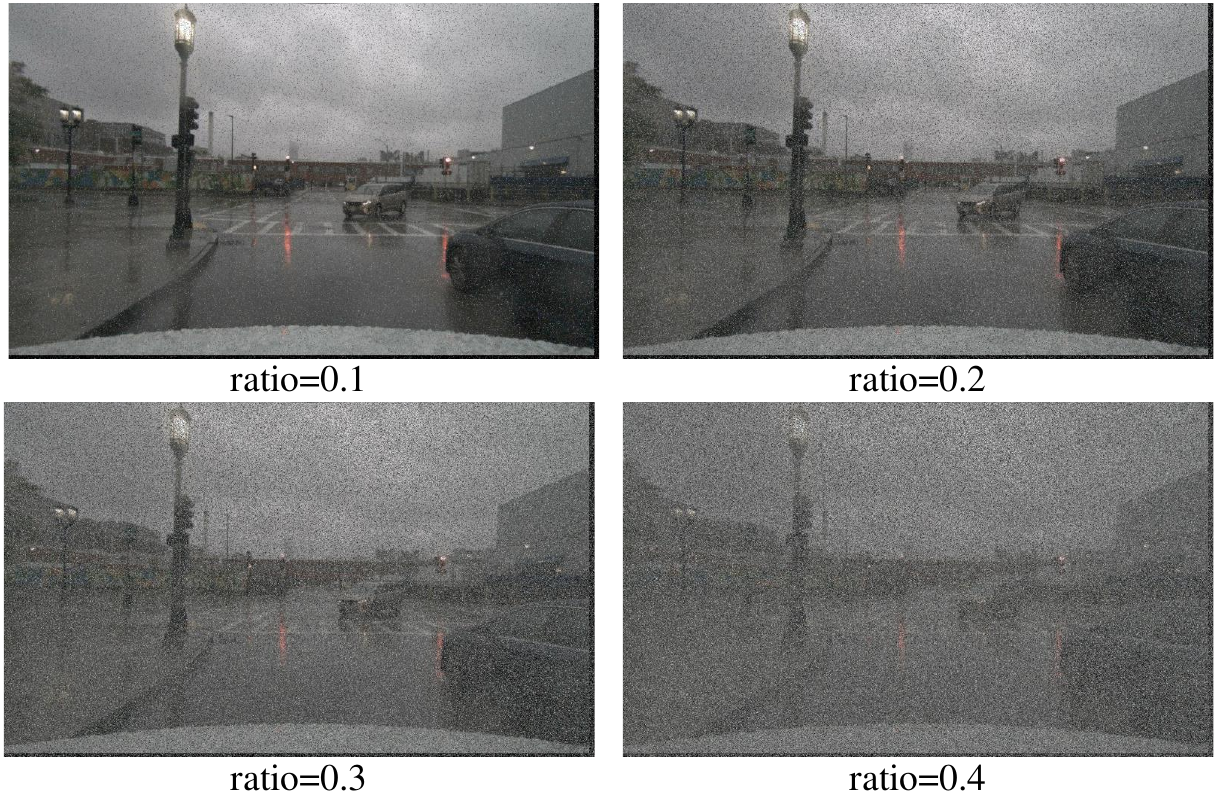}
    \caption{Examples of PixelShuffle and other augmentation techniques. These augmentations are crucial for enhancing the model's robustness against various image corruptions.}
    \label{fig:safedrivessr_augmentation}
\end{figure}

\subsubsection{Team \textcolor{robo_blue}{SafeDriveSSR}}
This team developed the \textbf{SafeMapSSR} framework to address the challenge of robust map segmentation through a comprehensive approach that includes advanced data augmentation, model scaling, and effective post-processing strategies. Their framework is built upon the BEVerse \cite{zhang2022beverse} architecture, with significant enhancements tailored to the map segmentation task. They optimized the model for high robustness in environments affected by various sensor corruptions such as fog, noise, and low light conditions.

\noindent\textbf{\faLightbulbO~Key Innovations:}
\begin{itemize}
    \item \textbf{Model Enhancement for Map Segmentation:} The BEVerse model architecture was modified to focus solely on map segmentation, optimizing the depth prediction range and employing a perspective-view map loss to improve accuracy.
    \item \textbf{Data Augmentation Combo:} Introduction of PixelShuffle augmentation, which randomly shuffles a certain percentage of pixels, alongside HSV augmentation, Cutout, and Dropout to enhance model generalization under corrupted conditions.
    \item \textbf{Temporal and Spatial Post-Processing:} Implementation of temporal ensembling and multi-scale test-time augmentation to refine the map segmentation outputs, ensuring consistency and accuracy even in dynamic scenes.
\end{itemize}

\noindent\textbf{\faGear~Implementation Details:}\\
The framework integrates several key components to maximize effectiveness, including fine-tuning the depth classification module in BEVerse \cite{zhang2022beverse} to better accommodate map segmentation needs and introducing a perspective-view segmentation head that processes map elements directly onto the camera plane. This enhancement significantly improves the model's ability to parse and segment complex map features from various camera angles, as illustrated in \cref{fig:safedrivessr_pipeline}. Additionally, a robust data augmentation strategy is proposed in preparing the model to handle unexpected or extreme variations in input data quality. Techniques such as PixelShuffle, which randomly shuffles a specified percentage of pixels, alongside HSV modifications, Cutout \cite{devries2017cutout}, and Dropout \cite{srivastava2014dropout}, are employed to train the model under simulated corrupted conditions, enhancing generalization across diverse operational scenarios. This strategy is shown in \cref{fig:safedrivessr_augmentation}. Post-processing techniques further leverage the temporal dimension of input data to refine and stabilize segmentation outputs, combining results from multiple frames to ensure consistency and accuracy even in dynamic scenes. This method is particularly effective for static map elements, ensuring reliable segmentation outputs over time.

\subsubsection{Team \textcolor{robo_blue}{CrazyFriday}}
This team developed the \textbf{MultiViewRobust} framework to address the challenge of robust map segmentation, leveraging multi-view architecture and advanced temporal information integration. The framework is based on the BEVerse \cite{zhang2022beverse} architecture but includes significant enhancements for handling real-world corruptions, such as fog, noise, and low light conditions that typically degrade the performance of map segmentation systems.

\noindent\textbf{\faLightbulbO~Key Innovations:}
\begin{itemize}
    \item \textbf{Enhanced Backbone Integration:} The team utilized large-scale backbones like Swin-L~\cite{liu2021swin} and EVA-02~\cite{fang2023eva} for feature extraction, which are crucial for maintaining robustness under various corruption scenarios. These backbones provide a strong foundation for feature richness and diversity.
    \item \textbf{Temporal and Multi-View Fusion:} By incorporating image-to-BEV transformation and multi-view features integration, the framework effectively handles temporal discrepancies and leverages spatial context, enhancing map segmentation accuracy.
    \item \textbf{Advanced Post-Processing Techniques:} Employing techniques like pillar pooling~\cite{lang2019pointpillars} and spatio-temporal BEV encoding, the framework refines the map segmentation outputs, ensuring high fidelity in the representation of dynamic and complex urban environments.
\end{itemize}

\noindent\textbf{\faGear~Implementation Details:}\\
The MultiViewRobust framework capitalizes on its robust architecture by incorporating advanced backbones pre-trained on extensive datasets like ImageNet-21K \cite{imagenet}, which significantly enhance feature extraction capabilities. For each timestamp, the framework efficiently aggregates multi-view features using a view transformer, extending coverage and translating these into a detailed point cloud. This point cloud is processed through pillar pooling to create precise BEV feature representations, which is essential for accurate map segmentation.
Following the feature extraction, the system implements temporal alignment to synchronize the BEV features from past timestamps with the current frame. This alignment, facilitated by known ego motions, is crucial for maintaining consistency across dynamic scenes. Spatio-temporal encoding further refines the temporal features, ensuring that the map decoders can effectively construct a semantic map. The decoders employ a simple yet effective MLP setup, translating robust features into precise map details, and optimizing the overall performance and accuracy of the map segmentation under varied conditions.

\begin{figure*}[t]
    \centering
    \includegraphics[width=\linewidth]{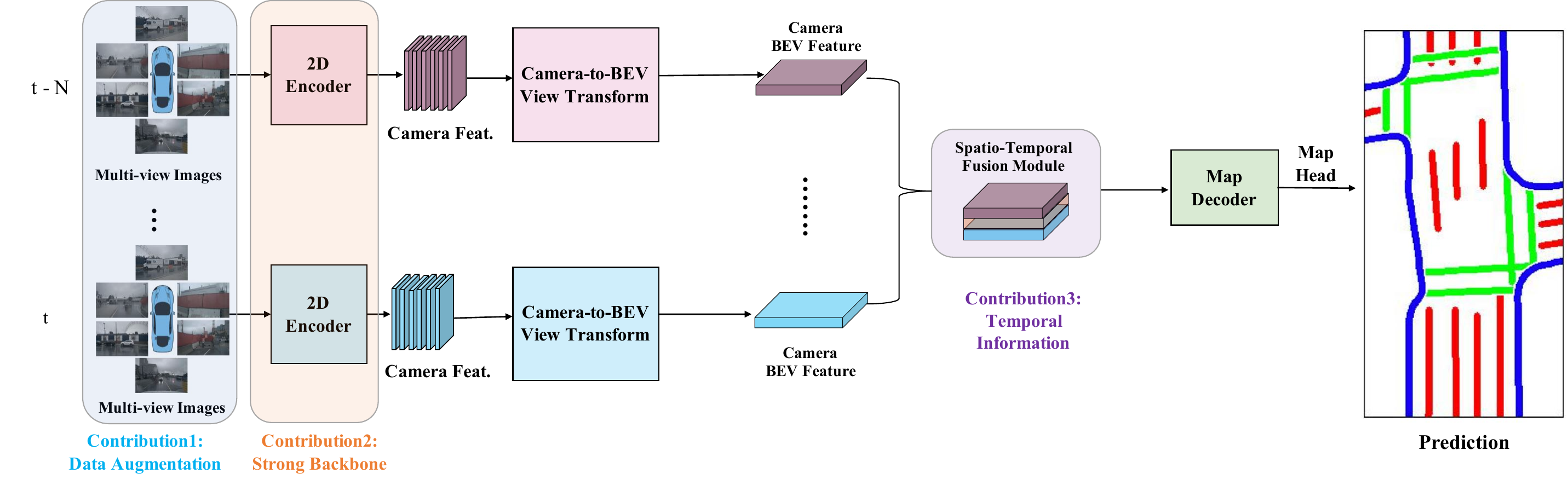}
    \caption{Overview of the DynamicBEV framework. This diagram illustrates the integration of advanced backbone, temporal fusion, and data augmentation strategies to tackle robust map segmentation challenges effectively.}
    \label{fig:samsung_overview}
\end{figure*}

\begin{figure*}[t]
    \centering
    \includegraphics[width=\linewidth]{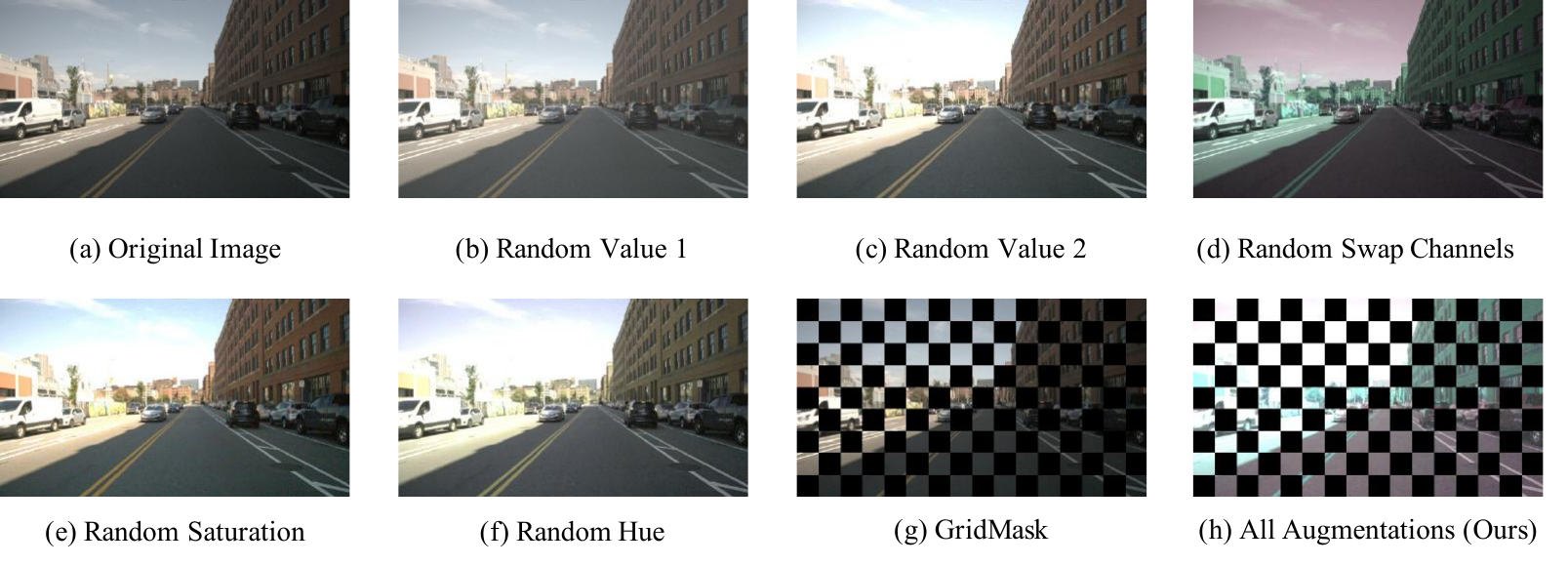}
    \caption{Examples of augmentation methods used in the DynamicBEV method. Techniques like Random Swapping Channels and GridMask are shown, highlighting their effectiveness in training the model to handle a variety of environmental conditions.}
    \label{fig:samsung_augmentation}
\end{figure*}

\subsubsection{Team \textcolor{robo_blue}{Samsung Research China-Advanced Research Lab}}
This team presented the \textbf{DynamicBEV} framework in the robust map segmentation track, aimed at enhancing the resilience of BEV map segmentation under diverse and challenging environmental conditions. Their approach integrates advanced temporal data fusion, a robust backbone for feature extraction, and strategic data augmentation methods to improve segmentation accuracy significantly.

\noindent\textbf{\faLightbulbO~Key Innovations:}
\begin{itemize}
    \item \textbf{Temporal Information Utilization:} The team implemented a temporal fusion module that integrates data across multiple frames, enhancing the robustness of the segmentation model by stabilizing the output over time.
    \item \textbf{Strong Backbone Employment:} Utilizing high-capacity models like Swin Transformers~\cite{liu2021swin} ensures deep and effective feature extraction, which is critical for maintaining performance under conditions like poor lighting and adverse weather.
    \item \textbf{Effective Data Augmentation Combo:} Techniques such as Random Swapping Channels, GridMask, and variations in hue and saturation were applied to train the model to be resilient against corruption and unexpected scenarios.
\end{itemize}

\noindent\textbf{\faGear~Implementation Details:}\\
The DynamicBEV framework, as depicted in illustrated in \cref{fig:samsung_overview}, employs the Swin Transformer \cite{liu2021swin} as a robust backbone within the BEVerse \cite{zhang2022beverse} architecture. This backbone is instrumental in processing high-resolution images from multiple camera views into a unified BEV map. The integration of temporal data through their sophisticated temporal fusion module plays a crucial role, merging information from successive frames to ensure that dynamic and static elements within the BEV are accurately and consistently represented across time. This approach is critical for maintaining map integrity and reliability. Additionally, the data augmentation strategy shown in \cref{fig:samsung_augmentation} employed by the team introduces a range of realistic variations during training, which significantly enhances the model’s resilience and generalization capabilities. Techniques such as Random Swapping Channels, GridMask, and varied adjustments in hue and saturation ensure that the model is well-prepared to handle common visual corruptions and unexpected scenarios encountered in real-world driving conditions. The integration of these advanced techniques allows the DynamicBEV framework to deliver superior performance in robust map segmentation tasks, showcasing their effectiveness in the challenge.

%% file: subsections/track3.tex
\begin{figure*}[t]
    \centering
    \includegraphics[width=\linewidth]{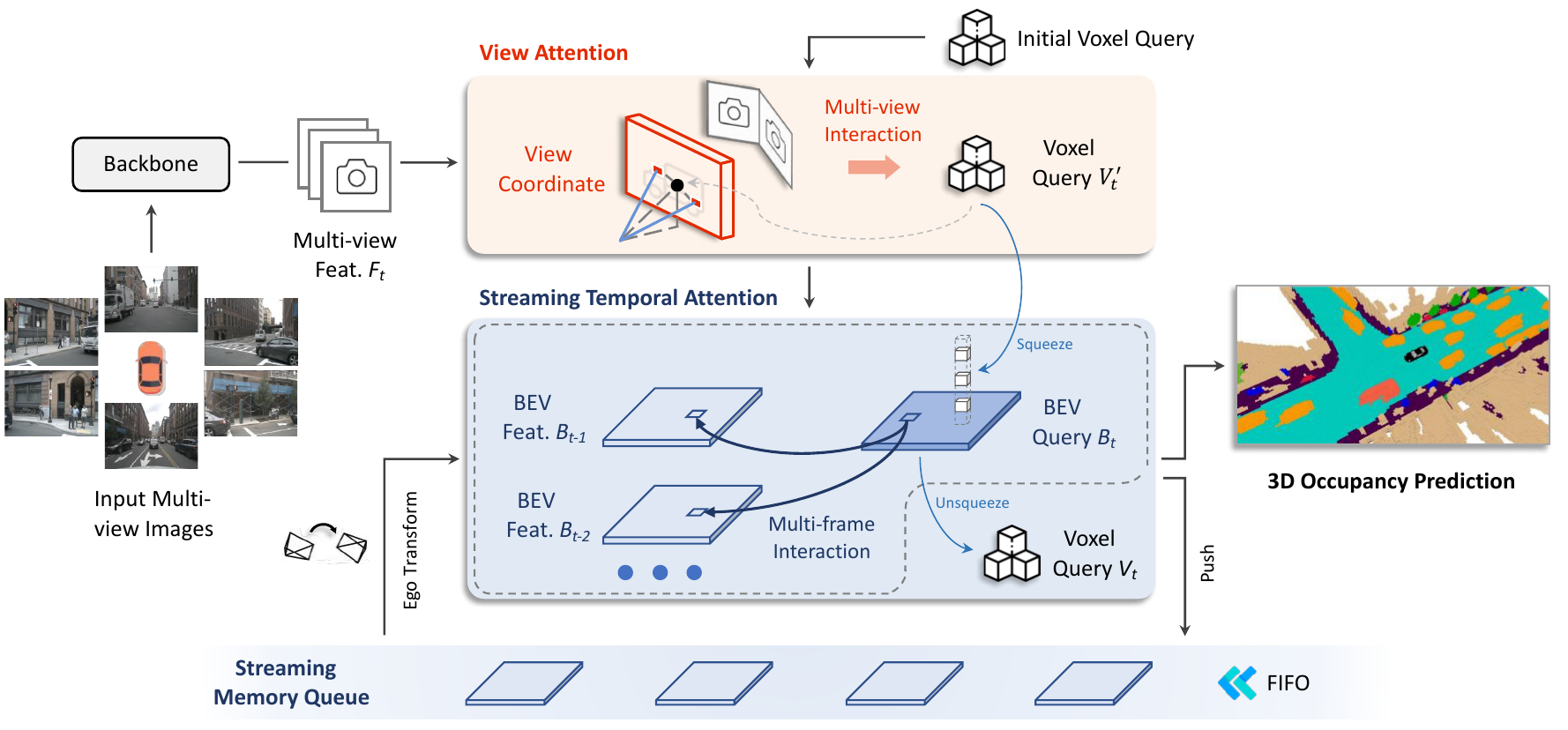}
    \caption{Overview of the ViewFormer Enhanced framework, showing the integration of multi-view features through view attention and the streaming temporal attention mechanism for robust 3D occupancy prediction.}
    \label{fig:viewformer_overview}
\end{figure*}

\begin{figure}[t]
    \centering
    \includegraphics[width=\linewidth]{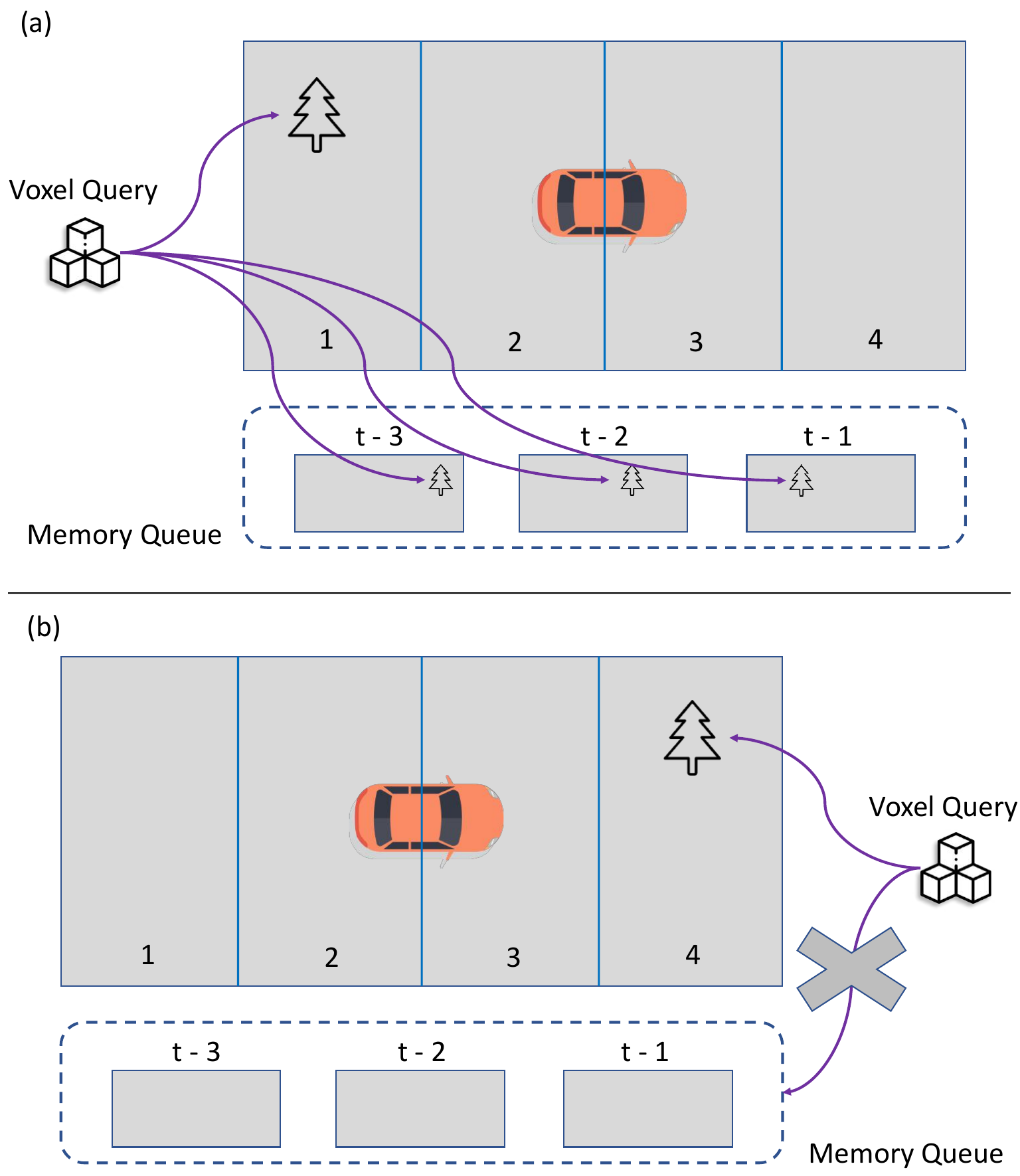}
    \caption{Illustration of the reverse video playback mechanism employed by ViewFormer Enhanced, demonstrating how future data is utilized to enhance occupancy prediction accuracy.}
    \label{fig:viewformer_reverse_playback}
\end{figure}

\begin{figure*}[t]
    \centering
    \includegraphics[width=\linewidth]{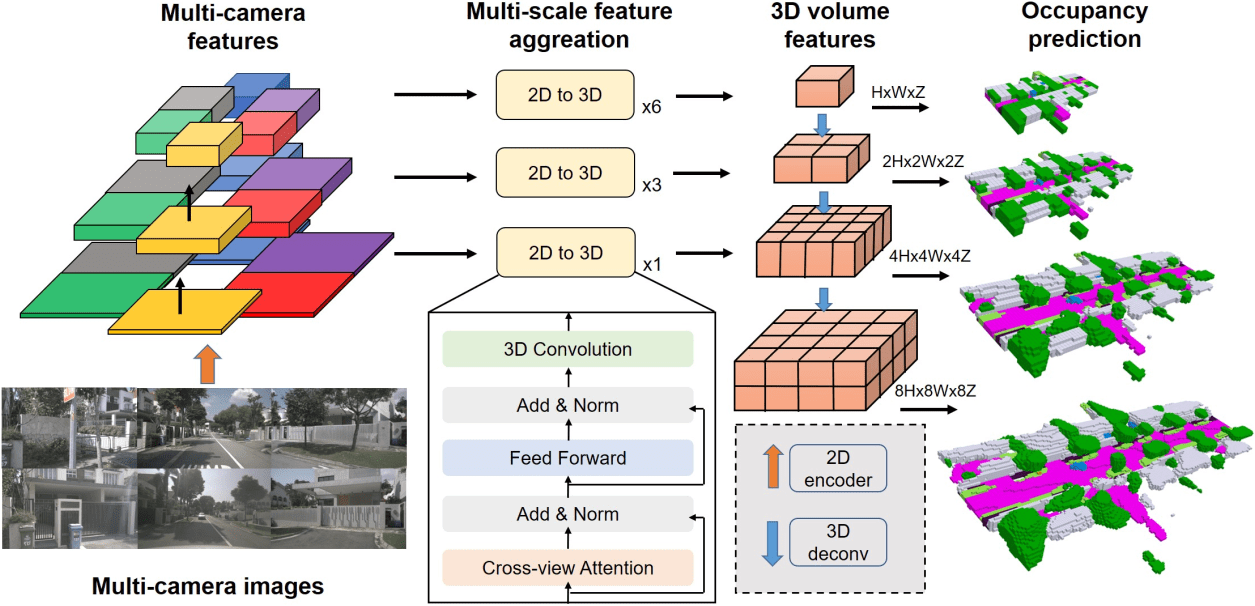}
    \caption{Overview of the SurroundOcc Enhanced framework, illustrating the integration of advanced backbones, fine-tuning strategies, and diverse model ensembling for robust occupancy prediction.}
    \label{fig:apecblue_framework_overview}
\end{figure*}

\subsubsection{Team \textcolor{robo_blue}{ViewFormer}}
This team introduced the \textbf{Enhanced ViewFormer} framework in the Robust Occupancy Prediction track, aiming to significantly boost the robustness and accuracy of occupancy predictions in autonomous driving systems. Leveraging sophisticated spatiotemporal modeling techniques, the framework is uniquely designed to handle complex dynamics and various environmental conditions using multi-view camera inputs. This approach ensures that the system not only maintains high accuracy but also adapts to sudden changes in the environment, which are common in real-world scenarios.

\noindent\textbf{\faLightbulbO~Key Innovations:}
\begin{itemize}
    \item \textbf{Spatial Interaction through View Attention:} Employing view-guided transformers, the team's approach allows for effective aggregation of multi-view image features into a unified 3D occupancy map, enhancing spatial accuracy.
    \item \textbf{Streaming Temporal Attention:} The team utilizes a memory mechanism to incorporate online video data efficiently for temporal interaction, reducing training time while maintaining consistency across temporal data.
    \item \textbf{Reverse Video Playback Mechanism:} As an offline extension, this innovative approach uses future frame data to enhance the accuracy of predictions by playing video sequences backward, effectively leveraging additional temporal information.
\end{itemize}

\noindent\textbf{\faGear~Implementation Details:}\\
The Enhanced ViewFormer framework utilizes a transformer-based architecture designed for robust multi-view 3D occupancy perception~\cite{li2022bevformer, liu2022petr, wang2022detr3d}. It processes features from multiple camera views through the attention mechanism, seamlessly integrating them into a coherent 3D space, effectively enhancing spatial accuracy as presented in \cref{fig:viewformer_overview}. This system incorporates a streaming memory queue for temporal modeling, which dynamically captures and integrates features over time, reducing inconsistencies caused by environmental changes and enhancing scene understanding~\cite{park2022time, wang2023exploring}.
Furthermore, the inclusion of a reverse video playback feature markedly improves model performance by using future data in a reverse sequence, allowing the model to access additional temporal information that enhances occupancy prediction accuracy. This feature, detailed in \cref{fig:viewformer_reverse_playback}, leverages future data to mitigate the limitations of real-time processing and provides a more comprehensive scene analysis.
These spatial and temporal processing techniques significantly enhance the framework's performance, contributing to its high accuracy in challenging conditions and demonstrating its robustness in adverse environments. This integration ensures that ViewFormer Enhanced remains highly accurate and reliable across varying conditions.

\subsubsection{Team \textcolor{robo_blue}{APEC Blue}}
Team APEC Blue introduced the \textbf{SurroundOcc Enhanced} framework in the Robust Occupancy Prediction track, aiming to refine occupancy prediction in complex driving environments. Their methodology emphasizes refining the baseline SurroundOcc model~\cite{wei2023surroundocc}, optimizing network structures for enhanced performance, and strategically ensembling multiple models to bolster robustness and accuracy.

\noindent\textbf{\faLightbulbO~Key Innovations:}
\begin{itemize}
    \item \textbf{Model Fine-tuning:} Focused fine-tuning of the baseline model on specific loss functions to direct learning more effectively and improve prediction precision.
    \item \textbf{Optimizing Network Structure:} Adjustments to network architecture, including experimenting with different backbones and voxel sizes, to optimize the spatial and feature resolution impacts on model performance.
    \item \textbf{Algorithmic Model Ensembling:} Integration of diverse models to improve the overall robustness and accuracy by leveraging the strengths of various approaches.
\end{itemize}

\noindent\textbf{\faGear~Implementation Details:}\\
The SurroundOcc Enhanced framework builds upon the SurroundOcc architecture, incorporating advanced backbones like VoVNet-99~\cite{lee2019energy} and ResNet-101~\cite{he2016deep} to compare their efficacy in feature extraction and their impact on model performance. The team explored different voxel sizes to understand their influence on the mIoU scores, which helped in tuning the model’s spatial resolution for better accuracy. Their fine-tuning strategy focused on minimizing loss from the final layer, which proved essential in directing the model's learning toward more relevant features for occupancy prediction. The general framework and approach to integrating these elements are shown in \cref{fig:apecblue_framework_overview}.
Additionally, the team employed a comprehensive ensembling strategy that combined models with varied backbones and training configurations to create a robust system capable of handling diverse and challenging scenarios, resulting in a notable improvement in mIoU scores across a range of test conditions.

\subsubsection{Team \textcolor{robo_blue}{hm.unilab}}
This team tested the state-of-the-art occupancy prediction method, \textbf{SurroundOcc}, adapting it for our challenge to enhance its robustness against out-of-distribution data. They focused on refining the model's backbone and integrating a combination of diverse loss functions to improve performance significantly under varied environmental conditions.

\noindent\textbf{\faLightbulbO~Key Innovations:}
\begin{itemize}
\item \textbf{Backbone Enhancement:} Adopting ResNet-101 \cite{he2016deep} as the backbone to leverage its deep network architecture, allowing for more effective and nuanced feature extraction crucial for occupancy prediction tasks.
\item \textbf{Comprehensive Loss Function Strategy:} Implementing a multi-faceted loss function approach that includes cross-entropy loss, segmentation scale loss, and geo scale loss, each contributing to the model's ability to fine-tune predictions more accurately across different scales and geometrical dimensions.
\end{itemize}

\noindent\textbf{\faGear~Implementation Details:}\
Team \textcolor{robo_blue}{hm.unilab}'s method involved several enhancements to the SurroundOcc framework. The use of the ResNet-101 \cite{he2016deep} backbone was pivotal in processing multi-view images to extract high-quality features that form the basis for accurate occupancy predictions. Their implementation of diverse loss functions aimed to optimize the model's performance by addressing different aspects of the prediction task, thereby enhancing the robustness of the model to various real-world disturbances such as weather changes and sensor failures. The team conducted a series of experiments to test different combinations of backbones and loss functions, ultimately achieving the highest mIoU score with ResNet-101 \cite{he2016deep} and their innovative loss strategy. This comprehensive approach not only improved the robustness of the occupancy prediction method but also ranked them 3rd in the competition, demonstrating the effectiveness of their modifications in challenging OoD scenarios.

%% file: subsections/track4.tex
\subsubsection{Team \textcolor{robo_blue}{HIT-AIIA}}
Team \textcolor{robo_blue}{HIT-AIIA} presented the \textbf{DINO-SD} (Dinov2-Surround Depth) model in the Robust Depth Estimation track, aimed at enhancing the reliability of depth estimation in autonomous driving through advanced multi-view image processing. Their approach leverages a combination of robust feature extraction and innovative depth prediction techniques to improve accuracy in corrupted environments.

\begin{figure*}[t]
    \centering
    \includegraphics[width=\linewidth]{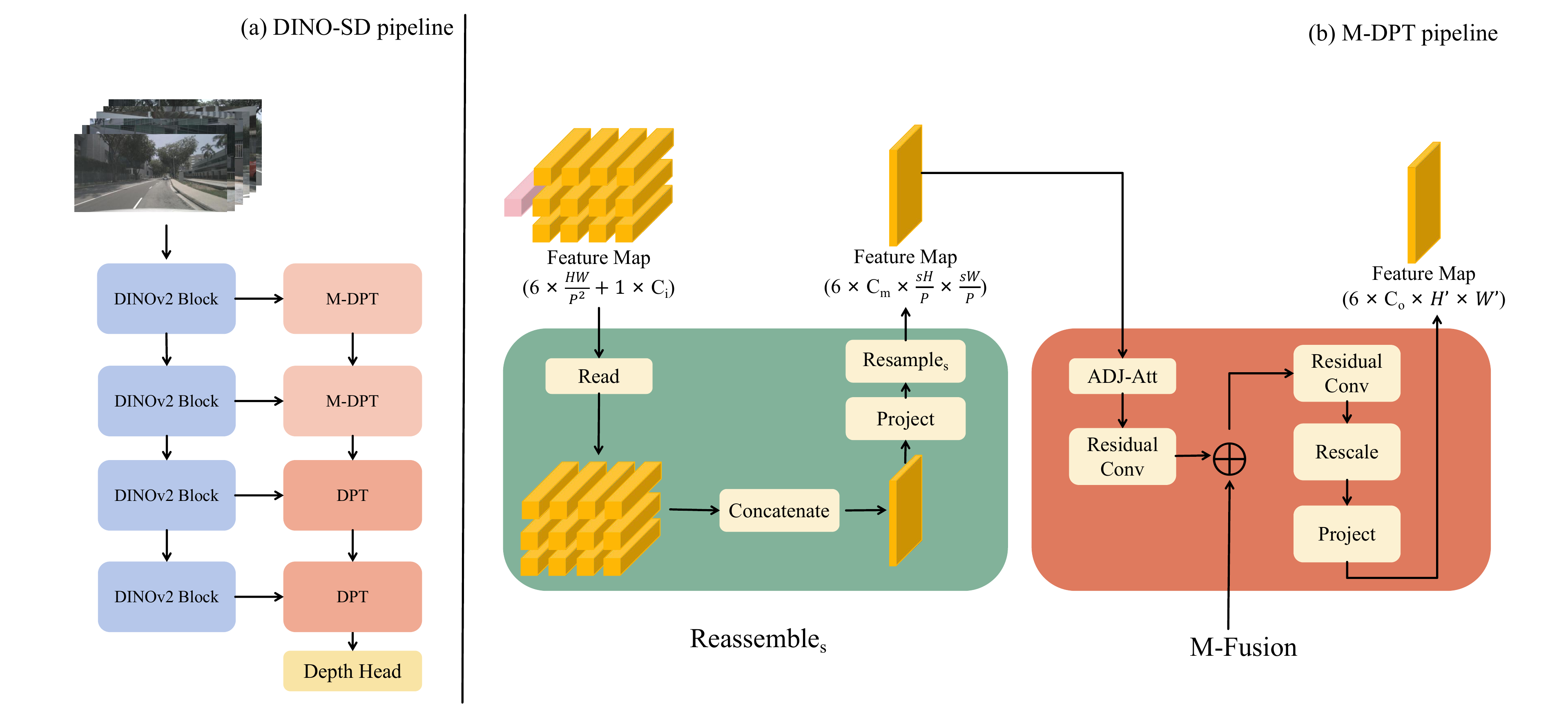}
    \caption{The DINO-SD pipeline. This figure shows the overall workflow of the DINO-SD model, emphasizing the integration of the DINOv2 encoder, M-DPT decoder, and the advanced data processing techniques used to enhance depth estimation accuracy.}
    \label{fig:dinosd_pipeline}
\end{figure*}

\begin{figure*}[t]
    \centering
    \begin{subfigure}{0.47\linewidth}
       \includegraphics[width=\linewidth]{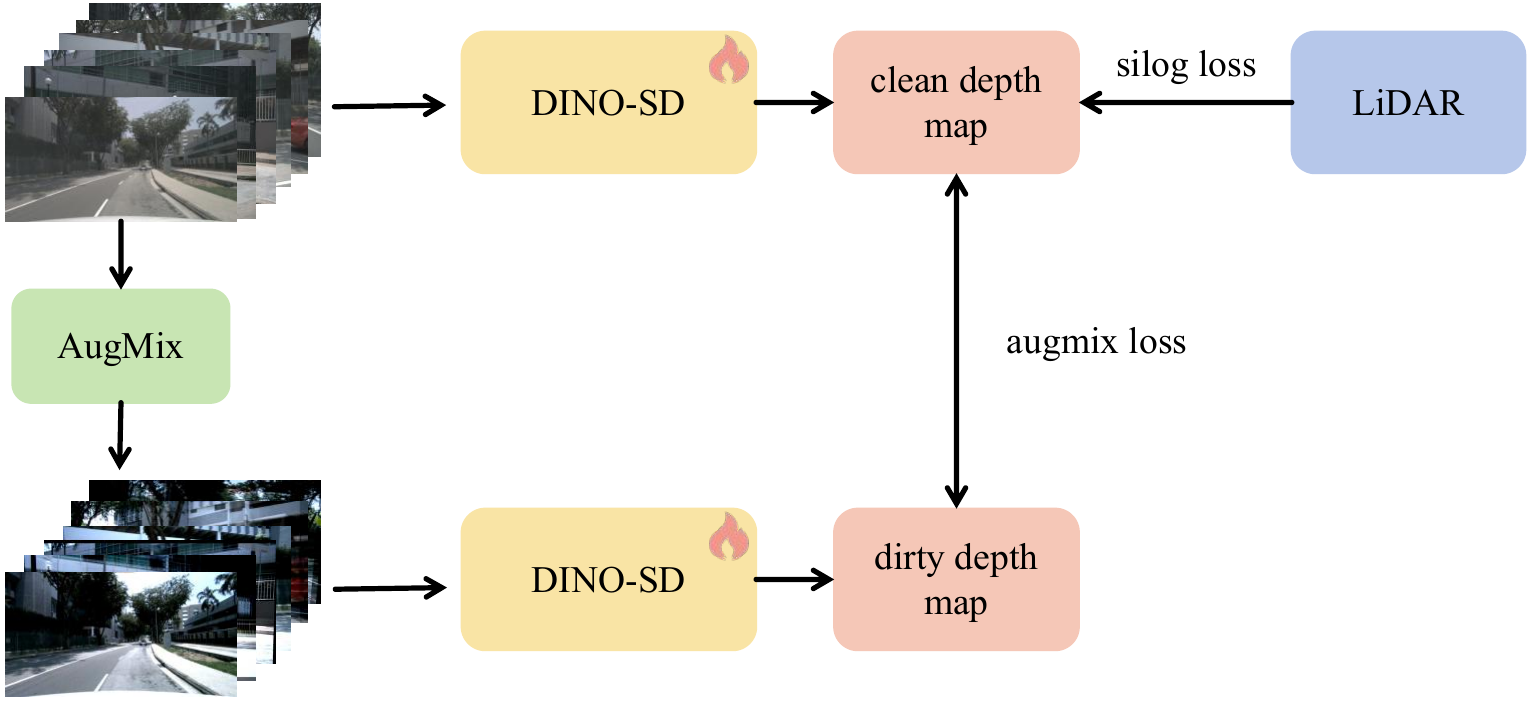}
        \caption{Training pipeline.}
        \label{fig:training-pipeline} 
    \end{subfigure}
    \hfill
    \begin{subfigure}{0.47\linewidth}
        \includegraphics[width=\linewidth]{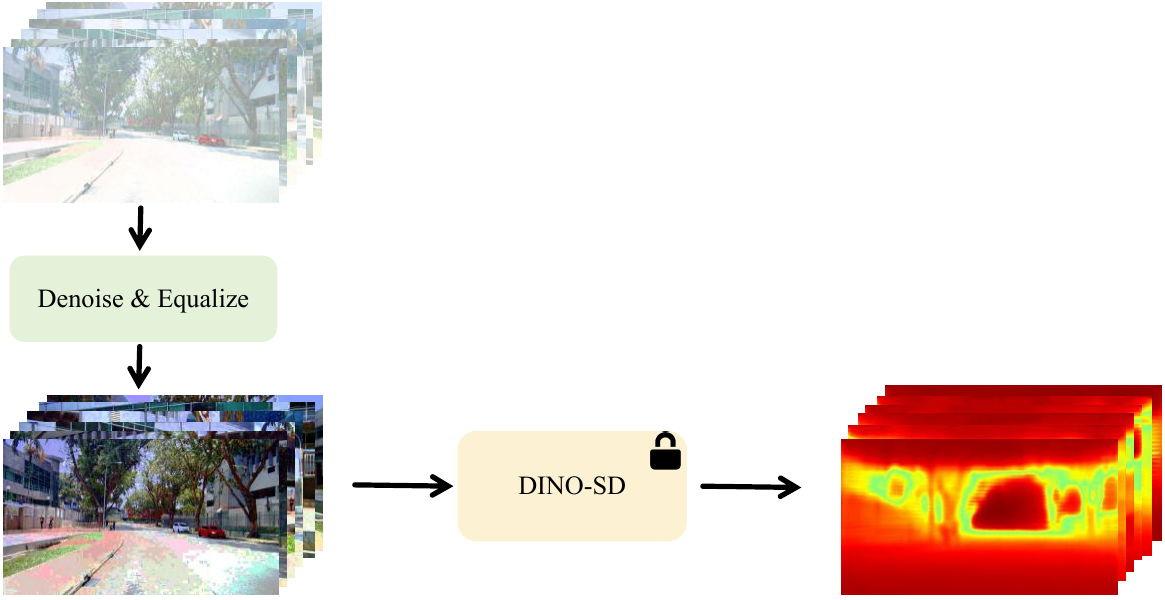}
        \caption{Testing pipeline.}
        \label{fig:testing-pipeline}
    \end{subfigure}
    \caption{Training and testing pipelines of the DINO-SD model. The two diagrams illustrate several useful image processing techniques, including denoising and equalization, employed to refine the performance of the depth estimation model in real-world conditions.}
    \label{fig:dinosd_testing_pipeline}
\end{figure*}

\begin{figure*}[t]
    \centering
    \includegraphics[width=\linewidth]{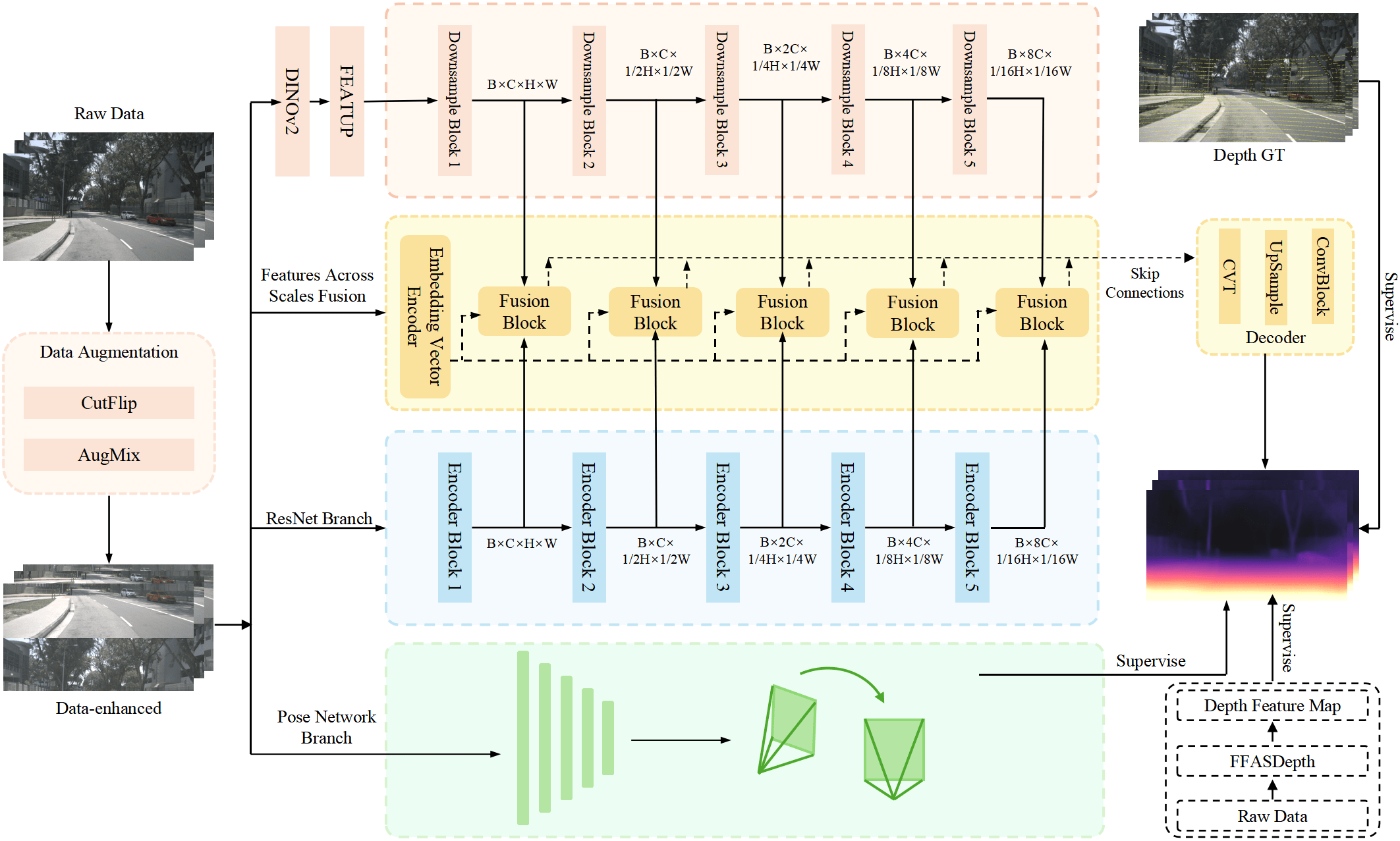}
    \caption{Overview of the FFASDepth framework. This figure shows the integration of dual-branch network architecture, channel-attention-based feature fusion, and semi-supervised data augmentations to enhance the robustness of depth estimation.}
    \label{fig:ffa_depth_overview}
\end{figure*}

\begin{figure*}[t]
    \centering
    \includegraphics[width=0.95\linewidth]{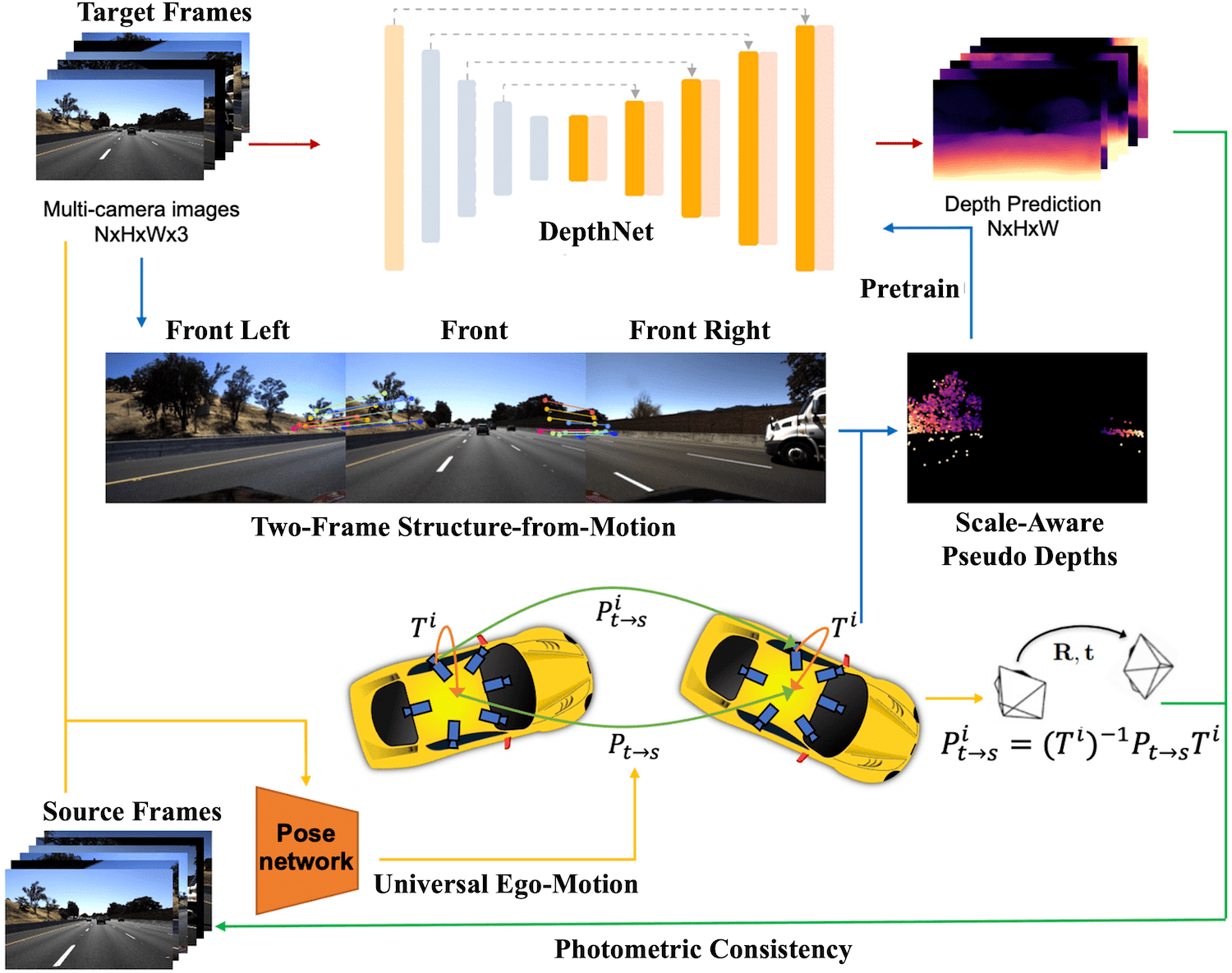}
    \caption{Overview of the MonoViT+TTA framework showing the integration of the MPVIT encoder and transformer-based depth decoding layers, highlighted by the use of advanced image restoration techniques at test time to enhance input quality.}
    \label{fig:custzs_monovit_structure}
\end{figure*}

\noindent\textbf{\faLightbulbO~Key Innovations:}
\begin{itemize}
    \item \textbf{Robust Feature Extraction:} Utilizing the pre-trained DINOv2~\cite{oquab2023dinov2} as an encoder, the team enhanced the capability of the model to handle out-of-distribution data by extracting robust image features that are less susceptible to common corruptions.
    \item \textbf{Multi-View Depth Prediction:} The team developed a multiview-DPT (M-DPT) decoder to fuse and decode features across multiple views, effectively improving the spatial consistency and accuracy of the depth maps produced.
    \item \textbf{Enhanced Data Augmentation:} AugMix~\cite{hendrycks2019augmix}, a method that combines different augmentation strategies, was used to train the model, increasing its resilience to various image corruptions.
\end{itemize}

\noindent\textbf{\faGear~Implementation Details:}\\
The DINO-SD model architecture is meticulously designed to process multi-view images and produce accurate and consistent depth maps efficiently. Utilizing the powerful DINOv2 as the backbone, the model benefits from high-quality feature extraction, which is crucial for robust performance in varied and challenging environments. The innovative M-DPT decoder, which includes adjacent-view attention mechanisms, significantly enhances feature fusion across different camera views, optimizing the spatial resolution of the output depth maps. This setup ensures that depth estimation is not only precise but also stable across different viewpoints, as depicted in \cref{fig:dinosd_pipeline}.
Further refining the model predictions, a depth head meticulously processes the fused features to generate finely detailed depth maps, capturing the subtleties of the surrounding environment. This enhanced processing capability is critical for applications such as autonomous driving, where accuracy and reliability are paramount. In addition to a robust training regime, the team implemented an advanced testing pipeline that incorporates denoising and equalization techniques to further improve the performance of the model under practical, noisy conditions typically found in real-world scenarios. The testing pipeline, which enhances the model’s adaptability to real-world conditions, is illustrated in \cref{fig:dinosd_testing_pipeline}.

\subsubsection{Team \textcolor{robo_blue}{BUAA Trans}}
This team introduced the \textbf{FFASDepth} (Fusing Features Across Scales Depth Estimation) framework, designed to enhance the robustness of depth estimation models in out-of-distribution scenarios, specifically for autonomous driving applications. Their innovative approach incorporates multi-branch network architectures and advanced data augmentation techniques to tackle the challenges posed by adverse weather and lighting conditions.

\noindent\textbf{\faLightbulbO~Key Innovations:}
\begin{itemize}
    \item \textbf{Multi-branch Network Architecture:} The FFASDepth employs DINOv2~\cite{oquab2023dinov2} and ResNet \cite{he2016deep} as backbones for multi-scale feature extraction, significantly improving the model's adaptability and accuracy under varied conditions.
    \item \textbf{Channel-Attention Based Feature Fusion:} The framework utilizes a novel channel-attention mechanism that enables effective redistribution and fusion of features from both branches, tailoring the feature integration process to enhance depth prediction.
    \item \textbf{Semi-supervised Data Augmentations:} Incorporating techniques like CutFlip and AugMix, the model enhances its generalization capabilities, making it robust against a broader range of image corruptions.
\end{itemize}

\noindent\textbf{\faGear~Implementation Details:}\\
The FFASDepth framework, depicted in \cref{fig:ffa_depth_overview}, enhances the SurroundDepth~\cite{wei2023surrounddepth} architecture by implementing a dual-branch strategy. This approach utilizes DINOv2 for processing high-level semantic features, essential for capturing the contextual nuances of the environment, while ResNet \cite{he2016deep} extracts detailed edge features critical for precise depth boundary delineation. Together, these branches cover a comprehensive range of feature scales vital for depth estimation in diverse conditions.
The integration of these features is accomplished through a sophisticated SENet-based channel-attention mechanism~\cite{kundu2023learning}. This system dynamically adjusts and merges features by weighing their relevance, guided by the context provided by the input image embeddings. These embeddings serve as a blueprint for channel fusion, focusing the model's attention on the most pertinent features for depth estimation and enhancing the overall precision and reliability of the output.
To further bolster the model’s robustness, FFASDepth incorporates semi-supervised learning techniques like AugMix \cite{hendrycks2019augmix}, which introduces a variety of controlled image distortions during training. This prepares the model to handle unexpected corruption effectively. Additionally, the CutFlip \cite{shao2023urcdc} technique challenges the model’s reliance on traditional vertical image alignment, promoting a more adaptable and comprehensive understanding of depth cues. These strategies ensure that the model delivers consistent and accurate depth predictions across dynamic and unpredictable environments.

\subsubsection{Team \textcolor{robo_blue}{CUSTZS}}
This team presented the \textbf{MonoViT+TTA} model in the Robust Depth Estimation track, specifically designed to enhance the robustness of depth estimation in adverse conditions. Their approach integrates innovative architectural modifications and testing strategies to improve the model's performance across a variety of challenging scenarios.

\noindent\textbf{\faLightbulbO~Key Innovations:}
\begin{itemize}
    \item \textbf{Integration of Transformer Architecture:} By incorporating the MonoViT~\cite{zhao2022monovit} architecture, which leverages transformers for depth estimation, the team enhanced the model's ability to process spatial contexts and complex textures more effectively than conventional CNN-based approaches.
    \item \textbf{Use of Test-Time Augmentation (TTA):} Implementing image restoration models at test time, such as Restormer~\cite{zamir2022restormer}, to refine input quality and thereby enhance the accuracy of depth predictions under diverse conditions.
    \item \textbf{Robust Training Pipeline:} Focusing on training robustness, the team employed a combination of depth estimation models, with MonoViT emerging as the most effective, particularly when enhanced with TTA strategies.
\end{itemize}

\noindent\textbf{\faGear~Implementation Details:}\\
The team's MonoViT+TTA framework is built on a dual strategy of advanced feature extraction using transformers and resilience enhancement through test-time augmentation. The MonoViT model serves as the backbone, incorporating a CNN-Transformer hybrid architecture that significantly extends the capacity for detailed and context-aware feature extraction. This setup is depicted in \cref{fig:custzs_monovit_structure}. 
For depth estimation, the encoder utilizes the MPVIT~\cite{lee2022mpvit} structure to integrate the robust capabilities of CNNs with the extensive contextual understanding of transformers, providing a comprehensive feature set that is effectively decoded to predict depth. The integration of Restormer at test time plays a crucial role in enhancing the input images, reducing noise, and correcting artifacts that typically degrade the performance of depth estimation models under adverse conditions.
The testing pipeline incorporates various image restoration techniques before feeding images into the depth estimation model, ensuring that the depth predictions are based on the clearest possible inputs. This method has demonstrated considerable improvements in model performance, particularly in terms of absolute relative error and precision metrics on challenging datasets like KITTI~\cite{geiger2013vision}.

%% file: subsections/track5.tex
\begin{figure*}[t]
    \centering
    \includegraphics[width=\linewidth]{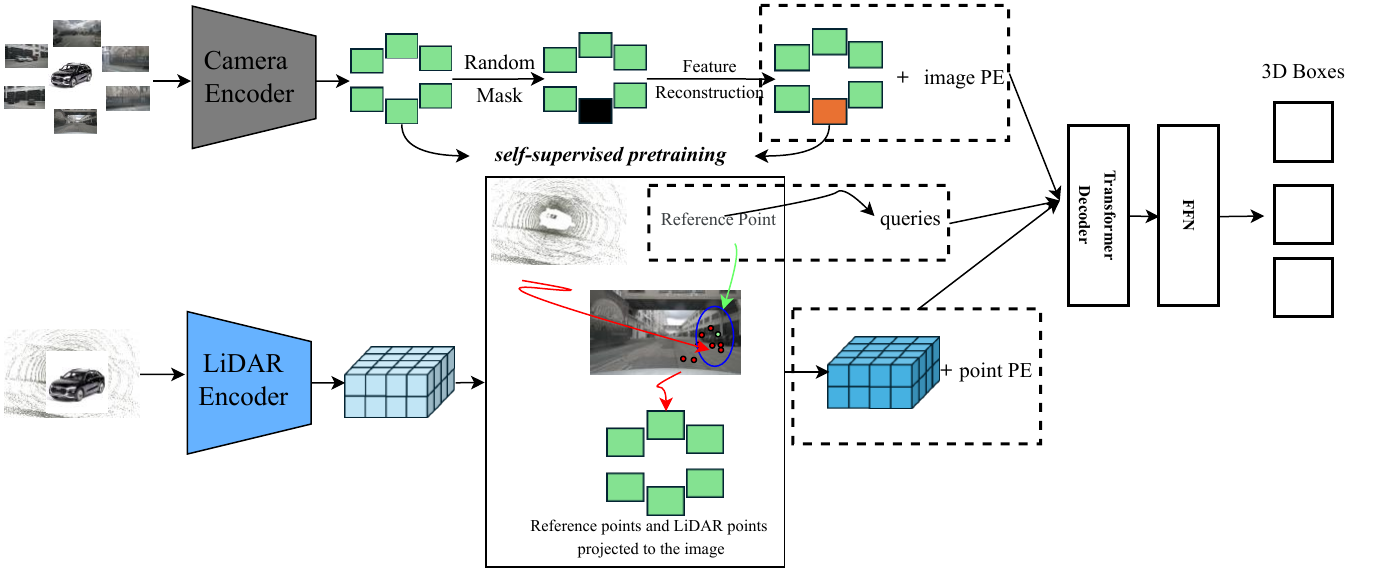}
    \caption{The ASF framework. This figure details the dual-modality integration and the strategic feature reconstruction and enhancement techniques that enable robust 3D object detection under sensor failure conditions.}
    \label{fig:asf_framework}
\end{figure*}

\subsubsection{Team \textcolor{robo_blue}{safedrive-promax}}
This team developed the \textbf{Against Sensor Failure (ASF)} model, targeting robust multi-modal BEV detection. This framework aims to enhance the resilience of multi-modal 3D object detection systems in scenarios where camera or LiDAR sensors fail or degrade. By focusing on advanced feature reconstruction and multi-modal feature enhancement, ASF addresses critical challenges that arise during sensor failures, ensuring reliable 3D object detection in diverse and adverse environmental conditions.

\noindent\textbf{\faLightbulbO~Key Innovations:}
\begin{itemize}
    \item \textbf{Self-Supervised Feature Reconstruction:} ASF employs a self-supervised pre-training process to reconstruct image features when cameras fail, using spatial cues to maintain detection capabilities without full sensor input.
    \item \textbf{Image Feature Enhancement for LiDAR (IEL):} This module enhances LiDAR data with corresponding image features to compensate for LiDAR sparsity or failures, significantly bolstering the system's robustness in multi-modal detection tasks.
    \item \textbf{Robust Fusion and Decoding Strategy:} Leveraging advanced cross-attention mechanisms, ASF integrates and decodes information from both imaging and LiDAR modalities to ensure precise object detection even under partial sensor failure.
\end{itemize}

\noindent\textbf{\faGear~Implementation Details:}\\
The ASF framework, as illustrated in \cref{fig:asf_framework}, operates on a foundation of the CMT~\cite{yan2023cross} architecture, adapted to handle dual-modal inputs effectively. During its self-supervised pre-training phase, the model masks random sections of image data, which are then reconstructed through a sophisticated decoder equipped with multiple transformer layers. This approach is not merely about replacing lost data but adapting the model's response to partial information, thereby ensuring continuous operation during sensor failures.
To address LiDAR sensor shortcomings, ASF simulates reduced sensor capability by thinning the point cloud data, which is then enriched with strategically projected image features. This is achieved by mapping reference points detected in the CMT onto corresponding image data, selectively enhancing LiDAR data with detailed image features. Such integration ensures that even with sparse or incomplete LiDAR data, the model can still deliver accurate and reliable object detection outputs.
These methods collectively improve the system's adaptability and fault tolerance, demonstrated by ASF's marked performance improvements over standard models under typical and stressed conditions.

\begin{figure}[t]
    \centering
    \includegraphics[width=\linewidth]{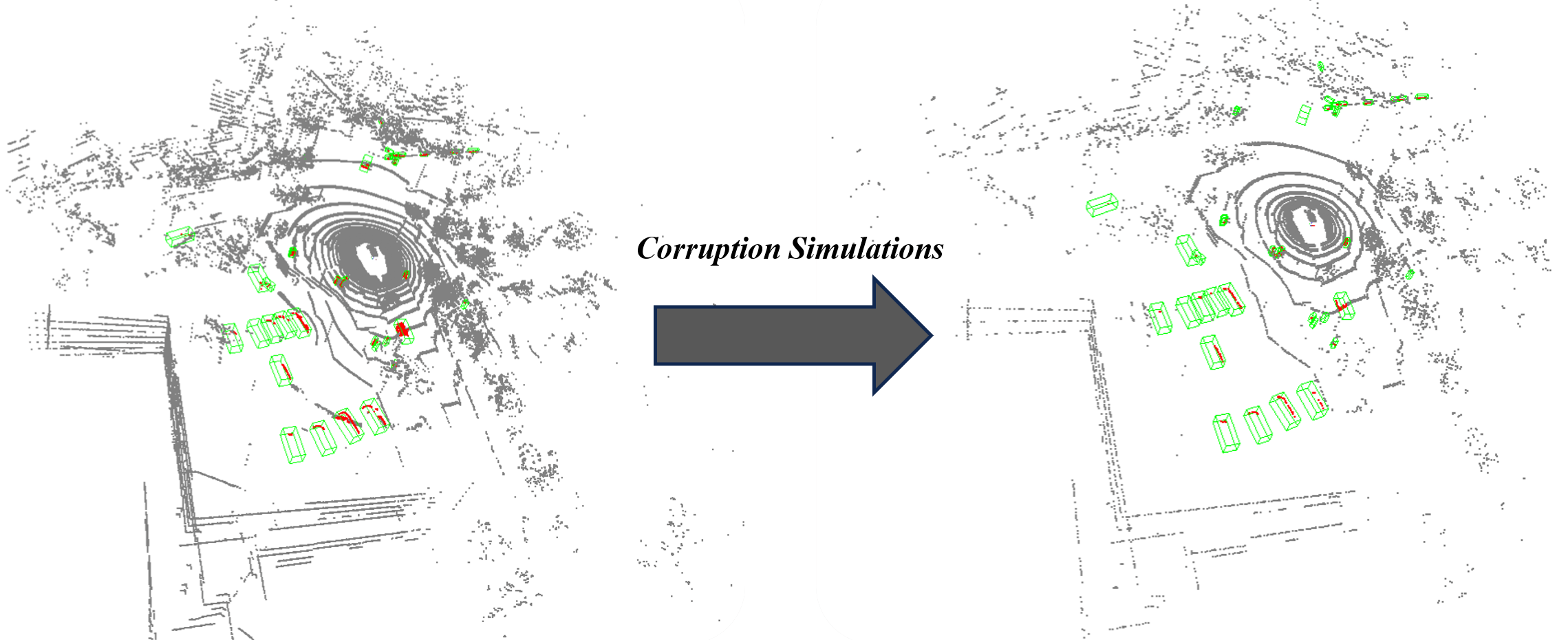}
    \caption{Corruption simulation strategy in RobuAlign, demonstrating robustness in the absence of LiDAR beams.}
    \label{fig:robualign_fig1}
\end{figure}

\begin{figure*}[t]
    \centering
    \includegraphics[width=\linewidth]{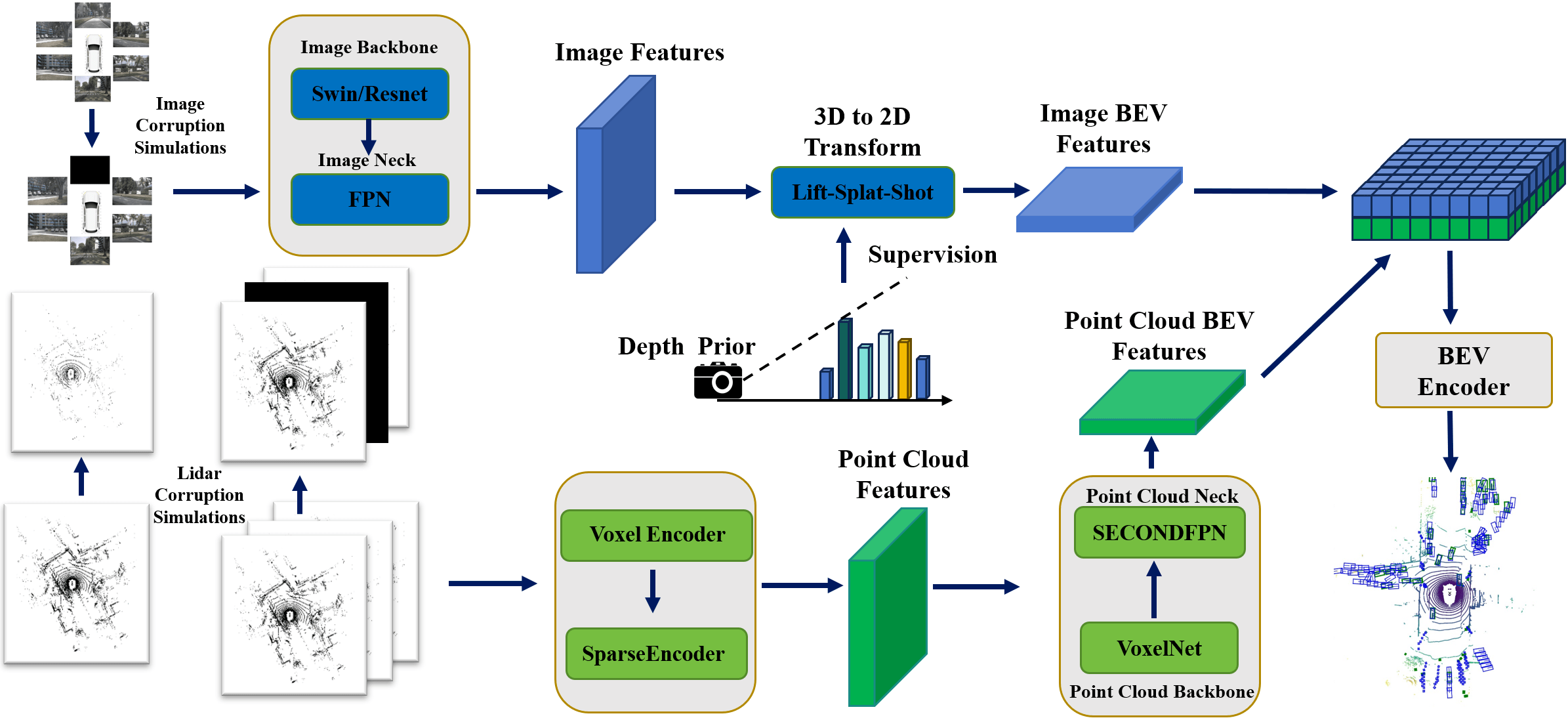}
    \caption{Overview of RobuAlign, highlighting the integration of image and LiDAR data into a unified BEV representation.}
    \label{fig:robualign_fig2}
\end{figure*}

\begin{figure}[t]
	\centering
	\begin{minipage}{0.49\textwidth}
		\includegraphics[width=\linewidth]{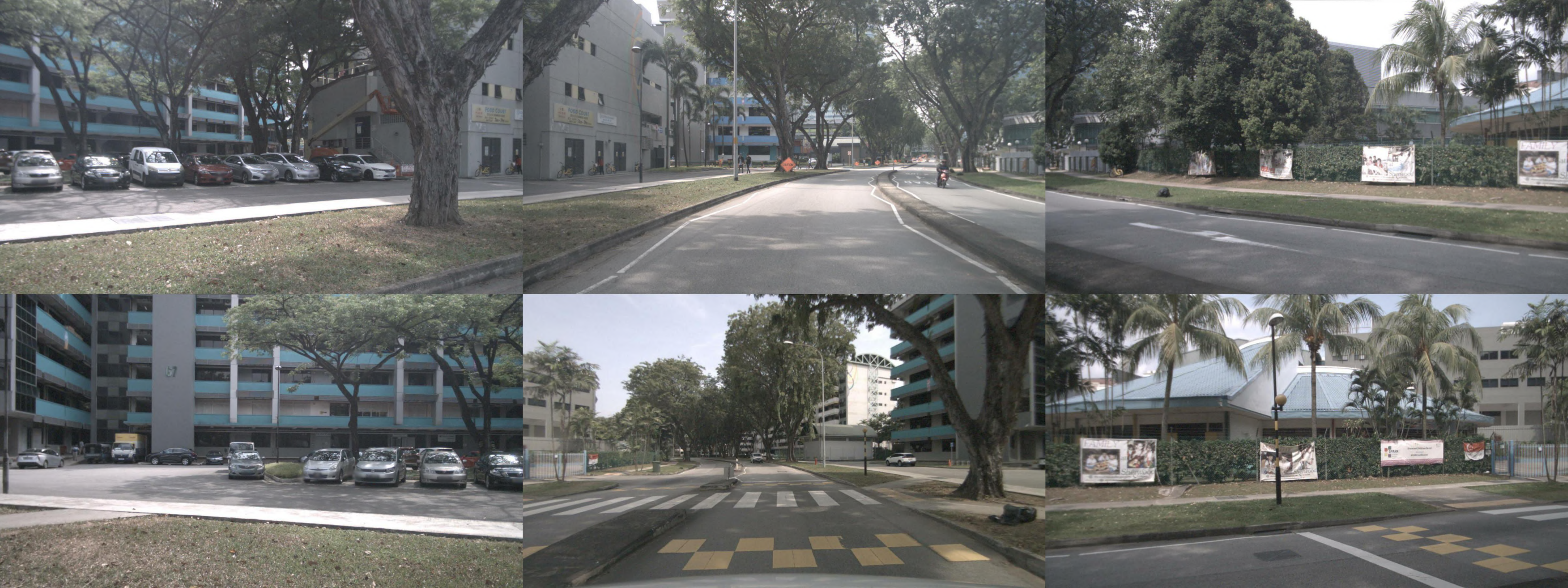}
		\subcaption{Normal Image}
	\end{minipage}
	\hfill
	\begin{minipage}{0.49\textwidth}
		\includegraphics[width=\linewidth]{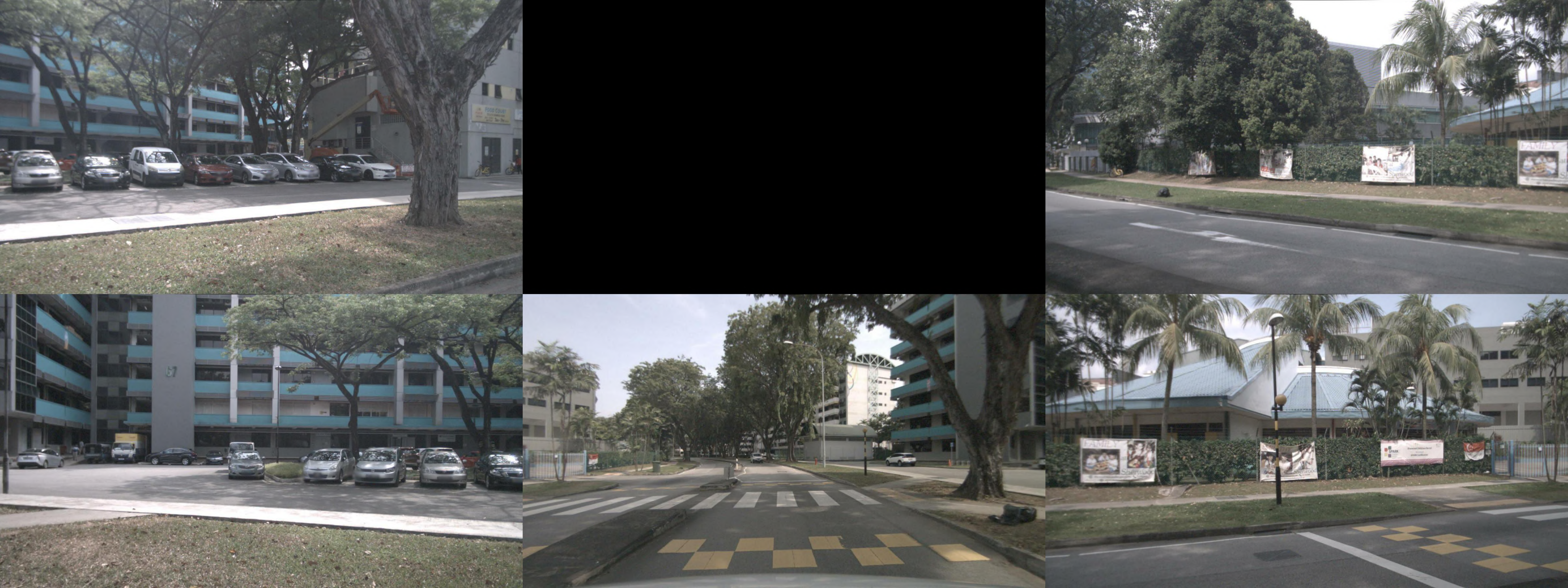}
		\subcaption{Corrupted Image}
	\end{minipage}
	\begin{minipage}{0.22\textwidth}		\includegraphics[width=0.9\linewidth]{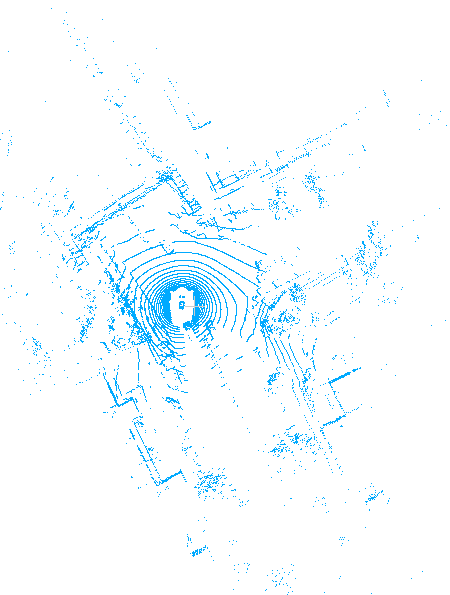}
		\subcaption{Normal Point Cloud}
	\end{minipage}
	\hfill
	\begin{minipage}{0.22\textwidth}
		\includegraphics[width=0.9\linewidth]{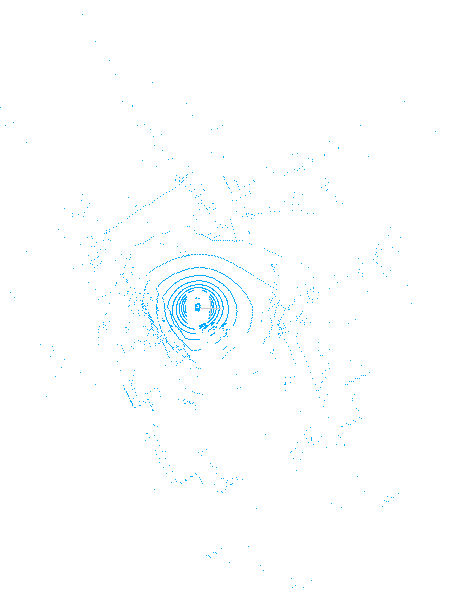}
		\subcaption{Corrupted Point Cloud}
	\end{minipage}
	\caption{The effects of normal and corrupted sensor data, which are used to train the RobuAlign model for enhanced robustness.}
	\label{fig:robualign_fig3}
\end{figure}

\begin{figure*}[t]
    \centering
    \includegraphics[width=\linewidth]{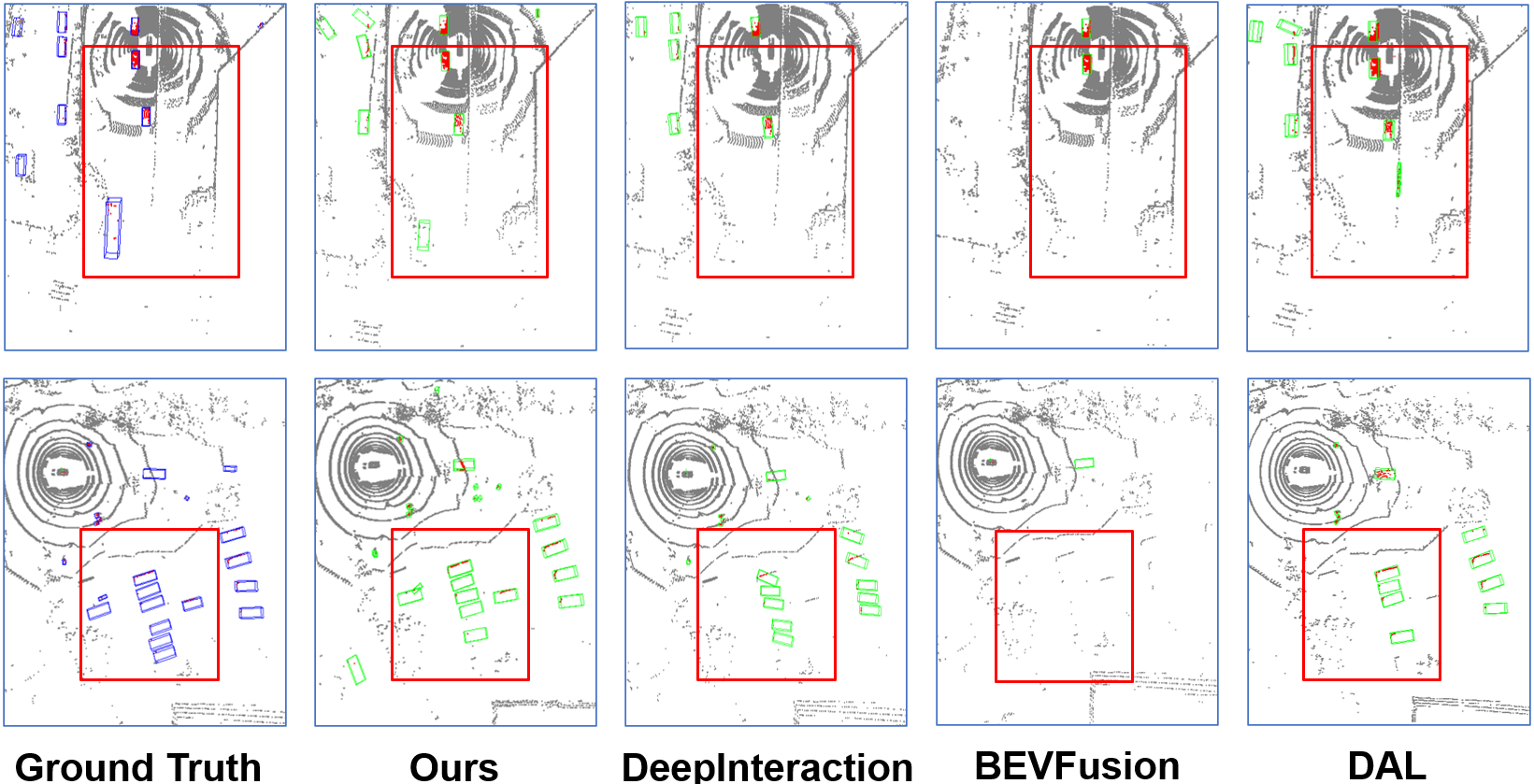}
    \caption{Comparative analysis of RobuAlign with other SOTA methods under sensor failure scenarios, highlighting superior detection accuracy.}
    \label{fig:robualign_fig4}
\end{figure*}

\subsubsection{Team \textcolor{robo_blue}{Ponyville Autonauts Ltd.}}
This team developed the \textbf{Cross Modal Transformer (CMT)} for the multi-modal BEV detection under scenarios that are critical to the perception robustness. This framework enhances the resilience of the baseline multi-modal 3D object detection systems under sensor failure conditions, particularly focusing on scenarios where camera or LiDAR sensors may be compromised. CMT utilizes a new approach to integrate data from both camera and LiDAR sensors without explicit view transformations, thus directly outputting accurate 3D bounding boxes.

\noindent\textbf{\faLightbulbO~Key Innovations:}
\begin{itemize}
    \item \textbf{Unified Multi-Modal Tokenization:} CMT processes images and point cloud data by converting them into a unified token format, allowing for seamless integration without the need for explicit view transformations.
    \item \textbf{Self-Supervised Feature Compensation:} The model uses self-supervised learning techniques to compensate for missing or corrupted sensor data, which has proven to enhance robustness under sensor failures.
    \item \textbf{Position-Guided Query Generator:} Inspired by advances in anchor-based detection systems, this feature generates queries based on 3D anchor points projected across modalities, improving the accuracy and efficiency of bounding box predictions.
\end{itemize}

\noindent\textbf{\faGear~Implementation Details:}\\
The CMT architecture employs dual backbones to extract features from ring-view camera images and LiDAR point clouds. These features are encoded into multi-modal tokens through innovative coordinate encoding, which optimizes the integration of 2D image depth with 3D LiDAR data. This integration occurs in a Transformer decoder where position-guided queries, derived from predefined anchor points, interact with the tokens to enhance detection accuracy by focusing on probable object locations.
The decoder refines these interactions to output precise object classifications and 3D bounding box coordinates, utilizing self-attention and cross-attention mechanisms to handle the complexities of data fusion. This setup is crucial for maintaining high performance even when sensors fail, as the model is trained and tested across a variety of simulated sensor failure scenarios. These simulations ensure that CMT can adapt to partial or complete loss of sensory data, maintaining reliable object detection in adverse conditions.
The robust training regimen, paired with the model's capacity to compensate for sensor inadequacies, is demonstrated in how it effectively handles real-world driving scenarios, significantly improving its utility for autonomous driving applications under challenging conditions.

\subsubsection{Team \textcolor{robo_blue}{HITSZrobodrive}}
Team \textcolor{robo_blue}{HITSZrobodrive} introduced the \textbf{RobuAlign} model in the Robust Multi-Modal BEV Detection track. This approach is crafted to ensure effective 3D object detection even under sensor failures, using a sophisticated corruption simulation strategy to robustly process multi-modal data.

\noindent\textbf{\faLightbulbO~Key Innovations:}
\begin{itemize}
    \item \textbf{Corruption Simulation Strategy:} RobuAlign simulates real-world sensor failures to train the model to maintain high performance despite data degradation.
    \item \textbf{Robust Multi-Modal Fusion:} The model employs a novel architecture to harmonize image and point cloud data, ensuring seamless integration and robust detection across varying sensor conditions.
    \item \textbf{Adaptive Data Augmentation:} Dynamic augmentation techniques are utilized to enhance the model’s resilience against unpredictable sensor failures and environmental challenges.
\end{itemize}

\noindent\textbf{\faGear~Implementation Details:}\\
RobuAlign utilizes an advanced dual architecture, utilizing ResNet \cite{he2016deep} and Swin Transformer \cite{liu2021swin} for processing image data and VoxelNet \cite{voxelnet} for LiDAR data extraction. This setup allows for robust feature extraction from both sensor modalities, which is essential for accurate object detection in adverse conditions. The model simulates sensor failures during preprocessing by masking images and downsampling LiDAR data (\cref{fig:robualign_fig1}). This corruption simulation trains the system to adapt to and compensate for real-world sensor failures.
To integrate these features, RobuAlign uses a multi-modal fusion module that merges image and LiDAR inputs into a cohesive BEV output, enhancing detection accuracy across varied sensor conditions (\cref{fig:robualign_fig2}). The adaptive data augmentation methods applied include dynamic masking of images and strategic LiDAR downsampling, which are essential for training the model to handle unexpected sensor outages and data corruption (\cref{fig:robualign_fig3}).
In addition to the robust fusion capabilities, the system also employs a novel image-to-BEV conversion technique using depth maps generated from the LiDAR data. This method ensures that the spatial relationships and topological structures are accurately maintained even when the input data quality is compromised. The effective handling of corrupted sensor data is demonstrated through rigorous testing under simulated conditions that mimic real-world challenges, showing a significant improvement in model performance and resilience compared to conventional methods.
The robust training and validation strategy includes extensive testing under simulated conditions of sensor failures, demonstrating the model's ability to maintain high accuracy and reliability. This is critical for applications in autonomous driving, where sensor reliability is crucial (\cref{fig:robualign_fig4}).

%% file: sections/5_conclusion.tex
\section{Discussions \& Future Directions}
\label{sec:discussions}

The 2024 RoboDrive Challenge has not only showcased the current state-of-the-art in autonomous driving perception technologies but also highlighted key areas for future research and development. This challenge has revealed both the robustness and the limitations of existing approaches under real-world and out-of-distribution conditions. The innovations presented by participants point toward several promising directions that could shape the future of autonomous vehicle technologies.

\textit{Integration of Advanced Sensor Technologies:}
A critical insight from this year’s challenge is the potential to enhance system performance through the adept integration of multi-sensor data. This is especially crucial under scenarios of sensor impairment. Future research could delve deeper into sophisticated sensor fusion techniques, particularly emphasizing the integration of underexploited modalities like radar and thermal imaging. Enhancing these fusion algorithms could substantially improve the operational reliability and accuracy of autonomous systems in adverse environmental conditions.

\textit{Advancements in Machine Learning Techniques:}
The challenge demonstrated the utility of self-supervised and semi-supervised learning approaches in correcting errors and augmenting features without extensive labeled datasets. Future endeavors could expand on the use of unsupervised learning techniques to further alleviate the constraints imposed by the scarcity of annotated data, potentially revolutionizing the efficiency of model training processes in autonomous driving applications.

\textit{Enhancing Algorithmic Robustness:} 
There is a pronounced need for developing algorithms that swiftly adapt to sudden environmental shifts and sensor degradation. Upcoming challenges could focus on crafting and evaluating algorithms tailored for dynamic, real-time adjustments to foster greater adaptability in autonomous driving systems.

\textit{Standardization and Benchmarking:} 
The diversity of methodologies and outcomes highlighted by the challenge underscores the necessity for standardized testing environments that more closely mirror the complexities encountered in real-world driving scenarios. Establishing rigorous benchmarks and uniform testing protocols would ensure that models are not only tested under controlled conditions but are also adept at navigating the unpredictable nuances of real-world operations.

\textit{Ethical and Safety Considerations:}
As autonomous technologies continue to evolve, addressing the ethical implications of their deployment in public spaces becomes imperative. Future research should prioritize the development of comprehensive safety protocols and ethical guidelines to govern the interactions between autonomous vehicles and human-operated vehicles. This will be crucial in managing mixed-traffic environments safely and ethically, ensuring that the integration of autonomous vehicles enhances overall road safety and efficiency.

The insights gained from this challenge are instrumental in directing future research efforts, setting new benchmarks in the field, and fostering a competitive yet collaborative environment for technological innovation. As we move forward, it is essential to continue pushing the boundaries of what is possible in autonomous driving technologies while ensuring these advancements align with broader societal and ethical standards.

\section{Conclusion}
\label{sec:conclusion}

The 2024 RoboDrive Challenge has significantly contributed to advancing the field of autonomous driving technologies, particularly in the domain of robust perception under diverse and challenging conditions. The competition has brought together some of the brightest minds from across the globe, fostering a collaborative environment that has spurred innovation and pushed the boundaries of what is possible in autonomous driving.
The solutions presented during the challenge have highlighted the critical importance of robustness and adaptability in autonomous vehicle systems, particularly the need for effective sensor fusion, advanced data augmentation, and innovative learning techniques. The insights gained from this challenge are expected to influence ongoing research and development, guiding the future of autonomous driving towards safer, more reliable, and more efficient systems.
As the field continues to evolve, it is clear that the journey toward fully autonomous driving is as much about technological innovation as it is about collaboration and continuous learning. The RoboDrive Challenge stands as a testament to the power of competitive innovation in driving progress, setting new standards for what can be achieved in the quest for autonomous mobility.
By drawing on the successes and learnings from this year's challenge, future iterations will undoubtedly continue to refine and redefine the capabilities of autonomous driving technologies, steering us toward a safer and more connected world.

\section*{Acknowledgments}
This work is under the programme DesCartes and is supported by the National Research Foundation, Prime Minister’s Office, Singapore, under its Campus for Research Excellence and Technological Enterprise (CREATE) programme. This work is also supported by the Ministry of Education, Singapore, under its MOE AcRF Tier 2 (MOET2EP20221- 0012), NTU NAP, and under the RIE2020 Industry Alignment Fund – Industry Collaboration Projects (IAF-ICP) Funding Initiative, as well as cash and in-kind contribution from the industry partner(s).

%% file: sections/6_appendix.tex
\section{Appendix}
\label{sec:appendix}
In this appendix, we summarize the public resources of this challenge (\cf \cref{sec:public_resources}), and the information of 1) the challenge organizers (\cf \cref{sec:organizers}), 2) the technical committee members (\cf \cref{sec:technical_committee}), as well as 3) the challenge participants from each track (\cf \cref{sec:teams_affiliations}). We also include the track report from each winning team (\cf \cref{sec:track_reports}).

\subsection{Public Resources}
\label{sec:public_resources}
\begin{itemize}
    \item \textbf{Website:}\\
    \url{https://robodrive-24.github.io}

    \item \textbf{Technical Report:}\\
    \url{https://arxiv.org/abs/2405.08816}
    
    \item \textbf{Workshop Recording (YouTube):}\\
    \url{https://youtu.be/83_FxdXNNQk}

    \item \textbf{Workshop Recording (Bilibili):}\\
    \url{https://www.bilibili.com/video/BV11H4y1M7mp}

    \item \textbf{Workshop Slides:}\\
    \url{https://robodrive-24.github.io/workshop.pdf}

    \item \textbf{Evaluation Server (Track 1):}\\
    \url{https://codalab.lisn.upsaclay.fr/competitions/17135}

    \item \textbf{Evaluation Server (Track 2):}\\
    \url{https://codalab.lisn.upsaclay.fr/competitions/17062}

    \item \textbf{Evaluation Server (Track 3):}\\
    \url{https://codalab.lisn.upsaclay.fr/competitions/17063}

    \item \textbf{Evaluation Server (Track 4):}\\
    \url{https://codalab.lisn.upsaclay.fr/competitions/17226}

    \item \textbf{Evaluation Server (Track 5):}\\
    \url{https://codalab.lisn.upsaclay.fr/competitions/17137}
    
\end{itemize}

\subsection{Organizers}
\label{sec:organizers}
\begin{itemize}
    \item \textbf{Team Members:}\\
    Lingdong Kong$^{1,2,3}$, Shaoyuan Xie$^4$, Hanjiang Hu$^5$, Yaru Niu$^5$, Wei Tsang Ooi$^2$, Benoit R. Cottereau$^{6,7}$, Lai Xing Ng$^8$, Yuexin Ma$^9$, Wenwei Zhang$^1$, Liang Pan$^1$, Kai Chen$^1$, and Ziwei Liu$^{10}$

    \item \noindent\textbf{Affiliations:}\\
    $^1$Shanghai AI Laboratory\\$^2$National University of Singapore\\$^3$CNRS@CREATE\\$^4$University of California, Irvine\\$^5$Carnegie Mellon University\\$^6$IPAL, CNRS IRL 2955, Singapore\\$^7$CerCo, CNRS UMR 5549, Université Toulouse III\\$^8$Institute for Infocomm Research, A*STAR\\$^9$ShanghaiTech University\\$^{10}$Nanyang Technological University, Singapore

    \item \noindent\textbf{Website:}\\
    \url{https://robodrive-24.github.io}
\end{itemize}

\subsection{Technical Committee}
\label{sec:technical_committee}
\begin{itemize}
    \item \textbf{Team Members:}\\
    Weichao Qiu$^1$ and Wei Zhang$^1$

    \item \noindent\textbf{Affiliations:}\\
    $^1$HUAWEI Noah's Ark Lab

    \item \noindent\textbf{Website:}\\
    \url{https://www.noahlab.com.hk}
\end{itemize}

\subsection{Teams \& Affiliations}
\label{sec:teams_affiliations}

\subsubsection{Track 1: Robust BEV Detection}

\paragraph{Team~ \textcolor{robo_blue}{DeepVision}}
\begin{itemize}
    \item \textbf{Team Members:}\\
    Xu Cao$^{1,2}$, Hao Lu$^{1,2}$, and Ying-Cong Chen$^{1,2}$
    \item \textbf{Email:}\\ \url{xcao635@connect.hkust-gz.edu.cn}
    \item \textbf{Affiliations:}\\
    $^1$The Hong Kong University of Science and Technology (Guangzhou)\\$^2$The Hong Kong University of Science and Technology
\end{itemize}

\paragraph{Team~ \textcolor{robo_blue}{Ponyville Autonauts Ltd.}}
\begin{itemize}
    \item \textbf{Team Members:}\\
    Caixin Kang$^1$, Xinning Zhou$^2$, Chengyang Ying$^2$, Wentao Shang$^3$, Xingxing Wei$^1$, and Yinpeng Dong$^2$
    \item \textbf{Email:}\\ \url{timkang666@gmail.com}
    \item \textbf{Affiliations:}\\
    $^1$Beihang University\\$^2$Tsinghua University\\$^3$Hefei University of Technology
\end{itemize}

\paragraph{Team~ \textcolor{robo_blue}{CyberBEV}}
\begin{itemize}
    \item \textbf{Team Members:}\\
    Bo Yang$^1$, Shengyin Jiang$^1$, Zeliang Ma$^1$, Dengyi Ji$^2$, and Haiwen Li$^1$
    \item \textbf{Email:}\\ \url{bobyang677@gmail.com}
    \item \textbf{Affiliations:}\\
    $^1$Beijing University of Posts and Telecommunications\\$^2$Beijing University of Technology
\end{itemize}

\subsubsection{Track 2: Robust Map Segmentation}

\paragraph{Team~ \textcolor{robo_blue}{SafeDrive-SSR}}
\begin{itemize}
    \item \textbf{Team Members:}\\
    Xingliang Huang$^1$ and Yu Tian$^2$
    \item \textbf{Email:}\\
    \url{huangxingliang20@mails.ucas.ac.cn}
    \item \textbf{Affiliations:}\\
    $^1$University of Chinese Academy of Sciences\\$^2$Tsinghua University
\end{itemize}

\paragraph{Team~ \textcolor{robo_blue}{CrazyFriday}}
\begin{itemize}
    \item \textbf{Team Members:}\\
    Genghua Kou$^1$, Fan Jia$^2$, Yingfei Liu$^2$, Tiancai Wang$^2$, and Ying Li$^1$
    \item \textbf{Email:}\\
    \url{koughua@bit.edu.cn}
    \item \textbf{Affiliations:}\\
    $^1$Beijing Institute of Technology\\$^2$Megvii Technology
\end{itemize}

\paragraph{Team~ \textcolor{robo_blue}{Samsung Research China-Advanced Research}}
\begin{itemize}
    \item \textbf{Team Members:}\\
    Xiaoshuai Hao$^1$, Yifan Yang$^1$, Hui Zhang$^1$, Mengchuan Wei$^1$, Yi Zhou$^1$,  Haimei Zhao$^2$, and Jing Zhang$^2$
    \item \textbf{Email:}\\
    \url{xshuai.hao@samsung.com}
    \item \textbf{Affiliations:}\\
    $^1$Samsung R\&D Institute China–Beijing\\$^2$The University of Sydney
\end{itemize}

\subsubsection{Track 3: Robust Occupancy Prediction}

\paragraph{Team~ \textcolor{robo_blue}{ViewFormer}}
\begin{itemize}
    \item \textbf{Team Members:}\\
    Jinke Li$^1$, Xiao He$^1$, and Xiaoqiang Cheng$^1$
    \item \textbf{Email:}\\
    \url{jinke.li@uisee.com}
    \item \textbf{Affiliations:}\\
    $^1$UISEE Foundation Research \& Development
\end{itemize}

\paragraph{Team~ \textcolor{robo_blue}{APEC Blue}}
\begin{itemize}
    \item \textbf{Team Members:}\\
    Bingyang Zhang$^1$, Lirong Zhao$^2$, Dianlei Ding$^1$, Fangsheng Liu$^1$, Yixiang Yan$^1$, and Hongming Wang$^1$
    \item \textbf{Email:}\\
    \url{zhangbingyang1@foxmail.com}
    \item \textbf{Affiliations:}\\
    $^1$Beijing APEC Blue Technology Co., Ltd\\$^2$Beihang University
\end{itemize}

\paragraph{Team~ \textcolor{robo_blue}{hm.unilab}}
\begin{itemize}
    \item \textbf{Team Members:}\\
    Nanfei Ye$^1$, Lun Luo$^1$, Yubo Tian$^1$, Yiwei Zuo$^1$, Zhe Cao$^1$, Yi Ren$^1$, Yunfan Li$^1$, Wenjie Liu$^1$, and Xun Wu$^1$
    \item \textbf{Email:}\\
    \url{hm.unilab@gmail.com}
    \item \textbf{Affiliations:}\\
    $^1$Haomo.ai
\end{itemize}

\subsubsection{Track 4: Robust Depth Estimation}

\paragraph{Team~ \textcolor{robo_blue}{HIT-AIIA}}
\begin{itemize}
    \item \textbf{Team Members:}\\
    Yifan Mao$^1$, Ming Li$^1$, Jian Liu$^1$, Jiayang Liu$^1$, Zihan Qin$^1$, Cunxi Chu$^1$, Jialei Xu$^1$, Wenbo Zhao$^1$, Junjun Jiang$^1$, and Xianming Liu$^1$
    \item \textbf{Email:}\\
    \url{maoyf1105@163.com}
    \item \textbf{Affiliations:}\\
    $^1$Harbin Institute of Technology
\end{itemize}

\paragraph{Team~ \textcolor{robo_blue}{BUAA-Trans}}
\begin{itemize}
    \item \textbf{Team Members:}\\
    Ziyan Wang$^1$, Chiwei Li$^1$, Shilong Li$^1$, Chendong Yuan$^1$, Songyue Yang$^1$, Wentao Liu$^1$, Peng Chen$^1$, and Bin Zhou$^1$
    \item \textbf{Email:}\\
    \url{cpeng@buaa.edu.cn}
    \item \textbf{Affiliations:}\\
    $^1$Beihang University
\end{itemize}

\paragraph{Team~ \textcolor{robo_blue}{CUSTZS}}
\begin{itemize}
    \item \textbf{Team Members:}\\
    Yubo Wang$^1$, Chi Zhang$^1$, and Jianhang Sun$^1$
    \item \textbf{Email:}\\
    \url{w1055025523@gmail.com}
    \item \textbf{Affiliations:}\\
    $^1$Changchun University of Science and Technology
\end{itemize}

\subsubsection{Track 5: Robust Multi-Modal BEV Detection}

\paragraph{Team~ \textcolor{robo_blue}{safedrive-promax}}
\begin{itemize}
    \item \textbf{Team Members:}\\
    Hai Chen$^1$, Xiao Yang$^1$, and Lizhong Wang$^1$
    \item \textbf{Email:}\\ \url{chber\_ahu@hotmail.com}
    \item \textbf{Affiliations:}\\
    $^1$Tsinghua University
\end{itemize}

\paragraph{Team~ \textcolor{robo_blue}{Ponyville Autonauts Ltd.}}
\begin{itemize}
    \item \textbf{Team Members:}\\
    Caixin Kang$^1$, Xinning Zhou$^2$, Chengyang Ying$^2$, Wentao Shang$^3$, Xingxing Wei$^1$, and Yinpeng Dong$^2$
    \item \textbf{Email:}\\ \url{timkang666@gmail.com}
    \item \textbf{Affiliations:}\\
    $^1$Beihang University\\
    $^2$Tsinghua University\\
    $^3$Hefei University of Technology
\end{itemize}

\paragraph{Team~ \textcolor{robo_blue}{HITSZrobodrive}}
\begin{itemize}
    \item \textbf{Team Members:}\\
    Dongyi Fu$^1$, Yongchun Lin$^1$, Huitong Yang$^2$, Haoang Li$^3$, Yadan Luo$^4$, Xianjing Cheng$^1$, and Yong Xu$^1$
    \item \textbf{Email:}\\
    \url{23s151124@stu.hit.edu.cn}
    \item \textbf{Affiliations:}\\
    $^1$Harbin Institute of Technology\\$^2$Guangdong University of Technology\\$^3$The Hong Kong University of Science and Technology (Guangzhou)\\$^4$The University of Queensland
\end{itemize}

\subsection{Track Reports}
\label{sec:track_reports}

\begin{itemize}
    \item Xu Cao, Hao Lu, and Ying-Cong Chen. ``Towards Robust Multi-Camera 3D Object Detection through Temporal Sequence Mix Augmentation'', Technical Report, 2024.

    \item Caixin Kang, Xinning Zhou, Chengyang Ying, Wentao Shang, Xingxing Wei, and Yinpeng Dong. ``MVE: Multi-View Enhancer for Robust Bird's Eye View Object Detection'', Technical Report, 2024.

    \item Bo Yang, Shengyin Jiang, Zeliang Ma, Dengyi Ji, and Haiwen Li. ``FocalAngle3D: An Angle-Enhanced Two-Stage Model for 3D Detection'', Technical Report, 2024.

    \item Xingliang Huang and Yu Tian. ``Models and Data Enhancements for Robust Map Segmentation in Autonomous Driving'', Technical Report, 2024.

    \item Xiaoshuai Hao, Yifan Yang, Hui Zhang, Mengchuan Wei, Yi Zhou, Haimei Zhao, and Jing Zhang. ``Using Temporal Information and Mixing-Based Data Augmentations for Robust HD Map Construction'', Technical Report, 2024.

    \item Genghua Kou, Fan Jia, Yingfei Liu, Tiancai Wang, and Ying Li. ``MultiViewRobust: Scaling Up Pretrained Models for Robust Map Segmentation'', Technical Report, 2024.

    \item Jinke Li, Xiao He, and Xiaoqiang Cheng. ``ViewFormer: Spatiotemporal Modeling for Robust Occupancy Prediction'', Technical Report, 2024.

    \item Bingyang Zhang, Lirong Zhao, Dianlei Ding, Fangsheng Liu, Yixiang Yan, and Hongming Wang. ``Robust Occupancy Prediction based on Enhanced SurroundOcc'', Technical Report, 2024.

    \item Nanfei Ye, Lun Luo, Xun Wu, Yubo Tian, Zhe Cao, Yunfan Li, Yiwei Zuo, Wenjie Liu, and Yi Ren. ``Improving Out-of-Distribution Robustness of Occupancy Prediction Networks with Advanced Loss Functions'', Technical Report, 2024.

    \item Yifan Mao, Ming Li, Jian Liu, Jiayang Liu, Zihan Qin, Chunxi Chu, Jialei Xu, Wenbo Zhao, Junjun Jiang, and Xianming Liu. ``DINO-SD for Robust Multi-View Supervised Depth Estimation'', Technical Report, 2024.

    \item Ziyan Wang, Chiwei Li, Shilong Li, Chendong Yuan, Songyue Yang, Wentao Liu, Peng Chen, and Bin Zhou. ``Fusing Features Across Scales: A Semi-Supervised Attention-Based Approach for Robust Depth Estimation'', Technical Report, 2024.

    \item Yubo Wang, Chi Zhang, and Jianhang Sun. ``SD-ViT: Performance and Robustness Enhancements of MonoViT for Multi-View Depth Estimation'', Technical Report, 2024.

    \item Hai Chen, Xiao Yang, and Lizhong Wang. ``ASF: Robust 3D Object Detection by Solving Sensor Failures'', Technical Report, 2024.

    \item Caixin Kang, Xinning Zhou, Chengyang Ying, Wentao Shang, Xingxing Wei, and Yinpeng Dong. ``Cross-Modal Transformers for Robust Multi-Modal BEV Detection'', Technical Report, 2024.

    \item Dongyi Fu, Yongchun Lin, Huitong Yang, Haoang Li, Yadan Luo, Xianjing Cheng, and Yong Xu. ``RobuAlign: Robust Alignment in Multi-Modal 3D Object Detection'', Technical Report, 2024.
\end{itemize}

%% file: main.bbl
\begin{thebibliography}{95}
\providecommand{\natexlab}[1]{#1}
\providecommand{\url}[1]{\texttt{#1}}
\expandafter\ifx\csname urlstyle\endcsname\relax
  \providecommand{\doi}[1]{doi: #1}\else
  \providecommand{\doi}{doi: \begingroup \urlstyle{rm}\Url}\fi

\bibitem[Barbu et~al.(2019)Barbu, Mayo, Alverio, Luo, Wang, Gutfreund, Tenenbaum, and Katz]{barbu2019objectnet}
Andrei Barbu, David Mayo, Julian Alverio, William Luo, Christopher Wang, Dan Gutfreund, Josh Tenenbaum, and Boris Katz.
\newblock Objectnet: A large-scale bias-controlled dataset for pushing the limits of object recognition models.
\newblock \emph{Advances in Neural Information Processing Systems}, 32, 2019.

\bibitem[Beemelmanns et~al.(2024)Beemelmanns, Zhang, and Eckstein]{beemelmanns2024multicorrupt}
Till Beemelmanns, Quan Zhang, and Lutz Eckstein.
\newblock Multicorrupt: A multi-modal robustness dataset and benchmark of lidar-camera fusion for 3d object detection.
\newblock \emph{arXiv preprint arXiv:2402.11677}, 2024.

\bibitem[Caesar et~al.(2020)Caesar, Bankiti, Lang, Vora, Liong, Xu, Krishnan, Pan, Baldan, and Beijbom]{caesar2020nuscenes}
Holger Caesar, Varun Bankiti, Alex~H Lang, Sourabh Vora, Venice~Erin Liong, Qiang Xu, Anush Krishnan, Yu Pan, Giancarlo Baldan, and Oscar Beijbom.
\newblock nuscenes: A multimodal dataset for autonomous driving.
\newblock In \emph{IEEE/CVF Conference on Computer Vision and Pattern Recognition}, pages 11621--11631, 2020.

\bibitem[Cao and De~Charette(2022)]{cao2022monoscene}
Anh-Quan Cao and Raoul De~Charette.
\newblock Monoscene: Monocular 3d semantic scene completion.
\newblock In \emph{IEEE/CVF Conference on Computer Vision and Pattern Recognition}, pages 3991--4001, 2022.

\bibitem[Chen et~al.(2023)Chen, Liu, Kong, Zhu, Ma, Li, Hou, Qiao, and Wang]{chen2023clip2Scene}
Runnan Chen, Youquan Liu, Lingdong Kong, Xinge Zhu, Yuexin Ma, Yikang Li, Yuenan Hou, Yu Qiao, and Wenping Wang.
\newblock Clip2scene: Towards label-efficient 3d scene understanding by clip.
\newblock In \emph{IEEE/CVF Conference on Computer Vision and Pattern Recognition}, pages 7020--7030, 2023.

\bibitem[Cheng et~al.(2024)Cheng, Yin, Wang, Chen, Wang, and Yang]{AFNet}
Junda Cheng, Wei Yin, Kaixuan Wang, Xiaozhi Chen, Shijie Wang, and Xin Yang.
\newblock Adaptive fusion of single-view and multi-view depth for autonomous driving.
\newblock \emph{arXiv preprint arXiv:2403.07535}, 2024.

\bibitem[Choy et~al.(2019)Choy, Gwak, and Savarese]{choy2019minkowski}
Christopher Choy, JunYoung Gwak, and Silvio Savarese.
\newblock 4d spatio-temporal convnets: Minkowski convolutional neural networks.
\newblock In \emph{IEEE/CVF Conference on Computer Vision and Pattern Recognition}, pages 3075--3084, 2019.

\bibitem[Deng et~al.(2009)Deng, Dong, Socher, Li, Li, and Fei-Fei]{imagenet}
Jia Deng, Wei Dong, Richard Socher, Li-Jia Li, Kai Li, and Li Fei-Fei.
\newblock Imagenet: A large-scale hierarchical image database.
\newblock In \emph{IEEE/CVF Conference on Computer Vision and Pattern Recognition}, pages 248--255, 2009.

\bibitem[DeVries and Taylor.(2017)]{devries2017cutout}
Terrance DeVries and Graham~W. Taylor.
\newblock Improved regularization of convolutional neural networks with cutout.
\newblock \emph{arXiv preprint arXiv:1708.04552}, 2017.

\bibitem[Dub{\'e} et~al.(2017)Dub{\'e}, Gawel, Sommer, Nieto, Siegwart, and Cadena]{dube2017online}
Renaud Dub{\'e}, Abel Gawel, Hannes Sommer, Juan Nieto, Roland Siegwart, and Cesar Cadena.
\newblock An online multi-robot slam system for 3d lidars.
\newblock In \emph{IEEE/RSJ International Conference on Intelligent Robots and Systems}, pages 1004--1011, 2017.

\bibitem[Eigen et~al.(2014)Eigen, Puhrsch, and Fergus]{eigen2014depth}
David Eigen, Christian Puhrsch, and Rob Fergus.
\newblock Depth map prediction from a single image using a multi-scale deep network.
\newblock In \emph{Advances in Neural Information Processing Systems}, 2014.

\bibitem[Fang et~al.(2023)Fang, Wang, Xie, Sun, Wu, Wang, Huang, Wang, and Cao]{fang2023eva}
Yuxin Fang, Wen Wang, Binhui Xie, Quan Sun, Ledell Wu, Xinggang Wang, Tiejun Huang, Xinlong Wang, and Yue Cao.
\newblock Eva: Exploring the limits of masked visual representation learning at scale.
\newblock In \emph{IEEE/CVF Conference on Computer Vision and Pattern Recognition}, pages 19358--19369, 2023.

\bibitem[Gaidon et~al.(2021)Gaidon, Shakhnarovich, Ambrus, Guizilini, Vasiljevic, Walter, Pillai, and Kolkin]{DDAD}
Adrien Gaidon, Greg Shakhnarovich, Rares Ambrus, Vitor Guizilini, Igor Vasiljevic, Matthew Walter, Sudeep Pillai, and Nick Kolkin.
\newblock The dense depth for autonomous driving (ddad) challenge.
\newblock \url{https://sites.google.com/view/mono3d-workshop}, 2021.

\bibitem[Ge et~al.(2021)Ge, Liu, Wang, Li, and Sun]{ge2021yolox}
Zheng Ge, Songtao Liu, Feng Wang, Zeming Li, and Jian Sun.
\newblock Yolox: Exceeding yolo series in 2021.
\newblock \emph{arXiv preprint arXiv:2107.08430}, 2021.

\bibitem[Geiger et~al.(2013)Geiger, Lenz, Stiller, and Urtasun]{geiger2013vision}
Andreas Geiger, Philip Lenz, Christoph Stiller, and Raquel Urtasun.
\newblock Vision meets robotics: The kitti dataset.
\newblock \emph{The International Journal of Robotics Research}, 32\penalty0 (11):\penalty0 1231--1237, 2013.

\bibitem[Godard et~al.(2019)Godard, Aodha, Firman, and Brostow]{godard2019monodepth2}
Clément Godard, Oisin~Mac Aodha, Michael Firman, and Gabriel~J. Brostow.
\newblock Digging into self-supervised monocular depth prediction.
\newblock In \emph{IEEE/CVF International Conference on Computer Vision}, pages 3828--3838, 2019.

\bibitem[Guo et~al.(2023)Guo, Yuan, Zhang, Yang, Zhang, Zhu, and Chen]{S3Depth}
Xianda Guo, Wenjie Yuan, Yunpeng Zhang, Tian Yang, Chenming Zhang, Zheng Zhu, and Long Chen.
\newblock A simple baseline for supervised surround-view depth estimation.
\newblock \emph{arXiv preprint arXiv:2303.07759}, 2023.

\bibitem[He et~al.(2016)He, Zhang, Ren, and Sun]{he2016deep}
Kaiming He, Xiangyu Zhang, Shaoqing Ren, and Jian Sun.
\newblock Deep residual learning for image recognition.
\newblock In \emph{IEEE/CVF Conference on Computer Vision and Pattern Recognition}, pages 770--778, 2016.

\bibitem[Hendrycks and Dietterich(2019)]{hendrycks2019benchmarking}
Dan Hendrycks and Thomas Dietterich.
\newblock Benchmarking neural network robustness to common corruptions and perturbations.
\newblock \emph{arXiv preprint arXiv:1903.12261}, 2019.

\bibitem[Hendrycks et~al.(2019)Hendrycks, Mu, Cubuk, Zoph, Gilmer, and Lakshminarayanan]{hendrycks2019augmix}
Dan Hendrycks, Norman Mu, Ekin~D Cubuk, Barret Zoph, Justin Gilmer, and Balaji Lakshminarayanan.
\newblock Augmix: A simple data processing method to improve robustness and uncertainty.
\newblock \emph{arXiv preprint arXiv:1912.02781}, 2019.

\bibitem[Hendrycks et~al.(2021)Hendrycks, Basart, Mu, Kadavath, Wang, Dorundo, Desai, Zhu, Parajuli, Guo, et~al.]{hendrycks2021many}
Dan Hendrycks, Steven Basart, Norman Mu, Saurav Kadavath, Frank Wang, Evan Dorundo, Rahul Desai, Tyler Zhu, Samyak Parajuli, Mike Guo, et~al.
\newblock The many faces of robustness: A critical analysis of out-of-distribution generalization.
\newblock In \emph{IEEE/CVF International Conference on Computer Vision}, pages 8340--8349, 2021.

\bibitem[Hong et~al.(2024)Hong, Kong, Zhou, Zhu, Li, and Liu]{hong20224dDSNet}
Fangzhou Hong, Lingdong Kong, Hui Zhou, Xinge Zhu, Hongsheng Li, and Ziwei Liu.
\newblock Unified 3d and 4d panoptic segmentation via dynamic shifting networks.
\newblock \emph{IEEE Transactions on Pattern Analysis and Machine Intelligence}, 46\penalty0 (5):\penalty0 3480--3495, 2024.

\bibitem[Hu et~al.(2022)Hu, Yang, Qiao, Liu, Zhao, and Wang]{hu2022seasondepth}
Hanjiang Hu, Baoquan Yang, Zhijian Qiao, Shiqi Liu, Ding Zhao, and Hesheng Wang.
\newblock Seasondepth: Cross-season monocular depth prediction dataset and benchmark under multiple environments.
\newblock In \emph{International Conference on Machine Learning Workshops}, 2022.

\bibitem[Huang and Huang(2022)]{huang2022bevdet4d}
Junjie Huang and Guan Huang.
\newblock Bevdet4d: Exploit temporal cues in multi-camera 3d object detection.
\newblock \emph{arXiv preprint arXiv:2203.17054}, 2022.

\bibitem[Huang et~al.(2021)Huang, Huang, Zhu, Ye, and Du]{huang2021bevdet}
Junjie Huang, Guan Huang, Zheng Zhu, Yun Ye, and Dalong Du.
\newblock Bevdet: High-performance multi-camera 3d object detection in bird-eye-view.
\newblock \emph{arXiv preprint arXiv:2112.11790}, 2021.

\bibitem[Huang et~al.(2023)Huang, Zheng, Zhang, Zhou, and Lu]{huang2023tri}
Yuanhui Huang, Wenzhao Zheng, Yunpeng Zhang, Jie Zhou, and Jiwen Lu.
\newblock Tri-perspective view for vision-based 3d semantic occupancy prediction.
\newblock In \emph{IEEE/CVF Conference on Computer Vision and Pattern Recognition}, pages 9223--9232, 2023.

\bibitem[Jiang et~al.(2023)Jiang, Zhang, Miao, Zhu, Gao, Hu, and Jiang]{jiang2023polarformer}
Yanqin Jiang, Li Zhang, Zhenwei Miao, Xiatian Zhu, Jin Gao, Weiming Hu, and Yu-Gang Jiang.
\newblock Polarformer: Multi-camera 3d object detection with polar transformer.
\newblock In \emph{AAAI Conference on Artificial Intelligence}, pages 1042--1050, 2023.

\bibitem[Kong et~al.(2023{\natexlab{a}})Kong, Liu, Chen, Ma, Zhu, Li, Hou, Qiao, and Liu]{kong2023rethinking}
Lingdong Kong, Youquan Liu, Runnan Chen, Yuexin Ma, Xinge Zhu, Yikang Li, Yuenan Hou, Yu Qiao, and Ziwei Liu.
\newblock Rethinking range view representation for lidar segmentation.
\newblock In \emph{IEEE/CVF International Conference on Computer Vision}, pages 228--240, 2023{\natexlab{a}}.

\bibitem[Kong et~al.(2023{\natexlab{b}})Kong, Liu, Li, Chen, Zhang, Ren, Pan, Chen, and Liu]{kong2023robo3d}
Lingdong Kong, Youquan Liu, Xin Li, Runnan Chen, Wenwei Zhang, Jiawei Ren, Liang Pan, Kai Chen, and Ziwei Liu.
\newblock Robo3d: Towards robust and reliable 3d perception against corruptions.
\newblock In \emph{IEEE/CVF International Conference on Computer Vision}, pages 19994--20006, 2023{\natexlab{b}}.

\bibitem[Kong et~al.(2023{\natexlab{c}})Kong, Niu, Xie, Hu, Ng, Cottereau, Zhao, Zhang, Wang, Ooi, Zhu, Song, Liu, Zhang, Yu, Jing, Li, Qi, Jin, Chen, Hou, Zhang, Kan, Lin, Peng, Li, Xu, Yang, Yao, Wu, Kuai, Liu, Jiang, Huang, Li, Chen, Zhang, Ao, Li, Chen, Luo, Zhao, and Yu]{kong2023robodepth_challenge}
Lingdong Kong, Yaru Niu, Shaoyuan Xie, Hanjiang Hu, Lai~Xing Ng, Benoit Cottereau, Ding Zhao, Liangjun Zhang, Hesheng Wang, Wei~Tsang Ooi, Ruijie Zhu, Ziyang Song, Li Liu, Tianzhu Zhang, Jun Yu, Mohan Jing, Pengwei Li, Xiaohua Qi, Cheng Jin, Yingfeng Chen, Jie Hou, Jie Zhang, Zhen Kan, Qiang Lin, Liang Peng, Minglei Li, Di Xu, Changpeng Yang, Yuanqi Yao, Gang Wu, Jian Kuai, Xianming Liu, Junjun Jiang, Jiamian Huang, Baojun Li, Jiale Chen, Shuang Zhang, Sun Ao, Zhenyu Li, Runze Chen, Haiyong Luo, Fang Zhao, and Jingze Yu.
\newblock The robodepth challenge: Methods and advancements towards robust depth estimation.
\newblock \emph{arXiv preprint arXiv:2307.15061}, 2023{\natexlab{c}}.

\bibitem[Kong et~al.(2023{\natexlab{d}})Kong, Niu, Xie, Hu, Ng, Cottereau, Zhao, Zhang, Wang, Ooi, et~al.]{kong2023challenge}
Lingdong Kong, Yaru Niu, Shaoyuan Xie, Hanjiang Hu, Lai~Xing Ng, Benoit~R Cottereau, Ding Zhao, Liangjun Zhang, Hesheng Wang, Wei~Tsang Ooi, et~al.
\newblock The robodepth challenge: Methods and advancements towards robust depth estimation.
\newblock \emph{arXiv preprint arXiv:2307.15061}, 2023{\natexlab{d}}.

\bibitem[Kong et~al.(2023{\natexlab{e}})Kong, Quader, and Liong]{kong2023conDA}
Lingdong Kong, Niamul Quader, and Venice~Erin Liong.
\newblock Conda: Unsupervised domain adaptation for lidar segmentation via regularized domain concatenation.
\newblock In \emph{IEEE International Conference on Robotics and Automation}, pages 9338--9345, 2023{\natexlab{e}}.

\bibitem[Kong et~al.(2023{\natexlab{f}})Kong, Ren, Pan, and Liu]{kong2022laserMix}
Lingdong Kong, Jiawei Ren, Liang Pan, and Ziwei Liu.
\newblock Lasermix for semi-supervised lidar semantic segmentation.
\newblock In \emph{IEEE/CVF Conference on Computer Vision and Pattern Recognition}, pages 21705--21715, 2023{\natexlab{f}}.

\bibitem[Kong et~al.(2023{\natexlab{g}})Kong, Xie, Hu, Cottereau, Ng, and Ooi]{kong2023robodepth_benchmark}
Lingdong Kong, Shaoyuan Xie, Hanjiang Hu, Benoit Cottereau, Lai~Xing Ng, and Wei~Tsang Ooi.
\newblock The robodepth benchmark for robust out-of-distribution depth estimation under corruptions.
\newblock \url{https://github.com/ldkong1205/RoboDepth}, 2023{\natexlab{g}}.

\bibitem[Kong et~al.(2024{\natexlab{a}})Kong, Xie, Hu, Ng, Cottereau, and Ooi]{kong2024robodepth}
Lingdong Kong, Shaoyuan Xie, Hanjiang Hu, Lai~Xing Ng, Benoit Cottereau, and Wei~Tsang Ooi.
\newblock Robodepth: Robust out-of-distribution depth estimation under corruptions.
\newblock \emph{Advances in Neural Information Processing Systems}, 36, 2024{\natexlab{a}}.

\bibitem[Kong et~al.(2024{\natexlab{b}})Kong, Xu, Ren, Zhang, Pan, Chen, Ooi, and Liu]{kong2024lasermix2}
Lingdong Kong, Xiang Xu, Jiawei Ren, Wenwei Zhang, Liang Pan, Kai Chen, Wei~Tsang Ooi, and Ziwei Liu.
\newblock Multi-modal data-efficient 3d scene understanding for autonomous driving.
\newblock \emph{arXiv preprint arXiv:2405.05258}, 2024{\natexlab{b}}.

\bibitem[Kretzschmar et~al.(2022)Kretzschmar, Liniger, Alvarez, Wang, Casser, Yu, Pavone, Li, Geiger, Ondruska, Li, Angelov, Leonard, and Gool]{ArgoverseStereo}
Henrik Kretzschmar, Alex Liniger, Jose~M. Alvarez, Yan Wang, Vincent Casser, Fisher Yu, Marco Pavone, Bo Li, Andreas Geiger, Peter Ondruska, Li~Erran Li, Dragomir Angelov, John Leonard, and Luc~Van Gool.
\newblock The argoverse stereo competition.
\newblock \url{CVPR2022.wad.vision}, 2022.

\bibitem[Kundu et~al.(2023)Kundu, Lu, Zhang, Liu, and Beerel]{kundu2023learning}
Souvik Kundu, Shunlin Lu, Yuke Zhang, Jacqueline Liu, and Peter~A Beerel.
\newblock Learning to linearize deep neural networks for secure and efficient private inference.
\newblock \emph{arXiv preprint arXiv:2301.09254}, 2023.

\bibitem[Lahiri et~al.(2024)Lahiri, Ren, and Lin]{vehicles}
Somnath Lahiri, Jing Ren, and Xianke Lin.
\newblock Deep learning-based stereopsis and monocular depth estimation techniques: A review.
\newblock \emph{Vehicles}, 6\penalty0 (1):\penalty0 305--351, 2024.

\bibitem[Lang et~al.(2019)Lang, Vora, Caesar, Zhou, Yang, and Beijbom]{lang2019pointpillars}
Alex~H Lang, Sourabh Vora, Holger Caesar, Lubing Zhou, Jiong Yang, and Oscar Beijbom.
\newblock Pointpillars: Fast encoders for object detection from point clouds.
\newblock In \emph{IEEE/CVF Conference on Computer Vision and Pattern Recognition}, pages 12697--12705, 2019.

\bibitem[Lee et~al.(2013)Lee, Fraundorfer, and Pollefeys]{lee2013robust}
Gim~Hee Lee, Friedrich Fraundorfer, and Marc Pollefeys.
\newblock Robust pose-graph loop-closures with expectation-maximization.
\newblock In \emph{IEEE/RSJ International Conference on Intelligent Robots and Systems}, pages 556--563, 2013.

\bibitem[Lee et~al.(2019)Lee, Hwang, Lee, Bae, and Park]{lee2019energy}
Youngwan Lee, Joong-won Hwang, Sangrok Lee, Yuseok Bae, and Jongyoul Park.
\newblock An energy and gpu-computation efficient backbone network for real-time object detection.
\newblock In \emph{IEEE/CVF Conference on Computer Vision and Pattern Recognition Workshops}, 2019.

\bibitem[Lee et~al.(2022)Lee, Kim, Willette, and Hwang]{lee2022mpvit}
Youngwan Lee, Jonghee Kim, Jeffrey Willette, and Sung~Ju Hwang.
\newblock Mpvit: Multi-path vision transformer for dense prediction.
\newblock In \emph{IEEE/CVF Conference on Computer Vision and Pattern Recognition}, pages 7287--7296, 2022.

\bibitem[Li et~al.(2024)Li, Sima, Dai, Wang, Lu, Wang, Zeng, Li, Yang, Deng, Tian, Xie, Xie, Chen, Li, Li, Gao, Jia, Liu, Shi, Lin, and Qiao]{li20224bev}
Hongyang Li, Chonghao Sima, Jifeng Dai, Wenhai Wang, Lewei Lu, Huijie Wang, Jia Zeng, Zhiqi Li, Jiazhi Yang, Hanming Deng, Hao Tian, Enze Xie, Jiangwei Xie, Li Chen, Tianyu Li, Yang Li, Yulu Gao, Xiaosong Jia, Si Liu, Jianping Shi, Dahua Lin, and Yu Qiao.
\newblock Delving into the devils of bird’s-eye-view perception: A review, evaluation and recipe.
\newblock \emph{IEEE Transactions on Pattern Analysis and Machine Intelligence}, 46\penalty0 (4):\penalty0 2151--2170, 2024.

\bibitem[Li et~al.(2022{\natexlab{a}})Li, Wang, Wang, and Zhao]{li2022hdmapnet}
Qi Li, Yue Wang, Yilun Wang, and Hang Zhao.
\newblock Hdmapnet: An online hd map construction and evaluation framework.
\newblock In \emph{International Conference on Robotics and Automation}, pages 4628--4634, 2022{\natexlab{a}}.

\bibitem[Li et~al.(2023{\natexlab{a}})Li, Ge, Yu, Yang, Wang, Shi, Sun, and Li]{li2023bevdepth}
Yinhao Li, Zheng Ge, Guanyi Yu, Jinrong Yang, Zengran Wang, Yukang Shi, Jianjian Sun, and Zeming Li.
\newblock Bevdepth: Acquisition of reliable depth for multi-view 3d object detection.
\newblock In \emph{AAAI Conference on Artificial Intelligence}, pages 1477--1485, 2023{\natexlab{a}}.

\bibitem[Li et~al.(2023{\natexlab{b}})Li, Li, Liu, Gong, Li, Chen, Wang, Li, Jiang, Yu, Wang, Zhao, Yu, and Feng]{li2023sscbench}
Yiming Li, Sihang Li, Xinhao Liu, Moonjun Gong, Kenan Li, Nuo Chen, Zijun Wang, Zhiheng Li, Tao Jiang, Fisher Yu, Yue Wang, Hang Zhao, Zhiding Yu, and Chen Feng.
\newblock Sscbench: Monocular 3d semantic scene completion benchmark in street views.
\newblock \emph{arXiv preprint arXiv:2306.09001}, 2023{\natexlab{b}}.

\bibitem[Li et~al.(2023{\natexlab{c}})Li, Yu, Choy, Xiao, Alvarez, Fidler, Feng, and Anandkumar]{li2023voxformer}
Yiming Li, Zhiding Yu, Christopher Choy, Chaowei Xiao, Jose~M Alvarez, Sanja Fidler, Chen Feng, and Anima Anandkumar.
\newblock Voxformer: Sparse voxel transformer for camera-based 3d semantic scene completion.
\newblock In \emph{IEEE/CVF Conference on Computer Vision and Pattern Recognition}, pages 9087--9098, 2023{\natexlab{c}}.

\bibitem[Li et~al.(2022{\natexlab{b}})Li, Wang, Li, Xie, Sima, Lu, Qiao, and Dai]{li2022bevformer}
Zhiqi Li, Wenhai Wang, Hongyang Li, Enze Xie, Chonghao Sima, Tong Lu, Yu Qiao, and Jifeng Dai.
\newblock Bevformer: Learning bird’s-eye-view representation from multi-camera images via spatiotemporal transformers.
\newblock In \emph{European Conference on Computer Vision}, pages 1--18, 2022{\natexlab{b}}.

\bibitem[Lin et~al.(2022)Lin, Lin, Pei, Huang, and Su]{lin2022sparse4d}
Xuewu Lin, Tianwei Lin, Zixiang Pei, Lichao Huang, and Zhizhong Su.
\newblock Sparse4d: Multi-view 3d object detection with sparse spatial-temporal fusion.
\newblock \emph{arXiv preprint arXiv:2211.10581}, 2022.

\bibitem[Liu et~al.(2024{\natexlab{a}})Liu, Huang, Zhang, Yao, Zhang, Wan, Ye, and Zhou]{liu2024ray}
Feng Liu, Tengteng Huang, Qianjing Zhang, Haotian Yao, Chi Zhang, Fang Wan, Qixiang Ye, and Yanzhao Zhou.
\newblock Ray denoising: Depth-aware hard negative sampling for multi-view 3d object detection.
\newblock \emph{arXiv preprint arXiv:2402.03634}, 2024{\natexlab{a}}.

\bibitem[Liu et~al.(2022)Liu, Wang, Zhang, and Sun]{liu2022petr}
Yingfei Liu, Tiancai Wang, Xiangyu Zhang, and Jian Sun.
\newblock Petr: Position embedding transformation for multi-view 3d object detection.
\newblock In \emph{European Conference on Computer Vision}, pages 531--548, 2022.

\bibitem[Liu et~al.(2023{\natexlab{a}})Liu, Chen, Li, Kong, Yang, Xia, Bai, Zhu, Ma, Li, Qiao, and Hou]{liu2023uniseg}
Youquan Liu, Runnan Chen, Xin Li, Lingdong Kong, Yuchen Yang, Zhaoyang Xia, Yeqi Bai, Xinge Zhu, Yuexin Ma, Yikang Li, Yu Qiao, and Yuenan Hou.
\newblock Uniseg: A unified multi-modal lidar segmentation network and the openpcseg codebase.
\newblock In \emph{IEEE/CVF International Conference on Computer Vision}, pages 21662--21673, 2023{\natexlab{a}}.

\bibitem[Liu et~al.(2023{\natexlab{b}})Liu, Yan, Jia, Li, Gao, Wang, and Zhang]{liu2023petrv2}
Yingfei Liu, Junjie Yan, Fan Jia, Shuailin Li, Aqi Gao, Tiancai Wang, and Xiangyu Zhang.
\newblock Petrv2: A unified framework for 3d perception from multi-camera images.
\newblock In \emph{IEEE/CVF International Conference on Computer Vision}, pages 3262--3272, 2023{\natexlab{b}}.

\bibitem[Liu et~al.(2024{\natexlab{b}})Liu, Kong, Cen, Chen, Zhang, Pan, Chen, and Liu]{liu2023segment}
Youquan Liu, Lingdong Kong, Jun Cen, Runnan Chen, Wenwei Zhang, Liang Pan, Kai Chen, and Ziwei Liu.
\newblock Segment any point cloud sequences by distilling vision foundation models.
\newblock In \emph{Advances in Neural Information Processing Systems}, 2024{\natexlab{b}}.

\bibitem[Liu et~al.(2024{\natexlab{c}})Liu, Kong, Wu, Chen, Li, Pan, Liu, and Ma]{liu2024m3net}
Youquan Liu, Lingdong Kong, Xiaoyang Wu, Runnan Chen, Xin Li, Liang Pan, Ziwei Liu, and Yuexin Ma.
\newblock Multi-space alignments towards universal lidar segmentation.
\newblock In \emph{IEEE/CVF Conference on Computer Vision and Pattern Recognition}, 2024{\natexlab{c}}.

\bibitem[Liu et~al.(2021)Liu, Lin, Cao, Hu, Wei, Zhang, Lin, and Guo]{liu2021swin}
Ze Liu, Yutong Lin, Yue Cao, Han Hu, Yixuan Wei, Zheng Zhang, Stephen Lin, and Baining Guo.
\newblock Swin transformer: Hierarchical vision transformer using shifted windows.
\newblock In \emph{IEEE/CVF International Conference on Computer Vision}, pages 10012--10022, 2021.

\bibitem[Liu et~al.(2023{\natexlab{c}})Liu, Tang, Amini, Yang, Mao, Rus, and Han]{liu2023bevfusion}
Zhijian Liu, Haotian Tang, Alexander Amini, Xinyu Yang, Huizi Mao, Daniela~L. Rus, and Song Han.
\newblock Bevfusion: Multi-task multi-sensor fusion with unified bird's-eye view representation.
\newblock In \emph{IEEE International Conference on Robotics and Automation}, pages 2774--2781, 2023{\natexlab{c}}.

\bibitem[Ma et~al.(2022)Ma, Wang, Bai, Yang, Hou, Wang, Qiao, Yang, Manocha, and Zhu]{ma2022vision}
Yuexin Ma, Tai Wang, Xuyang Bai, Huitong Yang, Yuenan Hou, Yaming Wang, Yu Qiao, Ruigang Yang, Dinesh Manocha, and Xinge Zhu.
\newblock Vision-centric bev perception: A survey.
\newblock \emph{arXiv preprint arXiv:2208.02797}, 2022.

\bibitem[Mendes et~al.(2016)Mendes, Koch, and Lacroix]{mendes2016icp}
Ellon Mendes, Pierrick Koch, and Simon Lacroix.
\newblock Icp-based pose-graph slam.
\newblock In \emph{IEEE International Symposium on Safety, Security, and Rescue Robotics}, pages 195--200, 2016.

\bibitem[Miao et~al.(2023)Miao, Liu, Chen, Gong, Xu, Hu, and Zhou]{miao2023occdepth}
Ruihang Miao, Weizhou Liu, Mingrui Chen, Zheng Gong, Weixin Xu, Chen Hu, and Shuchang Zhou.
\newblock Occdepth: A depth-aware method for 3d semantic scene completion.
\newblock \emph{arXiv preprint arXiv:2302.13540}, 2023.

\bibitem[Milioto et~al.(2019)Milioto, Vizzo, Behley, and Stachniss]{milioto2019rangenet++}
Andres Milioto, Ignacio Vizzo, Jens Behley, and Cyrill Stachniss.
\newblock Rangenet++: Fast and accurate lidar semantic segmentation.
\newblock In \emph{IEEE/RSJ International Conference on Intelligent Robots and Systems}, pages 4213--4220, 2019.

\bibitem[Moravec and Elfes(1985)]{moravec1985high}
Hans Moravec and Alberto Elfes.
\newblock High resolution maps from wide angle sonar.
\newblock In \emph{IEEE International Conference on Robotics and Automation}, pages 116--121, 1985.

\bibitem[Oquab et~al.(2023)Oquab, Darcet, Moutakanni, Vo, Szafraniec, Khalidov, Fernandez, Haziza, Massa, El-Nouby, et~al.]{oquab2023dinov2}
Maxime Oquab, Timoth{\'e}e Darcet, Th{\'e}o Moutakanni, Huy Vo, Marc Szafraniec, Vasil Khalidov, Pierre Fernandez, Daniel Haziza, Francisco Massa, Alaaeldin El-Nouby, et~al.
\newblock Dinov2: Learning robust visual features without supervision.
\newblock \emph{arXiv preprint arXiv:2304.07193}, 2023.

\bibitem[Park et~al.(2022)Park, Xu, Yang, Keutzer, Kitani, Tomizuka, and Zhan]{park2022time}
Jinhyung Park, Chenfeng Xu, Shijia Yang, Kurt Keutzer, Kris~M Kitani, Masayoshi Tomizuka, and Wei Zhan.
\newblock Time will tell: New outlooks and a baseline for temporal multi-view 3d object detection.
\newblock In \emph{International Conference on Learning Representations}, 2022.

\bibitem[Peng et~al.(2023)Peng, Chen, Fu, Liang, and Cheng]{peng2023bevsegformer}
Lang Peng, Zhirong Chen, Zhangjie Fu, Pengpeng Liang, and Erkang Cheng.
\newblock Bevsegformer: Bird's eye view semantic segmentation from arbitrary camera rigs.
\newblock In \emph{IEEE/CVF Winter Conference on Applications of Computer Vision}, pages 5935--5943, 2023.

\bibitem[Ramirez et~al.(2023)Ramirez, Tosi, Stefano, Timofte, Costanzino, andSamuele Salti, Mattoccia, Shi, Zhang, A, Jin, Li, Li, Liu, Zhang, Wang, and Yin]{NTIRE}
Pierluigi~Zama Ramirez, Fabio Tosi, Luigi~Di Stefano, Radu Timofte, Alex Costanzino, Matteo~Poggi andSamuele Salti, Stefano Mattoccia, Jun Shi, Dafeng Zhang, Yong A, Yixiang Jin, Dingzhe Li, Chao Li, Zhiwen Liu, Qi Zhang, Yixing Wang, and Shi Yin.
\newblock Ntire 2023 challenge on hr depth from images of specular and transparent surfaces.
\newblock In \emph{IEEE/CVF Conference on Computer Vision and Pattern Recognition Workshops}, pages 1384--1395, 2023.

\bibitem[Recht et~al.(2019)Recht, Roelofs, Schmidt, and Shankar]{recht2019imagenet}
Benjamin Recht, Rebecca Roelofs, Ludwig Schmidt, and Vaishaal Shankar.
\newblock Do imagenet classifiers generalize to imagenet?
\newblock In \emph{International Conference on Machine Learning}, pages 5389--5400. PMLR, 2019.

\bibitem[Shan et~al.(2020)Shan, Englot, Meyers, Wang, Ratti, and Rus]{shan2020lio}
Tixiao Shan, Brendan Englot, Drew Meyers, Wei Wang, Carlo Ratti, and Daniela Rus.
\newblock Lio-sam: Tightly-coupled lidar inertial odometry via smoothing and mapping.
\newblock In \emph{IEEE/RSJ international conference on intelligent robots and systems}, pages 5135--5142, 2020.

\bibitem[Shao et~al.(2023)Shao, Pei, Chen, Li, Liu, and Li]{shao2023urcdc}
Shuwei Shao, Zhongcai Pei, Weihai Chen, Ran Li, Zhong Liu, and Zhengguo Li.
\newblock Urcdc-depth: Uncertainty rectified cross-distillation with cutflip for monocular depth estimation.
\newblock \emph{IEEE Transactions on Multimedia}, 2023.

\bibitem[Spencer et~al.(2023{\natexlab{a}})Spencer, Qian, Russell, Hadfield, Graf, Adams, Schofield, Elder, Bowden, Cong, Mattoccia, Poggi, Suri, Tang, Tosi, Wang, Zhang, Zhang, and Zhao]{MDEC}
Jaime Spencer, C.~Stella Qian, Chris Russell, Simon Hadfield, Erich Graf, Wendy Adams, Andrew~J. Schofield, James~H. Elder, Richard Bowden, Heng Cong, Stefano Mattoccia, Matteo Poggi, Zeeshan~Khan Suri, Yang Tang, Fabio Tosi, Hao Wang, Youmin Zhang, Yusheng Zhang, and Chaoqiang Zhao.
\newblock The monocular depth estimation challenge.
\newblock In \emph{IEEE/CVF Winter Conference on Applications of Computer Vision Workshops}, pages 623--632, 2023{\natexlab{a}}.

\bibitem[Spencer et~al.(2023{\natexlab{b}})Spencer, Qian, Trescakova, Russell, Hadfield, Graf, Adams, Schofield, Elder, Bowden, Anwar, Chen, Chen, Cheng, Dai, Hoa, Hossain, Huang, Jing, Li, Li, Li, Liu, Mattoccia, Mercelis, Nam, Poggi, Qi, Ren, Tang, Tosi, Trinh, Uddin, Umair, Wang, Wang, Wang, Xiang, Xu, Yin, Yu, Zhang, and Zhao]{MDEC2}
Jaime Spencer, C.~Stella Qian, Michaela Trescakova, Chris Russell, Simon Hadfield, Erich Graf, Wendy Adams, Andrew~J. Schofield, James Elder, Richard Bowden, Ali Anwar, Hao Chen, Xiaozhi Chen, Kai Cheng, Yuchao Dai, Huynh~Thai Hoa, Sadat Hossain, Jianmian Huang, Mohan Jing, Bo Li, Chao Li, Baojun Li, Zhiwen Liu, Stefano Mattoccia, Siegfried Mercelis, Myungwoo Nam, Matteo Poggi, Xiaohua Qi, Jiahui Ren, Yang Tang, Fabio Tosi, Linh Trinh, S~M~Nadim Uddin, Khan~Muhammad Umair, Kaixuan Wang, Yufei Wang, Yixing Wang, Mochu Xiang, Guangkai Xu, Wei Yin, Jun Yu, Qi Zhang, and Chaoqiang Zhao.
\newblock The second monocular depth estimation challenge.
\newblock In \emph{IEEE/CVF Conference on Computer Vision and Pattern Recognition Workshops}, pages 3063--3075, 2023{\natexlab{b}}.

\bibitem[Srivastava et~al.(2014)Srivastava, Hinton, Krizhevsky, Sutskever, and Salakhutdinov]{srivastava2014dropout}
Nitish Srivastava, Geoffrey Hinton, Alex Krizhevsky, Ilya Sutskever, and Ruslan Salakhutdinov.
\newblock Dropout: a simple way to prevent neural networks from overfitting.
\newblock \emph{The Journal of Machine Learning Research}, 15\penalty0 (1):\penalty0 1929--1958, 2014.

\bibitem[Sun et~al.(2020)Sun, Kretzschmar, Dotiwalla, Chouard, Patnaik, Tsui, Guo, Zhou, Chai, Caine, et~al.]{sun2020scalability}
Pei Sun, Henrik Kretzschmar, Xerxes Dotiwalla, Aurelien Chouard, Vijaysai Patnaik, Paul Tsui, James Guo, Yin Zhou, Yuning Chai, Benjamin Caine, et~al.
\newblock Scalability in perception for autonomous driving: Waymo open dataset.
\newblock In \emph{IEEE/CVF Conference on Computer Vision and Pattern Recognition}, pages 2446--2454, 2020.

\bibitem[Thrun(2002)]{thrun2002probabilistic}
Sebastian Thrun.
\newblock Probabilistic robotics.
\newblock \emph{Communications of the ACM}, 45\penalty0 (3):\penalty0 52--57, 2002.

\bibitem[Tian et~al.(2024)Tian, Jiang, Yun, Mao, Yang, Wang, Wang, and Zhao]{tian2024occ3d}
Xiaoyu Tian, Tao Jiang, Longfei Yun, Yucheng Mao, Huitong Yang, Yue Wang, Yilun Wang, and Hang Zhao.
\newblock Occ3d: A large-scale 3d occupancy prediction benchmark for autonomous driving.
\newblock \emph{Advances in Neural Information Processing Systems}, 36, 2024.

\bibitem[Wang et~al.(2023)Wang, Liu, Wang, Li, and Zhang]{wang2023exploring}
Shihao Wang, Yingfei Liu, Tiancai Wang, Ying Li, and Xiangyu Zhang.
\newblock Exploring object-centric temporal modeling for efficient multi-view 3d object detection.
\newblock In \emph{IEEE/CVF International Conference on Computer Vision}, pages 3621--3631, 2023.

\bibitem[Wang et~al.(2022)Wang, Guizilini, Zhang, Wang, Zhao, and Solomon]{wang2022detr3d}
Yue Wang, Vitor~Campagnolo Guizilini, Tianyuan Zhang, Yilun Wang, Hang Zhao, and Justin Solomon.
\newblock Detr3d: 3d object detection from multi-view images via 3d-to-2d queries.
\newblock In \emph{Conference on Robot Learning}, pages 180--191. PMLR, 2022.

\bibitem[Watson et~al.(2019)Watson, Firman, Brostow, and Turmukhambetov]{watson2019hints}
Jamie Watson, Michael Firman, Gabriel~J. Brostow, and Daniyar Turmukhambetov.
\newblock Self-supervised monocular depth hints.
\newblock In \emph{IEEE/CVF International Conference on Computer Vision}, pages 2162--2171, 2019.

\bibitem[Wei et~al.(2023{\natexlab{a}})Wei, Zhao, Zheng, Zhu, Rao, Huang, Lu, and Zhou]{wei2023surrounddepth}
Yi Wei, Linqing Zhao, Wenzhao Zheng, Zheng Zhu, Yongming Rao, Guan Huang, Jiwen Lu, and Jie Zhou.
\newblock Surrounddepth: Entangling surrounding views for self-supervised multi-camera depth estimation.
\newblock In \emph{Conference on Robot Learning}, pages 539--549. PMLR, 2023{\natexlab{a}}.

\bibitem[Wei et~al.(2023{\natexlab{b}})Wei, Zhao, Zheng, Zhu, Zhou, and Lu]{wei2023surroundocc}
Yi Wei, Linqing Zhao, Wenzhao Zheng, Zheng Zhu, Jie Zhou, and Jiwen Lu.
\newblock Surroundocc: Multi-camera 3d occupancy prediction for autonomous driving.
\newblock In \emph{IEEE/CVF International Conference on Computer Vision}, pages 21729--21740, 2023{\natexlab{b}}.

\bibitem[Xie et~al.(2023)Xie, Kong, Zhang, Ren, Pan, Chen, and Liu]{xie2023robobev}
Shaoyuan Xie, Lingdong Kong, Wenwei Zhang, Jiawei Ren, Liang Pan, Kai Chen, and Ziwei Liu.
\newblock Robobev: Towards robust bird's eye view perception under corruptions.
\newblock \emph{arXiv preprint arXiv:2304.06719}, 2023.

\bibitem[Xie et~al.(2024)Xie, Kong, Zhang, Ren, Pan, Chen, and Liu]{xie2024benchmarking}
Shaoyuan Xie, Lingdong Kong, Wenwei Zhang, Jiawei Ren, Liang Pan, Kai Chen, and Ziwei Liu.
\newblock Benchmarking and improving bird's eye view perception robustness in autonomous driving.
\newblock \emph{arXiv preprint arXiv:2405.17426}, 2024.

\bibitem[Yan et~al.(2023)Yan, Liu, Sun, Jia, Li, Wang, and Zhang]{yan2023cross}
Junjie Yan, Yingfei Liu, Jianjian Sun, Fan Jia, Shuailin Li, Tiancai Wang, and Xiangyu Zhang.
\newblock Cross modal transformer: Towards fast and robust 3d object detection.
\newblock In \emph{IEEE/CVF International Conference on Computer Vision}, pages 18268--18278, 2023.

\bibitem[Yang et~al.(2023)Yang, Chen, Tian, Tao, Zhu, Zhang, Huang, Li, Qiao, Lu, et~al.]{yang2023bevformer}
Chenyu Yang, Yuntao Chen, Hao Tian, Chenxin Tao, Xizhou Zhu, Zhaoxiang Zhang, Gao Huang, Hongyang Li, Yu Qiao, Lewei Lu, et~al.
\newblock Bevformer v2: Adapting modern image backbones to bird's-eye-view recognition via perspective supervision.
\newblock In \emph{IEEE/CVF Conference on Computer Vision and Pattern Recognition}, pages 17830--17839, 2023.

\bibitem[Yang et~al.(2018)Yang, Zhu, Nian, Feng, Qu, and Ma]{yang2018robust}
Sheng Yang, Xiaoling Zhu, Xing Nian, Lu Feng, Xiaozhi Qu, and Teng Ma.
\newblock A robust pose graph approach for city scale lidar mapping.
\newblock In \emph{IEEE/RSJ International Conference on Intelligent Robots and Systems}, pages 1175--1182, 2018.

\bibitem[Yao et~al.(2018)Yao, Luo, Li, Fang, and Quan]{mvsnet}
Yao Yao, Zixin Luo, Shiwei Li, Tian Fang, and Long Quan.
\newblock Mvsnet: Depth inference for unstructured multi-view stereo.
\newblock In \emph{European Conference on Computer Vision}, pages 767--783, 2018.

\bibitem[Yin et~al.(2021)Yin, Zhou, and Krahenbuhl]{yin2021center}
Tianwei Yin, Xingyi Zhou, and Philipp Krahenbuhl.
\newblock Center-based 3d object detection and tracking.
\newblock In \emph{IEEE/CVF Conference on Computer Vision and Pattern Recognition}, pages 11784--11793, 2021.

\bibitem[Zamir et~al.(2022)Zamir, Arora, Khan, Hayat, Khan, and Yang]{zamir2022restormer}
Syed~Waqas Zamir, Aditya Arora, Salman Khan, Munawar Hayat, Fahad~Shahbaz Khan, and Ming-Hsuan Yang.
\newblock Restormer: Efficient transformer for high-resolution image restoration.
\newblock In \emph{IEEE/CVF Conference on Computer Vision and Pattern Recognition}, pages 5728--5739, 2022.

\bibitem[Zendel et~al.(2022)Zendel, Dai, Fernandez, Geiger, Koltun, Kontschieder, Kortylewski, Lin, Sattler, Scharstein, Schilling, Uhrig, and Wulff]{RVC}
Oliver Zendel, Angela Dai, Xavier~Puig Fernandez, Andreas Geiger, Vladen Koltun, Peter Kontschieder, Adam Kortylewski, Tsung-Yi Lin, Torsten Sattler, Daniel Scharstein, Hendrik Schilling, Jonas Uhrig, and Jonas Wulff.
\newblock The robust vision challenge.
\newblock \url{http://www.robustvision.net}, 2022.

\bibitem[Zhang et~al.(2022)Zhang, Zhu, Zheng, Huang, Huang, Zhou, and Lu]{zhang2022beverse}
Yunpeng Zhang, Zheng Zhu, Wenzhao Zheng, Junjie Huang, Guan Huang, Jie Zhou, and Jiwen Lu.
\newblock Beverse: Unified perception and prediction in birds-eye-view for vision-centric autonomous driving.
\newblock \emph{arXiv preprint arXiv:2205.09743}, 2022.

\bibitem[Zhao et~al.(2022)Zhao, Zhang, Poggi, Tosi, Guo, Zhu, Huang, Tang, and Mattoccia]{zhao2022monovit}
Chaoqiang Zhao, Youmin Zhang, Matteo Poggi, Fabio Tosi, Xianda Guo, Zheng Zhu, Guan Huang, Yang Tang, and Stefano Mattoccia.
\newblock Monovit: Self-supervised monocular depth estimation with a vision transformer.
\newblock In \emph{International Conference on 3D Vision}, pages 668--678, 2022.

\bibitem[Zhou and Tuzel(2018)]{voxelnet}
Yin Zhou and Oncel Tuzel.
\newblock Voxelnet: End-to-end learning for point cloud based 3d object detection.
\newblock In \emph{IEEE/CVF Conference on Computer Vision and Pattern Recognition}, pages 4490--4499, 2018.

\bibitem[Zhu et~al.(2019)Zhu, Jiang, Zhou, Li, and Yu]{cbgs}
Benjin Zhu, Zhengkai Jiang, Xiangxin Zhou, Zeming Li, and Gang Yu.
\newblock Class-balanced grouping and sampling for point cloud 3d object detection.
\newblock \emph{arXiv preprint arXiv:1908.09492}, 2019.

\bibitem[Zhu et~al.(2020)Zhu, Su, Lu, Li, Wang, and Dai]{zhu2020deformable}
Xizhou Zhu, Weijie Su, Lewei Lu, Bin Li, Xiaogang Wang, and Jifeng Dai.
\newblock Deformable detr: Deformable transformers for end-to-end object detection.
\newblock \emph{arXiv preprint arXiv:2010.04159}, 2020.

\end{thebibliography}
